\documentclass[10pt,twocolumn,letterpaper]{article}

\usepackage{iccv}
\usepackage{times}
\usepackage{epsfig}
\usepackage{graphicx}
\usepackage{amsmath}
\usepackage{amssymb}
\usepackage{pifont}

\usepackage{booktabs}
\usepackage{arydshln}
\usepackage{algorithm}% http://ctan.org/pkg/algorithm
\usepackage{algpseudocode}% http://ctan.org/pkg/algorithmicx
\usepackage{multirow}
\usepackage{url}

\newcommand{\cmark}{\ding{51}}%
\newcommand{\xmark}{\ding{55}}%

\newcommand{\cD}{\mathcal{D}}

\newcommand{\cO}{\mathcal{O}}

\newcommand{\cY}{\mathcal{Y}}

\newcommand{\bw}{\mathbf{w}}

\newcommand{\bmu}{\boldsymbol{\mu}}
\newcommand{\bx}{\mathbf{x}}

\newcommand{\bb}{\mathbf{b}}

\newcommand{\btheta}{\boldsymbol{\theta}}

\newcommand{\bxi}{\boldsymbol{\xi}}

\newcommand{\bbR}{\mathbb{R}}
\newcommand{\bbN}{\mathbb{N}}

\newcommand{\suppl}{Appendix}

\usepackage[pagebackref=true,breaklinks=true,colorlinks,bookmarks=false]{hyperref}

\usepackage[capitalize]{cleveref}
\crefname{section}{Sec.}{Secs.}
\Crefname{section}{Section}{Sections}
\Crefname{table}{Table}{Tables}
\crefname{table}{Tab.}{Tabs.}

\iccvfinalcopy % *** Uncomment this line for the final submission

\def\iccvPaperID{} % *** Enter the ICCV Paper ID here

\ificcvfinal\fi

\begin{document}

%\title{Is Continuous Adaptation of Features Essential in Class-Incremental Learning?}

\title{First Session Adaptation: A Strong Replay-Free Baseline for\\ Class-Incremental Learning}
% not sure whether features, feature extractors, neural network body, etc. is better

% Do we really need learning continuously in Continual Learning?
% Is Continuous Adaptation of Features Necessary in Continual Learning?
% Is Continual Adaptation of the Neural Network Body Useful for Continual Learning?

\author{Aristeidis Panos\\
University of Cambridge\\
\and
Yuriko Kobe\\
University of Cambridge\\
\and
Daniel Olmeda Reino\\
Toyota Motor Europe\\
\and
Rahaf Aljundi\\
Toyota Motor Europe\\
\and
Richard E. Turner\\
University of Cambridge\\
}

\maketitle
\ificcvfinal\thispagestyle{empty}\fi

\begin{abstract}
In Class-Incremental Learning (CIL) an image classification system is exposed to new classes in each learning session and must be updated incrementally. Methods approaching this problem have updated both the classification head and the feature extractor body at each session of CIL. In this work, we develop a baseline method, First Session Adaptation (FSA), that sheds light on the efficacy of existing CIL approaches, and allows us to assess the relative performance contributions from head and body adaption. FSA adapts a pre-trained neural network body only on the first learning session and fixes it thereafter; a head based on linear discriminant analysis (LDA), is then placed on top of the adapted body, allowing exact updates through CIL. 
FSA is replay-free i.e.~it does not memorize examples from previous sessions of continual learning.
To empirically motivate FSA, we first consider a diverse selection of 22 image-classification datasets, evaluating different heads and body adaptation techniques in high/low-shot offline settings. We find that the LDA head performs well and supports CIL out-of-the-box. 
We also find that Featurewise Layer Modulation (FiLM) adapters are highly effective in the few-shot setting, and full-body adaption in the high-shot setting. 
Second, we empirically investigate various CIL settings including high-shot CIL and few-shot CIL, including settings that have previously been used in the literature. We show that FSA significantly improves over the state-of-the-art in 15 of the 16 settings considered. FSA with FiLM adapters is especially performant in the few-shot setting. These results indicate that current approaches to continuous body adaptation are not working as expected.
Finally, we propose a measure that can be applied to a set of unlabelled inputs which is predictive of the benefits of body adaptation.

%Finally, we use offline baselines to form an upper bound on the benefits of body adaptation in 22 benchmarks. We show that a pre-trained network with no body adaptation often performs well, especially in the few-shot setting, and recommend the benchmarks the CIL community should use in the future.
%
\end{abstract}

\section{Introduction}
% general intro to continual learning
Continual learning (CL) is needed to bring machine learning models to many real-life applications. After a model is trained on a given set of data, and once deployed in  its test environment, it is likely that new classes will naturally emerge. For example, in an autonomous driving scenario, new types of road transportation and traffic signage can be encountered.  The deployed model needs to efficiently acquire this new knowledge with minimal cost (e.g.~annotation requirements) without deteriorating the performance on existing classes of objects. The process of efficiently acquiring new knowledge while preserving what is already captured by the model is what continual learning methods target.

% types of continual learning / why it arises
Continual learning can be mainly divided into task incremental learning and class incremental learning. Task incremental learning (TIL) sequentially learns independent sets of heterogeneous tasks and is out of this paper's scope. Class incremental learning (CIL), in contrast, assumes that all classes (already learnt and future ones) form part of a single classification task. 
% perhaps add one or two more sentences about general context?
In class-incremental learning, data arrive in a sequence of sessions with new classes appearing as the sessions progress. Critically, the data in each session cannot be stored in its entirety and cannot be revisited. The goal is to train a single classification model under these constraints. 

In both task incremental and class incremental learning, the data distribution will change as the sessions progress. However, these changes tend to be smaller in real-world class incremental learning settings than in the task incremental setting. For, example consider a human support robot learning about new objects in a home in an incremental manner (CIL) vs.~the same robot learning to adapt to different homes (TIL).

Practically, CIL has two major uses. First, in situations where large amounts of data arrive in each session and so retraining the model is computationally prohibitive.  Second, in situations where data are not allowed to be revisited due to privacy reasons such as under General Data Protection Regulation (GDPR). The latter situation is relevant for applications such as personalization where small numbers of data points are available i.e.~few-shot continual learning.
So-called replay-free methods are necessary for such settings, as samples of previous sessions are not allowed to be memorized, but CIL is known to be challenging in such settings.

%In CIL forgetting can be severe when samples of previous sessions are not memorized. Replay-free methods thus become of high importance in such scenario. 

Current SOTA methods for class incremental learning start from a pre-trained backbone~\cite{lesort2022scaling,ostapenko2022continual,wang2022learning,wang2022dualprompt} and then adapt the features at each session of continual learning. The use of a pre-trained backbone has been shown to lead to strong performance especially in the few-shot CIL setting due to the lack of data~\cite{peng2022few}. However, it is unclear to what extent continuous adaption of the features in each session is helpful. In general, in CIL there is a trade-off between adaptation (which helps adapt to the new statistics of the target domain) and catastrophic forgetting (whereby earlier classes are forgotten as the representation changes).  When memorization of previous samples is restricted, the benefits from adapting the feature extractor body to learn better features might well be out-weighted by the increase in forgetting of old classes. Moreover, body adaptation in earlier sessions is arguably more critical (e.g.~adapting the backbone to the new domain in the first session), whilst it is less essential in later sessions where the changes in the ideal representation between sessions are smaller.

This work explores under what conditions continuous adaptation of the body is beneficial, both in principle and in current practice. In order to do this, we develop a new replay-free method, inspired by the first encoding method of \cite{bateni2020improved}, called First Session Adaptation (FSA). FSA adapts the body of a pre-trained neural network on only the first session of continual learning. We investigate adapting all body parameters and the use of Feature-wise Layer Modulation (FiLM) adapters \cite{perez2018film} which learn only a small number of parameters. The head, in contrast, is adapted at each session using an approach that is similar to Linear Discriminate Analysis (LDA), which suffers from no forgetting, and improves over the Nearest Class Mean classifier (NCM) \cite{mensink2013distance} approach. The efficacy of this general approach, including comparisons to a standard linear head, is first motivated through experiments in the offline setting (\cref{sec:head_comparisons}).    
%
%We then consider continual learning, first considering the high-shot setting (high-shot CIL). Next, we consider few-shot continual learning following previous approaches that employ an initial session with a large number of data points and few-shot sessions thereafter (few-shot+ CIL). Finally, we consider few-shot continual learning in which each task contains only a small amount of data (few-shot CIL).  
We then carry out experiments under three CIL settings. First, we consider the high-shot setting (high-shot CIL). The other two settings consider few-shot continual learning. Specifically, one setting follows previous approaches that employ an initial session with a large number of data points and few-shot sessions thereafter (few-shot+ CIL) while the other exclusively includes sessions with only a small amount of data (few-shot CIL).  

The contributions of the paper are as follows: (1) We develop a replay-free CIL baseline, namely FSA, that is extremely simple to implement and performs well in many different scenarios. (2) We empirically motivate FSA through a set of offline experiments that evaluate different forms of neural network head and body adaptation;  (3) We then compare FSA to a set of strong continual learning baseline methods in a fair way using the same pre-trained backbone for all methods. (4) We show that the FSA-FiLM baseline performs well in the high-shot CIL setting, outperforming the SOTA whilst avoiding data  memorization. (5) In the few-shot learning settings, FSA outperforms existing continual learning methods on eight benchmarks, often by a substantial margin, and is statistically tied with the best performing method on the one remaining dataset. (6) Finally, we propose a measure that can be applied to a set of unlabelled inputs which is predictive of the benefits of body adaptation.
 %   \item We find the FiLM approach to be especially useful when there is relatively little data in the target domain and it is relatively far from the source domain used for pre-training. 
%\end{itemize}

%[?] three important facets of algorithms: regularization (keeping new parameters close to old ones), memory-based approaches, architecture - position work in this context: alternative to regularization, could be combined with a memory (but don't look at it here), explore the use of different architectures

\section{Related Work}
%In this work, we focus on continual learning for classification problems. The goal is to learn a representation that is beneficial for all classes being learnt and to incrementally construct a classifier with limited or restricted access to data of classes learnt in the past. This setting is referred to as Class Incremental Learning. This setting differs from Task Incremental learning scenario that focuses on learning a shared representation while deploying a separate classifier for each task. We refer to \cite{de2021continual} for a survey.
% \textbf{Task Incremental Learning}
% One form of continual learning is to assume that each training session belongs to a separate task that is learned using a separate classification head. Methods usually rely on regularization to prevent forgetting of previous sessions. However, it is assumed that the task ID is known at test time. 
%\aresnote{Cite suggested papers by CVPR's reviewers.}

Class Incremental Learning is more challenging than task-incremental continual learning as a shared output layer is used for  the classes of different learning sessions. We refer to \cite{de2021continual} for a survey on both class and task incremental learning. While softmax cross entropy loss is considered a standard in classification models, it is shown to be a source of class interference in continual learning~\cite{wu2019large,caccia2022new}. Thus, recent works either focus on fixing the softmax classifier~\cite{caccia2022new,Ahn_2021_ICCV,zhao2020maintaining} or deploy  nearest class mean classifiers (NCM) as an alternative~\cite{rebuffi2017icarl,davari2022probing,kang2022class}. In this work, we deploy an LDA classifier that uses the mean embedding of each class and an incrementally updated covariance  matrix shared across all classes. We show that this approach is comparable to a softmax classifier in the offline setting and that it outperforms NCM.

Due to the challenging nature of class incremental learning, some methods  employ a buffer of stored samples from previous sessions~\cite{prabhu2020gdumb,yan2021dynamically}. \cite{prabhu2020gdumb} suggests a simple baseline, GDumb, that performs offline training on both buffer data and new session data at each incremental step. GDumb shows strong performance compared to sophisticated continual learning solutions. Here we show that our simple yet effective solution often outperforms variants of GDumb that have a large memory without leveraging any buffer of stored samples.

In addition to the issues arising from the shared classification layer, learning a feature extractor that can provide representative features of all classes is a key element in continual learning. Some class incremental learning methods leverage large initial training session or start from a pretrained network, e.g.,~\cite{wang2022learning,hersche2022constrained,wu2021striking} and show improved overall continual learning performance without questioning the role of such pre-training steps and whether the learned representations have in fact improved. Recently, \cite{ostapenko2022continual} studies continual learning with strong pre-trained models and proposes deploying exemplar-based replay in the latent space using a small multilayer network on top of the fixed pre-trained model. The authors show that such latent replay improves performance, especially for tasks that are different from  the distribution used to train the initial model. In this work, we show that adaptation of pre-trained models is essential for strong continual learning performance. However, different from all existing continual learning works, we show that adapting the representation only in the first session is sufficient to obtain representative features for the classes being learned and that this forms a strong baseline for future continual learning methods. 

In addition to the incremental classification in the full data regime, we focus on the few-shot scenario where only a few labeled examples per class are provided at each training session.
% Recently \cite{wu2022class} examines training with a model pre-trained on a large number of base classes with different settings of CL.  They branch the network after some layer and then after learning new classes, they fuse the branches.
\cite{hersche2022constrained} utilize a meta-trained network for few-shot classification and explore multiple design choices   including a no-adaptation baseline and a method to optimize orthogonal prototypes with the last classification layer only. The method heavily relies on a pre-training and meta-training step on a large subset of classes from the original dataset. In this work, we show that the meta-training step is not essential and that a simple LDA head is sufficient  without the reliance on a big initial training stage. 
% Mode 1: no network retrained with prototypes as the average of class samples (features).
% Mode 2: optimize bipolarized prototypes by adding noise to the average prototypes. The FC layer is trained using stored activations to maximize the similarities between the class activations and the prototypes. 
% Mode 3: Instead of bipolarizing the prototypes they optimize nudged prototypes: prototypes that minimize inter class similarities and remain close to the initial avenged prototypes. Then they optimize the FC layer to align the features of the stored activations with the nudged prototypes. 
%
In  \cite{kang2022class} 
the authors propose a new distillation loss functioning at the feature map level where importance values are estimated per feature map along with replay and consider an NCM classifier at test time.
\cite{zhou2022forward}
heavily relies on the training of an initial session using a specific loss function that takes into account classes that will be added in future. It further requires that the total number of classes to be known \emph{a priori} which is unrealistic for real applications.
Our solution adapts FILM parameters to the first session data and fixes the representation for the remaining training sessions. We show that surprisingly few samples in the first session are sufficient for  adaptation  and that our solution is more powerful than current few-shot continual learning solutions. 

Finally, the LDA head has been also utilized for CIL in~\cite{hayes2020lifelong}. However, in ~\cite{hayes2020lifelong} a slightly different regularized covariance matrix was used and the pre-trained feature extractor was frozen across sessions without fine-tuning at all. As we show in our experiments, this strategy is sub-optimal and fine-tuning the FiLM parameters offers a significant performance boost.

% {\color{blue}Finally, the LDA head has been also utilized for CIL in~\cite{hayes2020lifelong}. However, in their work, the use slightly different regularized covariance matrix and they do not fine-tune the pre-trained feature extractor across sessions at all. As we show in our experiments, this strategy is sub-optimal and fine-tuning the FiLM parameters offers a significant performance boost. }

\section{Proposed Algorithm}\label{sec:methodology}
In this section, first we discuss the main components of the proposed methods FSA/FSA-FiLM, namely body adaptation techniques and classifier heads. Then we formally introduce the two methods for tackling CIL.
\subsection{Problem Formulation}
In CIL, we are given a dataset $\cD_s = \{ \bx_{i,s}, y_{i,s} \}_{i=1}^{N_s}$ for each session $s \in \{1, \ldots, S\}$, where $X_s = \{ \bx_{i,s} \}_{i=1}^{N_s}$ is a set of images and $Y_s = \{ y_{i,s} \}_{i=1}^{N_s}$ is the set of the corresponding labels with $y_{i,s} \in \cY_s$. Here $\cY_s$ is the label space of session $s$. 
%A key characteristic of 
It is common in CIL that label spaces are mutually exclusive across sessions, i.e.~$\cY_s \cap \cY_{s^{\prime}} = \varnothing,~\forall s \neq s^{\prime}$ and we only have access to $\cD_s$ in the current session $s$ to train our model. The proposed FSA method can naturally handle sessions with overlapping label spaces too, however, we follow the former paradigm in our experiments. Another typical assumption in CIL is that the data in all sessions come from the same dataset. We also adopt this assumption in this work, although our experiments on DomainNet are a step toward considering different datasets in each session. 

In session $s$, we will use models defined as 
\begin{equation}
f_s (\bx) = W_s^{\top} g_{\btheta_s}(\bx)  + \bb_s, \label{eq:model}
\end{equation}
where $g_{\btheta_s}(\bx) \in \bbR^d$ is a feature extractor backbone\footnote{In this paper, we interchangeably use the terms feature extractor, backbone, and body.} with session-dependent parameters $\btheta_s$. The linear classifier head comprises the class weights $W_s \in \bbR^{d \times |\cY_{1:s}|}$ and biases $\bb_s \in \bbR^{|\cY_{1:s}|}$, and $\cY_{1:s} = \cup_{j=1}^s \cY_j$ is the label space of all distinct classes we have seen up to session $s$. Ideally, when a new dataset $\cD_s$ is available at the $s$-th session, the model should be able to update the backbone parameters $\btheta$ and the linear classifier head $W_s$ with the minimum computational overhead and without compromising performance over the previously seen classes $\cY_{1:s}$.

%\vspace*{-0.3 cm}
\paragraph{Backbone adaptation.} Traditionally, the adaptation of the body at the current session $s$ requires updating the full set of  the network parameters $\btheta_s$ (full-body adaptation). Options include using specifically designed loss functions that mitigate catastrophic forgetting or using a memory buffer that stores a subset of the previously encountered data. See \cite{mai2022online} for a review of continual learning techniques. Nevertheless, when training data availability is scarce relative to the size of the model then full-body adaptation ceases to be suitable, and few-shot learning techniques \cite{wang2020generalizing}, such as meta-learning \cite{hospedales2021meta} and transfer learning \cite{yosinski2014transferable} become favorable.

%Briefly, a meta-learner is trained on a large number of tasks in order to learn-how-to-learn in order to generalize well.  when new tasks arrive while in transfer learning.
%
%we start with a pre-trained backbone on a large upstream dataset \cite{kolesnikov2020big} and then we update the full set or a subset of the model parameters on a downstream task. 

Recent works in few-shot learning \cite{requeima2019fast} showed that a pre-trained backbone can be efficiently adapted by keeping its parameters $\btheta_s$ frozen and introducing Feature-wise Linear Modulation (FiLM) \cite{perez2018film} layers with additional parameters $\bxi_s$ which scale and shift the activations produced by a convolutional layer. 
In \cite{requeima2019fast}, these parameters are generated by a meta-trained network when a downstream task is given. Alternatively, $\bxi_s$ can be learned (fine-tuned) by the downstream data as in \cite{shysheya2022fit}. In this work, we consider both meta-learned (Supplement) and fine-tuned (\cref{sec:experiments}) FiLM parameters.

%\vspace*{-0.2 cm}
\paragraph{Classifier heads.}
In both offline and CIL settings, (linear) classifier heads can be divided into two groups; parametrized and parameter-free. Parametrized heads require the weights/biases in \cref{eq:model} to be updated by an iterative gradient-based procedure whilst for parameter-free heads, a closed-form formula is available that directly computes the weights/biases after the update of the backbone. We consider three different heads (for notational simplicity we consider the offline setting, and thus suppress session $s$ dependence), one parametrized, and two parameter-free.
\begin{itemize}
    \item The linear (learnable) head where $W, \bb$  are learned.
    
    \item The Nearest Class Mean (NCM) classifier where the weight and bias of the $k$-th class is given by
    \begin{equation}\label{eq:ncm}
        \bw_k = \hat{\bmu}_k~\text{and}~b_k = \ln \frac{|X^{(k)}|}{N}-\frac{1}{2} \hat{\bmu}_k^{\top} \hat{\bmu}_k  , 
    \end{equation}
    where $X^{(k)} = \{\bx_i : y_i = k \}$ is the set of images belonging in class $k$ and $\hat{\bmu}_k =  \frac{1}{|X^{(k)}|} \sum_{\bx \in X^{(k)}} g_{\btheta}(\bx)$ is the mean vector of the embedded images of class $k$.

    \item The Linear Discriminant Analysis (LDA) classifier with weights/biases defined as
    \begin{equation}\label{eq:lda}
        \bw_k = \Tilde{S}^{-1}\hat{\bmu}_k~\text{and}~b_k = \ln \frac{|X^{(k)}|}{N}-\frac{1}{2} \hat{\bmu}_k^{\top} \Tilde{S}^{-1} \hat{\bmu}_k ,
    \end{equation}
    where $\Tilde{S} = S + I_d$ is the regularized sample covariance matrix with sample covariance matrix $S = \frac{1}{|X|-1} \sum_{\bx \in X} (g_{\btheta}(\bx) - \hat{\bmu}) (g_{\btheta}(\bx) - \hat{\bmu})^{\top} $, $\hat{\bmu} =  \frac{1}{|X|} \sum_{\bx \in X} g_{\btheta}(\bx)$, and identity matrix $I_d \in \bbR^{d \times d}$.
\end{itemize}

Both NCM and LDA classifiers are suitable for CIL since their closed-form updates in \cref{eq:ncm} and \cref{eq:lda} support exact, computationally inexpensive, continual updates (via running averages) that incur no forgetting. Updating the linear head, in contrast, requires more computational effort and machinery when novel classes appear in a new session. Notice that setting $\Tilde{S} = I_d$ in LDA recovers NCM. 

%Notice that NCM is a special case of LDA when you set $\Tilde{S} = I_d$ for LDA. 

The continuous update of the LDA head is not as straightforward as it is for NCM, since LDA  also requires updating the sample covariance $S$. We discuss how this can be attained in the next section where we introduce our proposed method FSA/FSA-FiLM.

%Without loss of generality, we can assume that we have two sessions in total, i.e. $S=2$, and after creating the LDA head in the first session using \cref{eq:lda} for dataset $X_1$, we want to update the head using the data of the second session $X_2$. This can be done by computing and storing the following quantities:

%define CIL mathematically
%- define sessions

%backbones
%- general discussion 
%- FiLM

%heads
%- linear head
%- LDA head
%- explain how the head can updated incrementally

\subsection{First Session Adaptation}
Motivated by the efficient adaptation techniques and CIL-friendly classifiers discussed in the previous section, we propose two simple and computationally efficient CIL methods called First Session Adaptation (FSA) and FSA-FiLM. FSA is based on a full-body adaptation while FSA-FiLM uses FiLM layers to adapt the body. Both methods utilize pre-trained backbones. FSA (-FiLM)  (i) adapts the backbone only at the first session and then the backbone remains frozen for the rest of the sessions, and (ii) makes use of an LDA classifier which is continuously updated as new data become available. %Despite the computational convenience of LDA, we found empirically that adapting the body using an LDA head makes the optimization process  unstable and slow, leading to sub-optimal solutions. 
For adapting the body (either full or FiLM-based adaptation) at the first session,  we use a linear head and a cross-entropy loss first, and then after the optimization is over, the linear head is removed  and an LDA head is deployed instead, based on the optimized parameters $\btheta^*$ and \cref{eq:lda}. Updating the LDA head when the data of the next session becomes available, can be done by using  \cref{algo:lda_update} recursively until the last session $S$. Specifically, by having access to the running terms $\{A, \bb, \textrm{count}\}$, the sample covariance matrix is given by
\begin{equation}
S = \frac{1}{\textrm{count}-1} \left(A - \frac{1}{\textrm{count}} \bb \bb^{\top} \right). \label{eq:updated_sample_cov}
\end{equation}
\begin{algorithm}
  \caption{Update of sample covariance running terms}\label{algo:lda_update}
  \begin{algorithmic}[1]
    \Require $g(\cdot) \equiv$ a feature extractor backbone
    \Require $X_s = \{ \bx_{i,s} \}_{i=1}^{N_s}$: Images of session $s$
    \Require $A \in \bbR^ {d \times d}, \bb \in \bbR^d, \textrm{count} \in \bbN^*$ if $s > 1$
    \Function{IncUpdate}{$g(\cdot), X_s, A, \bb, \textrm{count}$}
       \If{$s = 1$} \Comment{Initialize $A, \bb, \textrm{count}$}
          \State $A\gets \mathbf{0}_{d \times d}, \bb \gets \mathbf{0}_d, \textrm{count}\gets 0$
          %\State $\bb \gets \mathbf{0}_d$
          %\State $\textrm{count}\gets 0$
        \EndIf
      \State $A\gets A + \sum_{\bx \in X_s} g(\bx) g(\bx)^{\top}$
      \State $\bb \gets \bb + \sum_{\bx \in X_s} g(\bx)$
      \State $\textrm{count}\gets \textrm{count} + N_s$
      \State \textbf{return} $A, \bb, \textrm{count}$ 
    \EndFunction
  \end{algorithmic}
\end{algorithm}
The time complexity of LDA scales as $\cO (d^3 + |\cY_s| d^2)$ and its space complexity is $\cO (d^2 +  |\cY_{1:s}| d)$ at the $s$-th session.
This computational burden is negligible compared to taking gradient steps for learning a linear head and as we will see in \cref{sec:head_comparisons} using covariance information boosts performance significantly against the vanilla NCM head.
%\vspace*{-0.3cm}

% Add differences Beyond Simple Meta-Learning:
% Multi-Purpose Models for Multi-Domain, Active
% and Continual Few-Shot Learning

%pre-trained backbone
%adapt on first session of continual learning using a linear head, fix the backbone thereafter
%replace linear head with LDA head and update head in an incremental way thereafter

\section{Experiments}\label{sec:experiments}
In this section, we present a series of experiments to investigate the performance of FSA (-FiLM) under different CIL settings. First, we detail the datasets used for the experiments and discuss our implementation. Second, we perform a large empirical comparison between LDA, linear, and NCM classifiers, in combination with different body adaptation techniques, in the offline setting. For the CIL settings, the results are compared to the vanilla method, \textit{No Adaptation (NA)}, where a pretrained backbone is combined with an LDA head and remains frozen across sessions. We also consider how these findings are affected by the number of shots available. Third, we compare FSA (-FiLM) with state-of-the-art continual learning methods and conduct ablation studies. Finally, we investigate the effect of adapting the backbone when the CIL dataset is ``similar'' to ImageNet-1k and discuss a similarity metric that indicates whether body adaptation is required.

\subsection{Datasets and Implementation details}\label{sec:datasets}
\paragraph{Datasets.} We employ a diverse collection of 26 image-classification datasets across the offline and CIL experiments. For the offline experiments of \cref{sec:head_comparisons}, we use the 19 datasets of VTAB \cite{zhai2019large}, a low-shot transfer learning benchmark, plus three additional datasets, FGVC-Aircraft \cite{maji2013fine}, Stanford Cars \cite{krause20133d}, and Letters \cite{deCampos2009character}. We refer to this group of datasets as VTAB+. For the CIL-based experiments, we choose 5 VTAB+ datasets from different domains, having an adequate number of classes to create realistic CIL scenarios. The datasets are CIFAR100, SVHN, dSprites-location, FGVC-Aircraft, Cars, and Letters while also including 4 extra datasets: CUB200 \cite{WahCUB_200_2011}, CORE50 \cite{lomonaco2017core50}, iNaturalist \cite{van2018inaturalist}, and DomainNet \cite{peng2019moment}. Exact details for each dataset are provided in the Supplement.

\paragraph{Training details.} All models are implemented with PyTorch \cite{paszke2019pytorch}. We use a pre-trained EfficientNet-B0 \cite{tan2019efficientnet} on Imagenet-1k as the main backbone for all methods. For the few-shot+ CIL experiment in \cref{sec:cil_comparisons}, we also consider two ResNet architectures, ResNet-18 and ResNet-20 \cite{he2016deep} to enable direct comparison to the original settings used in \cite{zhou2022forward}. All the deployed backbones (except ResNet-20, due to the unavailability of pre-trained weights on ImageNet-1k), are pre-trained on ImageNet-1k for all methods. We keep the optimization settings the same across all baselines for fairness. Optimization details are given in the Supplement.

%\vspace*{-0.3 cm}
\paragraph{Evaluation protocol.} We report the Top-1 accuracy after the last continual learning session evaluated on a test set including instances from all the previously seen classes. In the Supplement, we provide the accuracy after each session. To quantify the forgetting behavior, we use a scaled modification of the performance dropping rate \cite{tao2020few}, namely percent performance dropping rate (PPDR), defined as $ \text{PPDR} = 100 \times \frac{\mathcal{A}_1 - \mathcal{A}_S}{\mathcal{A}_1} \%$, where $\mathcal{A}_1$ denotes  test accuracy after the first session, and $\mathcal{A}_S$  the test accuracy after the last session.

\begin{figure*}[htb]
\centering
\includegraphics[width = \textwidth]{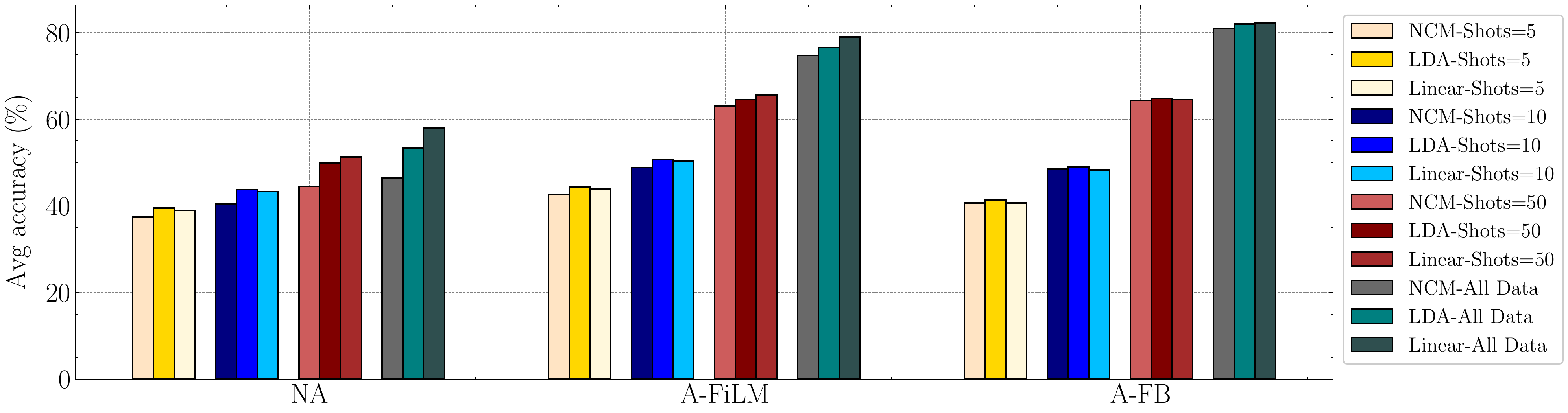} 
\caption{Average accuracy across all VTAB+ datasets using no-adaptation (NA), FiLM adaptation (A-FiLM), and full body adaptation (A-FB) for different classifier heads (NCM, LDA, Linear) and number of shots (5, 10, 50, All Data). The results correspond to the offline setting where all classes are available without any incremental learning.}
\label{fig:head_comparisons}
\end{figure*}

\subsection{Head Comparisons}\label{sec:head_comparisons}
To motivate the choice of the LDA head for FSA (-FiLM), we compare its average accuracy across all VTAB+ datasets with that of the linear and NCM classifiers under the \emph{offline} setting where all the classes are available and no incremental learning is required. For the body, we use no adaptation (NA) as a baseline (i.e.~using the unadapted original backbone), FiLM-based adaptation, and full-body adaptation while ranging the number of shots from 5 to 10, then 50, and finally using all available training data. As Fig.~\ref{fig:head_comparisons} illustrates, LDA consistently outperforms both NCM and linear heads when the number of shots is equal to 5 and 10, regardless of adaptation technique. Notice also that as the number of learnable parameters increases, the choice of the head plays a less significant role in the predictive power of the method. Overall, LDA performs  similarly to the linear head in the high-shot settings and  performs the best of all heads in the low-shot settings. Finally, the full covariance information of LDA provides a consistent advantage over its isotropic covariance counterpart, i.e.~the NCM classifier. This is in agreement with the results presented in~\cite{shysheya2022fit} for the LDA and NCM classifier. In addition to the results presented here, we have also tested meta-learning based body adaptation methods~\cite{bronskill2021memory,requeima2019fast,bateni2020improved} which support continual learning out of the box. We find these   perform poorly compared to the fine-tuning based methods. See next section for more results.

\begin{table*}
%\vspace*{1.5 cm}
%\begingroup
\setlength{\tabcolsep}{4.9pt} % Default value: 6pt
\begin{center}
\begin{small}
%\begin{sc}
\begin{tabular}{l c c c c c c c | c}
\toprule
%& \multicolumn{5}{c}{\textbf{Datasets}} \\
%\cmidrule(lr){2-6}
\textbf{Method} & \textbf{CIFAR100} & \textbf{CORE50} & \textbf{SVHN} &  \textbf{dSprites-loc} &  \textbf{FGVC-Aircraft} & \textbf{Cars}  & \textbf{Letters}  & \textbf{Avg Diff} \\
 \midrule
NA & 68.2 (26.8) & 82.6 (14.2) & 39.9 (51.1) & 20.6 (54.0) & 41.3 (1.7) & 43.3 (40.3) &  68.4 (24.1) & 0.0\\
\hdashline
E-EWC+SDC~\cite{yu2020semantic} & 32.4 (66.7) &  21.7 (78.1) & 39.5 (60.1) & 18.6 (81.4) & 25.6 (55.8) & 30.0 (62.6) & 33.6 (66.3) & -23.3\\
FACT~\cite{zhou2022forward} & 10.2 (89.4) & 22.0 (77.7) & 33.8 (65.9) & 6.4 (93.6) & 4.7 (90.3) & 0.6 (99.3) &  20.9 (79.1) & -38.0 \\
ALICE~\cite{peng2022few} & 52.4 (45.8) & 72.8 (25.8) & 46.1 (53.6) & 68.3 (31.7) & 39.8 (35.0)& 36.4 (56.0) & 75.7 (24.2) & +3.9\\
FSA & 62.8 (34.5) & 82.8 (15.5)  & 71.3 (26.6) & \textbf{91.5} (\textbf{8.5}) & 50.8 (6.3)  & 50.3 (36.6)  & 78.4 (21.4) & +17.7 \\
FSA-LL & 60.5 (37.2) & 79.0 (19.2) & 64.6 (33.1) & 91.3 (8.5) & 45.4 (21.7) & 45.7 (43.8) & 77.2 (22.7) & +14.2 \\
FSA-FiLM & \textbf{73.8} (\textbf{23.4}) & \textbf{85.4} (\textbf{13.3})  & \textbf{75.9} (\textbf{23.4}) &  76.9 (22.8)  &  \textbf{55.9} (\textbf{-5.7}) & \textbf{55.9} (\textbf{30.2})  & \textbf{79.7} (\textbf{20.0}) & \textbf{+19.9} \\
\midrule
GDumb~\cite{prabhu2020gdumb} & 54.5 (42.2) & 82.4 (15.0) & 78.2 (19.6) & 79.5 (12.9) & 25.3 (56.9)  & 14.2 (82.6) & 70.1 (27.0) & +5.7 \\
\midrule
Offline-FiLM & 78.2 & 88.4 & 93.1  & 98.5 & 67.5 & 67.3 & 85.2 & +30.6\\
\bottomrule
\end{tabular}
%\end{sc}
\end{small}
\end{center}
\caption{Last session's test accuracy (\%) ($\uparrow$) and the PPDR (\%) ($\downarrow$) in parentheses, for the high-shot CIL setting (\cref{sec:cil_comparisons}). The last column reports the average accuracy difference ($\uparrow$) across all datasets between  a baseline and NA. A pre-trained EfficientNet-B0 on Imagenet-1k is used as a backbone for all methods. For reference, we include the replay-based baseline GDumb where 1k images are used for the memory buffer.}
\label{table:cifar100_core50_fullshots}
%\endgroup
%\vspace*{-0.2cm}
\end{table*}

\subsection{Class-Incremental Learning Comparisons}\label{sec:cil_comparisons}
We consider three different CIL scenarios: (i) high-shot CIL, (ii) few-shot+ CIL, and (iii) few-shot CIL. We compare our FSA/FSA-FiLM methods with recent state-of-the-art FSCIL methods, including  Decoupled-Cosine~\cite{vinyals2016matching}, CEC~\cite{zhang2021few}, FACT~\cite{zhou2022forward}, and ALICE. Additionally, for the high-shot CIL setting, we consider a strong replay-based baseline for CIL,  GDumb~\cite{prabhu2020gdumb} and a competitive replay-free method, E-EWC+SDC~\cite{yu2020semantic}. Finally, we introduce an additional baseline adapter inspired by \cite{wu2022class}, called FSA-LL (Last Layer). In FSA-LL only the parameters of the backbone's last block are fine-tuned which can be compared to the FSA-FiLM adaption method. %\rahaf{It seems that we have never defined NA and what head it uses.}

%\vspace*{-0.2cm}
\paragraph{High-shot CIL.} In this setting, we consider all the available training data for each class while keeping the number of novel classes in each session low. We use CIFAR100, CORE50, SVHN, dSprites-loc, FGVC-Aircraft, Cars, and Letters for our experiments. For CIFAR100, CORE50, and FGVC-Aircraft, the first session consists of 10 classes, 4 for dSprites-loc, 16 for Cars, and 12 for letters. The rest of the sessions include 10 classes for CIFAR100, FGVC-Aircraft, 20 for Cars, 5 for CORE50 and Letters, 2 for SVHN and dSprites-loc. Therefore, the total number of incremental sessions for CIFAR100, FGVC-Aircraft, and Cars is 10 while for CORE50 we have 9 sessions. A pre-trained on Imagenet-1k EfficientNet-B0 is deployed as a backbone for all methods. FSA-FiLM outperforms all the competitors by a significant margin on all datasets except dSprites-loc. We attribute the performance gap on dSprites-loc due to the large number of data points (25k) and  the portion of the classes ($25\%$) in the first session. It is also apparent the efficiency of fine-tuning FiLM parameters over fine-tuning the parameters of the model's last layer; note the number of FiLM parameters is only $20.5$k while the last layer of an EfficientNet-B0 comprises $2.9$M parameters. Furthermore, fine-tuning the FiLM parameters offer almost $20\%$ accuracy increase on average compared to using a pretrained model without doing any further adaptation. Interestingly, the replay-free FSA-FILM is able to outperform significantly the replay-based method GDumb with a 1k memory buffer on most of the datasets; accuracy and time comparisons between FSA-FiLM and GDumb  with varying buffer sizes can be found in Figure 1 of the \suppl. Regarding FACT's perfromance, although it starts from a pre-trained backbone, it was developed under the assumption that the first session contains lots of data and a high fraction of the classes that will be encountered ($> 50 \%$) which is not true for the
setting treated in \Cref{table:cifar100_core50_fullshots}. FACT overfits in this setting and
this results in poor performance. It turns out that FACT’s
assumptions about the first session are a strong requirement
which is not necessary for obtaining good performance as
our FSA baseline shows.

\paragraph{Few-shot+ CIL.} This setting is the one that has most commonly been used for few-shot CIL (FSCIL) and it involves an initial session that contains a large number of classes (around 50-60\% of the total number of classes in the dataset) and all the available training images of these classes. The remaining sessions comprise a small number of classes and shots (typically 5/10-way 5-shot). Here we follow the exact FSCIL settings as described in \cite{zhou2022forward,peng2022few} for CIFAR100 and CUB200. We use ResNet-18/20 and EfficientNet-B0 as backbones. \Cref{table:fact_vs_our} summarizes the performance comparison between baselines. FSA performs on par with FACT on CIFAR100 when we use the original backbone used in \cite{zhou2022forward}, and it outperforms FACT by almost 10\%  and ALICE by 3.4\% when EfficientNet-B0 is utilized while FSA-FiLM exhibits the lowest PPDR score. Notice also that FSA is only marginally worse than its offline counterpart, meaning there is little room for continuous body adaptation to improve things further. For CUB200, FSA with an EfficientNet-B0 performs on par with ALICE. Interestingly, we observe that NA performs well on this dataset. This indicates that CUB200 is not far from ImageNet-1k. The current results of FSA set new SOTA performance on CIFAR100 for the FSCIL setting. %For CUB200, FSA with a ResNet-18 already outperforms all SOTA baselines while similar patterns are present when  EfficientNet-B0 is used. Interestingly, we observe that NA performs well on this dataset. This indicates that CUB200 is not far from ImageNet-1k. The current results of FSA/FSA-FiLM set new SOTA performance on CIFAR100 and CORE50 for the FSCIL setting.
\begin{table}[htb]

\setlength{\tabcolsep}{7.4pt} % Default value: 6pt
\begin{center}
    
\begin{small}
%\begin{sc}

\begin{tabular}{l c c c}
\toprule
& & \multicolumn{2}{c}{ \textbf{Datasets}} \\
\cmidrule(lr){3-4}  \textbf{Method} & \textbf{Backbone} &  \textbf{CIFAR100} &  \textbf{CUB200} \\
\midrule
CEC~\cite{zhang2021few} & \multirow{3}{*}{RN-20} & 49.1 (32.7)* & - \\
FACT~\cite{zhou2022forward} & & 52.1 (30.2)* & - \\ 
FSA & & 52.0 (30.8) & - \\ 
\hdashline
NA & \multirow{6}{*}{RN-18} & 50.4 (26.8) & 50.0 (29.3) \\ 
CEC~\cite{zhang2021few} & & - & 52.3 (31.1)* \\
FACT~\cite{zhou2022forward} & & 49.5 (34.8)* & 56.9 (25.0)* \\
ALICE~\cite{peng2022few} &  & 54.1 (31.5$)^{\dagger}$ & 60.1 (22.4$)^{\dagger}$ \\
FSA-FiLM &  & 55.2 (24.4) & 52.7 (27.6) \\ 
FSA & & 61.4 (25.1) & 57.6 (24.3) \\ 
\hdashline
NA & \multirow{6}{*}{EN-B0} & 55.2 (25.8) &  63.2 (\textbf{19.6}) \\ 
FACT~\cite{zhou2022forward} & & 56.5 (34.6) & 62.9 (23.3) \\ 
ALICE~\cite{peng2022few} & & 62.7 (28.4) &  \textbf{63.5} (22.2) \\ 
FSA-LL &  & 61.4 (25.7) & 55.9 (22.3) \\ 
FSA-FiLM &  & 61.8 (\textbf{22.4}) & 62.9 (20.4) \\ 
FSA &  &  \textbf{66.1} (24.6) & 63.4 (20.9) \\ 
\midrule
Offline & EN-B0 & 67.0 & 65.1 \\
\bottomrule
\end{tabular}
%\end{sc}
\end{small}
\end{center}
\caption{Baseline comparison under the few-shot+ CIL setting (\cref{sec:cil_comparisons}). We report the accuracy (\%) ($\uparrow$) of the last session and the PPDR (\%) ($\downarrow$) in parentheses. Asterisk (*) indicates that the reported results are from \cite{zhou2022forward} and $\dagger$ indicates results reported in \cite{peng2022few}. We use three different backbones, EfficientNet-B0 (EN-B0) and ResNet-18/20 (RN-18/20); EN-B0 and RN-18 are pre-trained on Imagenet-1k.}
\label{table:fact_vs_our}
\end{table}

%\vspace*{-.9cm}
\paragraph{Few-shot CIL.} The final setting, which is firstly introduced in this work, considers an alternative  FSCIL setting in which only a small number of data points are available in all sessions, including the first. We use 50 shots per session while the first session includes 20\% of the total number of classes of the dataset. Each of the remaining sessions includes around 10\% of the total number of classes; more details are available in the Supplement. We repeat experiments 5 times and we report the mean accuracy and PPDR in \Cref{table:50shot_CL_acc_ppd}. FSA-FiLM outperforms all the other baselines by a large margin in terms of both accuracy and PPDR, indicating that transfer learning is considerably advantageous for CIL when the data availability is low. Notice that both ALICE and FACT struggle to achieve good performance under this setting due to the limited amount of data in the first session. We find 
that FGVC-Aircraft exhibits a positive backward transfer behavior
that we attribute to a  difficult initial session followed by sessions that contain classes that are comparatively easier to distinguish. 

Finally, we demonstrate the suitability of the LDA head in FSA-FiLM compared to  a nearest class mean head, denoted FSA-FiLM-NCM. In the continual learning
setting, LDA gives far larger improvements over NCM; e.g. 10\% on average as presented in \Cref{table:50shot_CL_acc_ppd}. This highlights the importance of  a strong classification layer. Note that the incremental update of LDA does not require any replay buffer as opposed to a linear head.

\begin{table*}[htb]
%\vspace*{1.5 cm}
%\begingroup
\setlength{\tabcolsep}{4.8pt} % Default value: 6pt
\begin{center}
\begin{small}
%\begin{sc}
\begin{tabular}{l c c c c c c c | c}
\toprule
 \textbf{Method/Dataset} & \textbf{CIFAR100} & \textbf{SVHN} &  \textbf{dSprites-loc} &  \textbf{FGVC-Aircraft} & \textbf{Letters}  & \textbf{DomainNet} & \textbf{iNaturalist} & \textbf{Avg Diff} \\
 \midrule
NA & 57.4 (28.4) & 28.3 (61.5) & 11.9 (66.6) & 41.0 (-15.5) & 57.6 (29.9) & 69.0 (17.2) & 49.7 (4.3) & 0.0\\
\hdashline
FACT~\cite{zhou2022forward} & 16.8 (79.7) & 24.1 (66.2) & 11.7 (64.1) & 8.3 (79.9) & 49.8 (40.9) & 20.6 (75.6) & 14.3 (74.0) & -24.2\\
ALICE~\cite{peng2022few} & 58.0 (33.8) & 23.0 (65.9) & 23.0 (46.0) & 42.0 (27.5) & 66.5 (31.4) & 66.5 (25.1) & 47.9 (13.7) & +1.7\\
FSA & 60.3 (27.0) & 32.9 (53.5) & 33.7 (\textbf{41.3}) & 50.1 (-16.8) & 62.2 (28.5) & 70.3 (17.5) & 51.5 (\textbf{1.1}) & +6.6\\
FSA-LL & 62.0 (25.7) & 43.5 (47.0) & 18.8 (61.7) & 45.8 (-2.1) & 69.4 (26.1) & 67.6 (16.9) & 49.2 (14.1) & +5.9 \\
FSA-FiLM-NCM & 66.4 (24.8) & 38.8 (56.5) & 25.1 (59.4) & 42.5 (-7.5) & 57.1 (34.9) & 68.8 (18.5) & 51.5 (12.1) & +5.0\\
FSA-FiLM & \textbf{70.9} (\textbf{20.5}) & \textbf{51.3} (\textbf{43.5}) & \textbf{35.7} (43.1) & \textbf{55.8} (\textbf{-19.8}) & \textbf{73.4} (\textbf{22.1}) & \textbf{74.0} (\textbf{15.6}) & \textbf{58.8} (4.8) & +15.0 \\
\midrule
Offline-FiLM & 73.8 & 77.2 & 83.7 & 65.1 & 79.7 & 75.1 & 62.1 & +28.0 \\
\bottomrule
\end{tabular}
%\end{sc}
\end{small}
\end{center}
\caption{Accuracy (\%) ($\uparrow$)  of the last session and PPDR (\%) ($\downarrow$) in parentheses for the few-shot CIL setting of \cref{sec:cil_comparisons}. The last column reports the average accuracy difference ($\uparrow$) across all datasets between  a baseline and NA. A pre-trained EfficientNet-B0 on ImageNet-1k is used as a backbone for all methods. FSA-FiLM-NCM utilizes an NCM classifier while NA, FSA (-LL, FiLM) and Offline uses an LDA head.}
\label{table:50shot_CL_acc_ppd}
%\endgroup
\end{table*}

%-----------------------------------------------------------------------------
\begin{table}[ht]
%\begingroup
\setlength{\tabcolsep}{7.5pt} % Default value: 6pt
\begin{center}
\begin{footnotesize}
%\begin{sc}
\begin{tabular}{l c c c c c}
\toprule
\textbf{Dataset} & \textbf{NA} & \textbf{Aq} & \textbf{DevF} &  \textbf{Veh} \\
 \midrule
CIFAR100 & 57.4{\scriptsize $\pm 1.0$} & 66.1{\scriptsize $\pm 1.6$} & 66.2{\scriptsize $\pm 1.9$} & 67.9{\scriptsize $\pm 1.2$} \\
\midrule
\midrule
& \textbf{NA} & \textbf{El-R} & \textbf{F-Clp} &  \textbf{Tr-Sk} \\
\midrule
DomainNet & 69.0{\scriptsize $\pm 0.4$} & 70.6{\scriptsize $\pm 0.6$} & 71.7{\scriptsize $\pm 0.6$} & 72.8{\scriptsize $\pm 0.5$} \\
\bottomrule
\end{tabular}
%\end{sc}
\end{footnotesize}
\end{center}
\caption{Last session accuracy (\%) ($\uparrow$) of FSA-FiLM for three different session splits on CIFAR100 and DomainNet as described in \cref{sec:cil_nonhomogeneous}. For reference, we also include the accuracy of NA. Results are averaged over 5 runs (mean$\pm$std).}
\label{table:cifar100_domainnet_superclass}
%\endgroup
\end{table}
\subsection{Inhomogeneous Class-Incremental Learning}\label{sec:cil_nonhomogeneous}

FSA adapts the body only on the first session of continual learning and it is therefore likely to be sensitive to the particular classes which are present in this session. To investigate the degree of performance sensitivity for FSA, we devise a setting similar to the few-shot CIL setting of \cref{sec:cil_comparisons}, where each session includes classes that share some common attribute. We select CIFAR100 which provides superclass information, and DomainNet which consists of different domains and also has superclass information available.
We create three distinct CIL configurations for each dataset, each of which has different types of data in the first session. For CIFAR100, we split the data into 19 sessions. 
The first session includes 10 classes from super-classes (i) aquatic mammals and fish (\textit{Aq}), (ii) electric devices and furniture (\textit{DevF}), or (iii) Vehicles 1 and Vehicles 2 (\textit{Veh}). Each of the other 18 sessions include images from the remaining 18 super-classes.
Similarly, for DomainNet, we split the data into 6 sessions using 50 shots with 10 classes in each session. The first session includes 10 classes of (i) the ``electricity'' superclass of the real domain (\textit{El-R}), (ii) the ``furniture'' superclass of the clipart domain (\textit{F-Clp}), or (iii) the ``transportation'' superclass of the sketch domain (\textit{Tr-Sk}). In this way, we can vary the content of the first session and analyze the effect this has on performance.
\Cref{table:cifar100_domainnet_superclass} reveals that FSA-FiLM's performance is similar even if the data in the first session used for the body adaptation appears disparate from that contained in the remaining sessions. Even for DomainNet where the distribution shift of the data across sessions is considerable, performance is only marginally affected in each of the three settings. This provides evidence that adapting the body only on the first session achieves competitive performance regardless of the class order, given the assumption that the data come from a single dataset (albeit a varied one in the case of DomainNet).

\begin{figure}[ht]
\centering
\includegraphics[width =\linewidth]{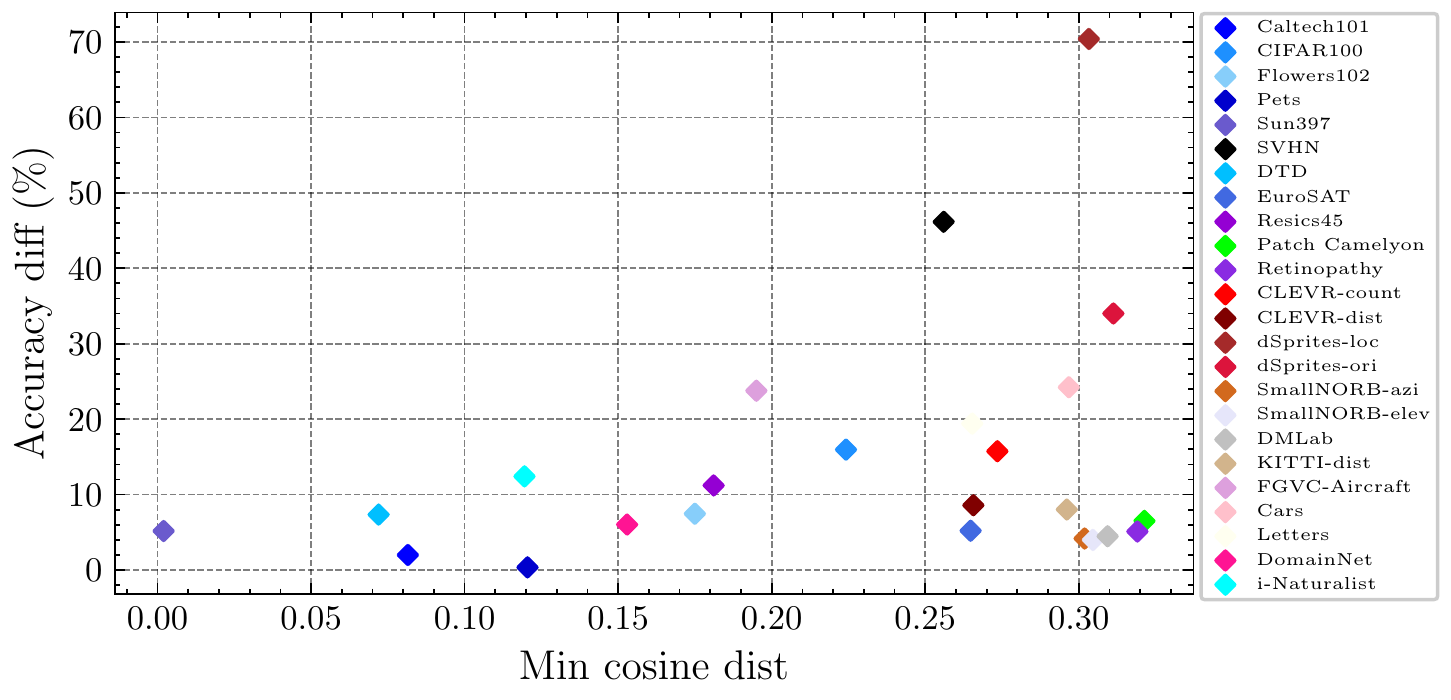} 
\caption{Scatter plot of the accuracy differences between FSA-FiLM and NA against the minimum cosine distance between a dataset and \emph{mini}Imagenet dataset evaluated using the NA method. We consider the offline setting with 50 shots. A pre-trained EfficientNet-B0 on ImageNet-1k
is used as a backbone.}
\label{fig:ch_index}
\end{figure}
%\vspace*{-.18cm}
\subsection{When to adapt the body?}\label{sec:body_adaptation}
% old name: The Effect of Body Adaptation in Low-Data Availability
Despite the strong performance of FSA (-FiLM) in the few-shot settings of \cref{sec:cil_comparisons}, there are cases where the NA method achieves very close accuracy to FSA (e.g.~see CUB200 results in \Cref{table:fact_vs_our}). This implies that there may be datasets where adaptation is not required and all we need is the pre-trained backbone. In order to decide whether we require body adaptation or not, 
%given the data of the first session, we utilize the Calinski-Harabasz (CH) index \cite{calinski1974dendrite} which quantifies how dense and well-separated are the classes. The score is evaluated on the training data of the first session through the embedded representations of the pre-trained backbone. Larger scores indicate that the pre-trained backbone is sufficient and no adaptation is required. 
we compute the minimum cosine distance in the embedding space of the pre-trained backbone between the downstream dataset and the \emph{mini}Imagenet~\cite{vinyals2016matching} dataset. We use \emph{mini}Imagenet ($60$k images) as a proxy of Imagenet-1k ($1.3$M images) to reduce the computational overhead of evaluating pairwise distances. This allows us to approximately measure the dissimilarity between the downstream dataset and Imagenet-1k.

\begin{figure}[ht]
\centering
\includegraphics[width = \linewidth]{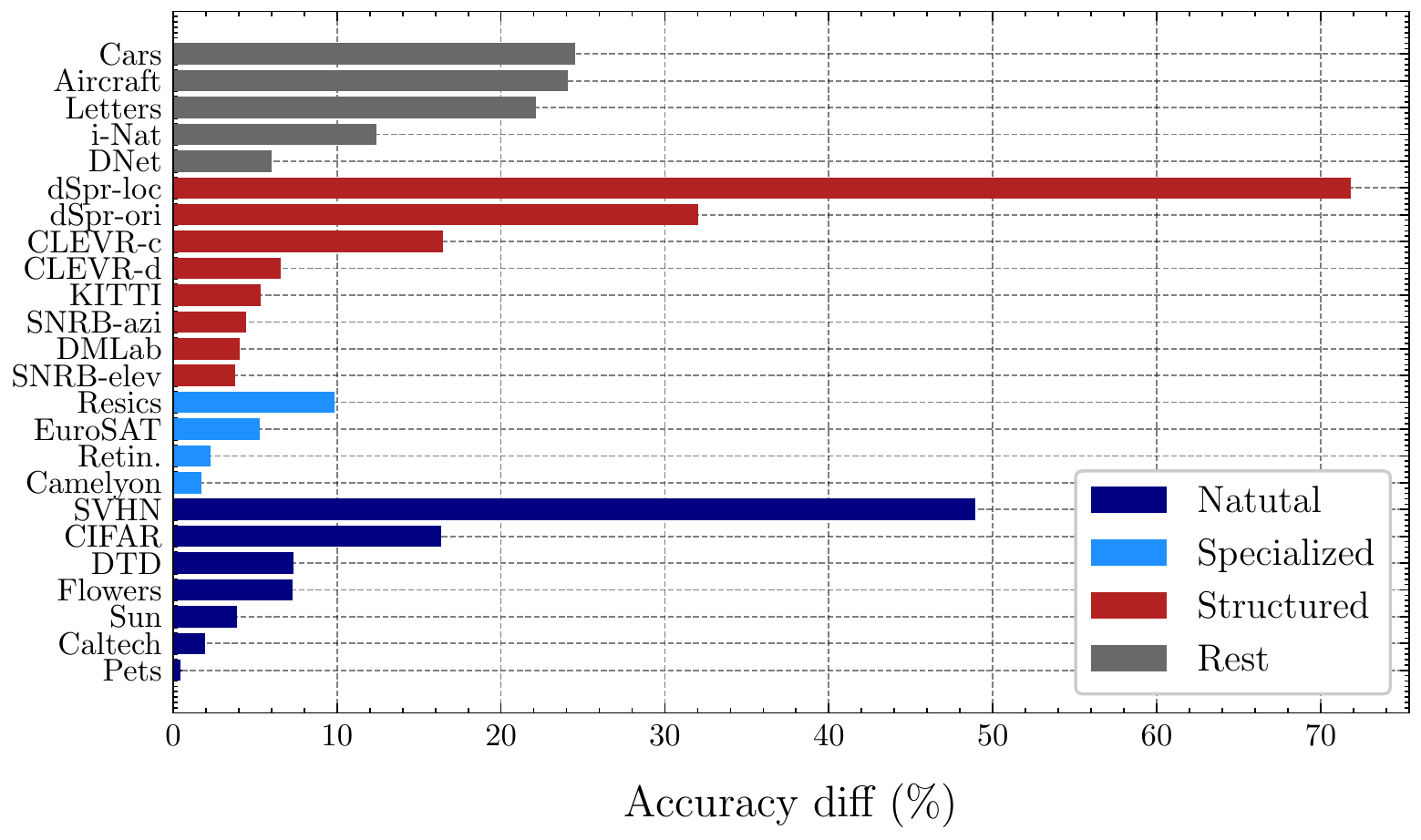} 
\caption{Bar plot of the accuracy differences between FSA-FiLM and NA for the offline case with 50 shots. %Notice that the benefits of body adaptation are more tangible compared to those in \cref{fig:ch_index} due to the larger number of data in this setting.
}
\label{fig:na_film_diff_50_shots_offline}
\end{figure}

%
%\vspace*{-.51cm}
%\vspace*{-.44cm}
%
%We first consider a challenging setting for FSA with only 5 classes and 50 shots per session. This results in sessions that are far smaller, on average than those considered in the few-shot CIL setting of \cref{sec:cil_comparisons}, making body adaptation non-trivial. We choose 12 datasets from VTAB+ and we run 12 few-shot CIL experiments. \cref{fig:ch_index} shows  the accuracy difference between FSA-FiLM and NA at the end of the last session as a function of the CH index. Only SVHN, FGVC-Aircraft, and Letters benefit significantly from body adaptation in this setting. If we use a threshold of 10 for the CH-index and use body adaptation below this number, but not above, we improve over adapting everywhere by  $2.4\%$ and improve over no-adaptation everywhere by $+7.3\%$.
%\vspace*{-.18cm}
\cref{fig:ch_index} shows  the accuracy difference between FSA-FiLM and NA as a function of the cosine distance for the offline setting with 50 shots whilst \cref{fig:na_film_diff_50_shots_offline} illustrates the same accuracy difference where the datasets are grouped based on the VTAB+ categorization. For this experiment, we use 24 datasets (VTAB+, DomainNet, and iNaturalist). We observe that adaptation is more beneficial for datasets in the structured domain, which contain images that are dissimilar to those of ImageNet-1k.

%\iffalse
\begin{table}[ht]
%\vspace*{-.3 cm}
%\begingroup
\setlength{\tabcolsep}{1.1pt} % Default value: 6pt
\begin{center}
\begin{footnotesize}
%\begin{sc}
\begin{tabular}{l c c c c c c c}
\toprule
  & \textbf{CFR100} & \textbf{SVHN} &  \textbf{dSp-loc} &  \textbf{FGVC} & \textbf{Letters}  & \textbf{DNet} & \textbf{iNat} \\
 \midrule
EffNet-B0 \tiny{(FSA-FiLM)} & 70.9 & \textbf{51.3} & \textbf{35.7} &\textbf{55.8} & \textbf{73.4} & 74.0 & 58.8 \\
ConvNext \tiny{(NA)} & \textbf{87.1} & 43.6 & 13.6 & 50.2 & 63.1 & \textbf{82.9} & \textbf{72.6} \\
\midrule
\midrule
ConvNext-CosD & 0.2 & 0.6 & 0.6 & 0.3 & 0.4 & 0.3 & 0.1 \\
\bottomrule
\end{tabular}
%\end{sc}
\end{footnotesize}
\end{center}
\caption{Accuracy comparison between a pre-trained ConvNext on ImageNet-22k without body adaptation (NA) and a pre-trained  EfficientNet-B0 on ImageNet-1k which is adapted on the first session (FSA-FiLM). We use the few-shot CIL setting  (\cref{sec:cil_comparisons}) and we report the accuracy (\%) ($\uparrow$) after the last session. Results are averaged over 5 runs (mean±std). We also include the minimum cosine distance using the pre-trained ConvNext as in \Cref{fig:ch_index}.}
\label{table:convnext_vs_film}
%\endgroup
\end{table}
%\fi

%\vspace*{-.14cm}
To stress the importance of adaptation on datasets far from ImageNet, we compare FSA-FiLM with an EfficientNet-B0 backbone to the no-adaptation method with a ConvNext \cite{liu2022convnet} pre-trained on Imagenet-22k. The total number of parameters for FSA-FiLM (backbone and FiLM parameters) is $\sim$4M while for ConvNext is 348M.  \Cref{table:convnext_vs_film} shows that a small adapted backbone can significantly surpass the accuracy of a much larger pre-trained backbone for datasets far from ImageNet. %The conclusion from these results also coincides with that of \cref{fig:ch_index}. 
We also compute the cosine distance measure for ConvNext and find that it is predictive of the performance of the unadapted ConvNext model and the benefits of adaptation. We show this in \Cref{table:convnext_vs_film}; note that for CIFAR100 the ConvNext's cosine distance values are small relative to the other datasets (indicating that CIFAR100 is close to the pretraining distribution) whereas for EfficientNet the values in \Cref{fig:ch_index} indicate that CIFAR100 is more out-of-distribution.

The main takeaway from these results is that when there is a large cosine distance (e.g.~SVHN, dSprites, FGVC, Letters) FiLM adaptation of a light-weight backbone performs well -- better even than applying LDA head adaptation to a much larger backbone trained on a much larger dataset.
As the community employs ever-larger models and datasets, these results indicate that adapters are likely to continue to bring improvements over simply learning a classifier head (the NA baseline).

\section{Discussion}\label{sec:discussion}
We have presented FSA (-FiLM), a simple yet effective replay-free baseline for CIL which adapts a pre-trained backbone via full body adaptation or FiLM layers only at the first session of continual learning and then utilizes a flexible incrementally updated LDA classifier on top of the body. Extensive experiments in several CIL scenarios have shown
that FSA outperforms previous SOTA baselines in most cases, thus, questioning the efficacy of current approaches to continuous body adaption. Furthermore, the experiments have revealed that FiLM layers are helpful when the amount of data available in the target domain is relatively low and the number of parameters in the neural network and the size of the source domain data set is large. For very large foundation models trained on very large source domain data sets, it is likely we are often in this situation and FiLM-like layers will be more effective than full-body adaptation, even when the target domain has a fairly large amount of data.

%\vspace*{-.05cm}
\noindent
\textbf{Limitations and future work.} The main limitation of this work is inextricably linked with the distributional assumptions of CIL in general. If the data distribution at each session shifts considerably (e.g.~the first session includes natural images while the next session includes images of numbers) then first session body adaptation is not suitable. In future work, we plan to combine FSA (-FiLM) with a memory method like GDumb to get the best of both worlds and deal with more challenging CIL scenarios.

\section*{Acknowledgements}
We are grateful to John Bronskill for his guidance on how to use the HPC cluster. We also thank John Bronskill, Aliaksandra Shysheya, and Massimiliano Patacchiola for useful discussions. The experiments were carried out using resources provided by the Cambridge Tier-2 system operated by the University of Cambridge Research Computing Service \url{https://www.hpc.cam.ac.uk}. This work was funded by Toyota Motor Europe.

%For very large foundation models trained on very large source domain data sets, it is likely we are often in this situation and adapters will be more effective than full-body adaptation, even when the target domain has a fairly large amount of data.  

%Extensions: combine first session adaptation with a memory method like GDumb to get the best of both worlds

%%%%%%%%% REFERENCES
{\small
\bibliographystyle{ieee_fullname}
\bibliography{refs}
}

\clearpage
\appendix

\section{Datasets}\label{sec:dataset_info}
The exact numbers of training samples and classes for each dataset used in the experiments of Section 4 in the main paper are given in \Cref{table:all_datasets_info}. For datasets with more than 120k training instances in VTAB+, due to hardware limitations, we randomly sample 120k images and the associated labels and we consider this subset as the full training set. For instance, when we use dSprites-location dataset in a 50-shot setting, we first randomly sample 120k examples and then we randomly pick 50 images for each one of the 16 classes. For DomainNet and iNaturalist we apply a different procedure (see \Cref{sec:splits} for details).

For evaluation, we consider the protocol used in \cite{zhai2019large} on the 19 datasets in VTAB, where a balanced dataset of $2$k images is created by randomly sampling images from the full test dataset. For FGVC-Aircraft, Cars, and Letters the full test dataset is utilized. We also use the full test dataset for evaluation in the high-shot Class-Incremental Learning (CIL) setting.

% Vtab+ datasets info
\begin{table*}[ht]
%\begingroup
\setlength{\tabcolsep}{7.8pt} % Default value: 6pt
\begin{center}
\begin{normalsize}
%\begin{sc}
\begin{tabular}{l c c  c  c}
\toprule
\textbf{Datasets}       &   \textbf{\# Classes} & \textbf{\# Train instances} & \textbf{ALL} & \textbf{CIL}  \\
\midrule
Caltech101~\cite{fei2006one} & $102$ & $3,060$ & \cmark & \xmark \\
CIFAR100~\cite{krizhevsky2009learning} & $100$ & $50,000$  & \cmark & \cmark \\
Flowers102~\cite{nilsback2008automated} & $102$ & $1,020$  & \cmark & \xmark \\
Pets~\cite{parkhi2012cats}  & $37$ & $3,680$ & \cmark & \xmark  \\
Sun397~\cite{xiao2010sun} & $397$  & $76,127$  & \cmark & \xmark \\
SVHN~\cite{netzer2011reading} & $10$ & $73,257$  & \cmark & \cmark  \\
DTD~\cite{cimpoi2014describing} & $47$ & $1,880$  & \cmark & \xmark  \\
\midrule
EuroSAT~\cite{helber2019eurosat} & $10$  & $27,000$ & \cmark & \xmark \\
Resics45~\cite{cheng2017remote} & $45$ & $31,500$ & \cmark & \xmark \\
Patch Camelyon~\cite{veeling2018rotation} & $2$  & $262,144$ & \cmark & \xmark \\
Retinopathy~\cite{kaggle2015diabetic} & $5$ & $35,126$  & \cmark & \xmark \\
\midrule
CLEVR-count~\cite{johnson2017clevr} & $8$ & $70,000$ & \cmark & \xmark \\
CLEVR-dist~\cite{johnson2017clevr} & $6$  & $70,000$ & \cmark & \xmark \\
dSprites-loc~\cite{dsprites17} & $16$ & $737,280$  & \cmark & \cmark \\
dSprites-ori~\cite{dsprites17} & $16$  & $737,280$ & \cmark & \xmark\\
SmallNORB-azi~\cite{lecun2004learning} & $18$ & $24,300$ & \cmark & \xmark  \\
SmallNORB-elev~\cite{lecun2004learning} & $9$ & $24,300$ & \cmark & \xmark  \\
DMLab~\cite{beattie2016deepmind} & $6$ & $65,550$  & \cmark & \xmark  \\
KITTI-dist~\cite{geiger2013vision} & $4$ & $6,347$  & \cmark & \xmark  \\
\midrule
FGVC-Aircraft~\cite{maji2013fine} & $100$ & $6,667$ & \cmark & \cmark   \\
Cars~\cite{krause20133d} & $196$ & $8,144$ & \cmark & \cmark  \\
Letters~\cite{deCampos2009character} & $62$ & $74,107$ & \cmark & \cmark  \\
\midrule
\midrule
DomainNet~\cite{peng2019moment} &  $60$ ($345^\dagger$) & $569,010$ & \xmark & \cmark   \\
iNaturalist~\cite{van2018inaturalist} & $100$ ($10,000^\dagger$)  & $500,000$ & \xmark & \cmark  \\
Core50~\cite{lomonaco2017core50} & $50$ & $119,894$ & \cmark & \cmark \\
CUB200~\cite{WahCUB_200_2011} & $200$ & $11,788$ & \xmark & \cmark  \\
\bottomrule
\end{tabular}
%\end{sc}
\end{normalsize}
\end{center}
\caption{Information concerning all datasets used in the experiments. $\dagger$ denotes the number of classes of the original dataset before they are modified for the continual learning scenarios (see \Cref{sec:splits} for more details). The first 22 datasets form the VTAB+ collection. We also indicate whether a dataset has been used in the offline experiments in Section 4.2 of the main paper which use all the available training data (ALL). Similarly, we indicate which datasets are considered for the Class-Incremental Learning settings in Section 4.}
\label{table:all_datasets_info}
%\endgroup
\end{table*}

%---------------------------------------------

\section{Dataset Information for the Class-Incremental Learning Experiments}\label{sec:splits}
% 3 tables; one for each CIL setting
In \Cref{table:cil_datasets_splits_info}, we present detailed information about the Class-Incremental Learning Experiments (CIL) experiments that are in Section 4.3 in the main paper, such as the number of total sessions, number of train instances, and classes per session. Next, we discuss the exact setup for DomainNet and iNaturalist.

\begin{table}[ht]

%\begingroup
\setlength{\tabcolsep}{2.2pt} % Default value: 6pt
\begin{center}
    
\begin{small}
%\begin{sc}
\begin{tabular}{l c c c c c c c}
\toprule
\textbf{CIL setting}  & \textbf{Datasets}  &   \textbf{$S$} & \textbf{$N_1$} & \textbf{$|\cY_1|$} & \textbf{$N_s$} & \textbf{$|\cY_s|$} \\
\midrule
 \multirow{4}{*}{High-shot} & CIFAR100 & $10$ & $5$k & $10$ & $5$k & $10$ \\
 & SVHN   & $5$ & $\sim 19$k & $2$ & $\sim 14$k & $2$  \\
 & dSprites-loc  & $7$ & $24$k & $4$ & $12$k & $2$  \\
 & FGVC-Aircraft & $10$ & $667$ & $10$ & $\sim 670$ & $10$  \\
 & Cars & $10$ & $652$ & $15$ & $\sim 830$ & $20$  \\
 & Letters & $11$ & $\sim 11$k & $12$ & $\sim 5$k & $5$  \\
 & Core50  & $9$ & $\sim 24$k & $10$ & $\sim 12$k & $5$  \\
\midrule
 \multirow{2}{*}{Few-shot+} & CIFAR100  & $9$ & $30$k & $60$ & $25$ & $5$  \\
 & CUB200   & $11$ & $3$k & $100$ & $50$ & $10$  \\
\midrule
 \multirow{7}{*}{Few-shot} & CIFAR100  & $9$ & $1$k & $20$ & $500$ & $10$  \\
 & SVHN   & $5$ & $100$ & $2$ & $100$ & $2$  \\
 & dSprites-loc  & $7$ & $200$ & $4$ & $100$ & $2$  \\
 & FGVC-Aircraft & $9$ & $1$k & $20$ & $500$ & $10$  \\
 & Cars & $9$ & $1484$ & $36$ & $\sim 830$ & $20$  \\
 & Letters & $11$ & $600$ & $12$ & $250$ & $5$  \\
 & DomainNet & $9$ & $600$ & $12$ & $300$ & $6$  \\
 & iNaturalist & $9$ & $1$k & $20$ & $500$ & $10$  \\
\bottomrule
\end{tabular}
%\end{sc}
\end{small}
\end{center}
\caption{Detailed CIL settings for the experiments of Section 4.3. We report the total number of sessions ($S$), the number of train instances ($N_1$), and the number of classes ($|\cY_1|$) of the first session and the rest of the sessions ($N_s, |\cY_s|, s>1$).}
\label{table:cil_datasets_splits_info}
%\endgroup
\end{table}

\paragraph{DomainNet \& iNaturalist.} DomainNet and iNaturalist are the only datasets for which we follow a different pre-processing procedure from the one described in \Cref{sec:dataset_info} in order to create a few-shot CIL scenario similar to the ones considered in the literature. This is due to the large number of classes (iNaturalist has $10,000$ classes) and different domains (DomainNet includes images from 6 domains) these datasets have.

\begin{table*}[ht]
\begin{center}
    
\setlength{\tabcolsep}{2.1pt} % Default value: 6pt
\begin{small}
%\begin{sc}
%\begin{tabular}{l c c c c c c c c c c c c}
\begin{tabular}{l c c c c c c}
\toprule
%\textbf{Domain}  & \textbf{Superclass}  &  \multicolumn{10}{c}{\textbf{Classes}}  \\
%\midrule  
%Clipart & Furniture & Couch (1) & Fence (2) & Streetlight (3) & Table (4) & Toothbrush (5) & Vase (6) & Bed (7) & Fireplace (8) & Teapot (9) & Lantern (10) \\
%Infograph & Mammal & Cat (11) & Dolphin (12) & Squirrel (13) & Zebra (14) & Cow (15) & Elephant (16) & Pig (17) & Tiger (18) & Dog (19) & Rabbit (20) \\
%Painting & Tool & Anvil (21) & Basket (22) & Rifle (23) & Axe (24) & Dumbbell (25) & Pliers (26) & Saw (27) & Skateboard (28) & Bandage (29) & Paint Can (30) \\
%Quickdraw & Cloth & Belt (31) & Camouflage (32) & Eyeglasses (33) & Helmet (34) & Necklace (35) & Rollerskates (36) & Sock (37) & Underwear (38) & Bowtie (39) & Crown (40) \\
%Real & Electricity & Calculator (41) & Computer (42) & Fan (43) & Oven (44) & Dishwasher (45) & %Headphones (46) & Microwave (47) & Radio (48) & Stereo (49) & Toaster (50) \\
%Sketch & Road Transportation & Ambulance (51) & Bus (52) & Motorbike (53) & Train (54) & Bicycle (55) & Car (56) & Truck (57) & Bulldozer (58) & Firetruck (59) & Tractor (60) \\
\textbf{Domain} & Clipart & Infograph & Painting & Quickdraw & Real & Sketch \\
\midrule
\textbf{Superclass} & Furniture & Mammal & Tool & Cloth & Electricity &  Road Transportation \\
\midrule
\multirow{10}{*}{\textbf{Classes}} & Clipart & Infograph & Painting & Quickdraw & Real & Sketch \\
 & Furniture & Mammal & Tool & Cloth & Electricity & Road Transportation \\
 & Couch (1) & Cat (11) & Anvil (21) & Belt (31) & Calculator (41) & Ambulance (51) \\
 & Fence (2) & Dolphin (12) & Basket (22) & Camouflage (32) & Computer (42) & Bus (52) \\
 & Streetlight (3) & Squirrel (13) & Rifle (23) & Eyeglasses (33) & Fan (43) & Motorbike (53) \\
 & Table (4) & Zebra (14) & Axe (24) & Helmet (34) & Oven (44) & Train (54) \\
 & Toothbrush (5) & Cow (15) & Dumbbell (25) & Necklace (35) & Dishwasher (45) & Bicycle (55) \\
 & Vase (6) & Elephant (16) & Pliers (26) & Rollerskates (36) & Headphones (46) & Car (56) \\
 & Bed (7) & Pig (17) & Saw (27) & Sock (37) & Microwave (47) & Truck (57) \\
 & Fireplace (8) & Tiger (18) & Skateboard (28) & Underwear (38) & Radio (48) & Bulldozer (58) \\
 & Teapot (9) & Dog (19) & Bandage (29) & Bowtie (39) & Stereo (49) & Firetruck (59) \\
 & Lantern (10) & Rabbit (20) & Paint Can (30) & Crown (40) & Toaster (50) & Tractor (60) \\
\bottomrule
\end{tabular}
%\end{sc}
\end{small}
\end{center}
\caption{DomainNet classes (class id in parentheses) used for the few-shot CIL setting.}
\label{table:domainnet_classes}
%\endgroup
\end{table*}

DomainNet is a large-scale dataset of $\sim 0.6$M images lying in 6 different domains (clipart, infograph, painting, quickdraw, real, sketch) and categorized into 365 distinct classes. These classes can be grouped into 24 superclasses: furniture, mammal, tool, cloth, electricity, building, office, human
body, road transportation, food, nature, cold-blooded,
music, fruit, sport, tree, bird, vegetable, shape, kitchen,
Water transportation, sky transportation, insects, and others. In our CIL experiments, we use 60 classes from the superclasses with an adequate number of instances ($>150$): furniture, mammal, tool, cloth, electricity, and road transportation. To the best of our knowledge, this is the first time such a dataset is considered for CIL problems. \Cref{table:domainnet_classes} summarizes the DomainNet classes we use for the CIL experiments. To build the 50-shot CIL setting of Section 4.3, we randomly sample 50 images per class and the rest of the images are used for evaluation. 

The iNaturalist is another large-scale dataset. comprising $\sim 2.7$ million images of $10,000$ species. The species can be divided into 10 general categories: amphibians, animalia, arachnids, birds, fungi, insects, mammals, mollusks, plants, and reptiles. Due to the dataset's large size, we have opted to use the ``mini'' version of the training dataset\footnote{We use the data from the 2021 competition, available at \url{https://github.com/visipedia/inat_comp/tree/master/2021}.} which has 50 images per class, and thus, this is the only dataset from the few-shot CIL experiments that we do not repeat for 5 times since the (mini) train dataset is already in a 50-shot setting. For evaluation, we use the validation data with 10 images per class. The number of classes considered for the CIL experiments is reduced from $10,000$ to 100; 10 classes per superclass (10 superclasses/sessions). Specific details are given in Tables   \ref{table:inaturalist_classes_1} and \ref{table:inaturalist_classes_2}.

\begin{table*}[ht]

%\begingroup
\setlength{\tabcolsep}{8.pt} % Default value: 6pt
\begin{center}
    
\begin{tiny}
%\begin{sc}
\begin{tabular}{l c c c c c c}
\toprule
  &  \multicolumn{5}{c}{\textbf{Superclass}}  \\
\cmidrule(lr){2-7} 
   & \textbf{Amphibians} & \textbf{Animalia} & \textbf{Arachnids} & \textbf{Birds} & \textbf{Fungi} \\
\textbf{Session} & \textbf{1} & \textbf{2} & \textbf{3} & \textbf{4} & \textbf{5} \\
\midrule  
 \multirow{10}{*}{Classes} & Ascaphus truei & Lumbricus terrestris & Eratigena duellica & Accipiter badius & Herpothallon rubrocinctum \\
 & Bombina orientalis & Sabella spallanzanii & Atypoides riversi & Accipiter cooperii & Chrysothrix candelaris \\
 & Bombina variegata & Serpula columbiana & Aculepeira ceropegia & Accipiter gentilis & Apiosporina morbosa \\
 & Anaxyrus americanus & Spirobranchus cariniferus & Agalenatea redii & Accipiter nisus & Acarospora socialis \\
 & Anaxyrus boreas & Hemiscolopendra marginata & Araneus bicentenarius & Accipiter striatus & Physcia adscendens \\
 & Anaxyrus cognatus & Scolopendra cingulata & Araneus diadematus & Accipiter trivirgatus & Physcia aipolia \\
 & Anaxyrus fowleri & Scolopendra heros & Araneus marmoreus & Aegypius monachus & Physcia millegrana \\
 & Anaxyrus punctatus & Scolopendra polymorpha & Araneus quadratus & Aquila audax & Physcia stellaris \\
 & Anaxyrus quercicus & Scutigera coleoptrata & Araneus trifolium & Aquila chrysaetos & Candelaria concolor \\
 & Anaxyrus speciosus & Ommatoiulus moreleti & Araniella displicata & Aquila heliaca & Cladonia chlorophaea \\
\bottomrule
\end{tabular}
%\end{sc}
\end{tiny}
\end{center}
\caption{Classes used from iNaturalist to create the few-shot CIL setting. (table continues to \Cref{table:inaturalist_classes_2}).}
\label{table:inaturalist_classes_1}
%\endgroup
\end{table*}

\begin{table*}[ht]

%\begingroup
\setlength{\tabcolsep}{7.pt} % Default value: 6pt
\begin{center}
    
\begin{tiny}
%\begin{sc}
\begin{tabular}{l c c c c c c}
\toprule
  &  \multicolumn{5}{c}{\textbf{Superclass}}  \\
\cmidrule(lr){2-7} 
   & \textbf{Insects} & \textbf{Mammals} & \textbf{Mollusks} & \textbf{Plants} & \textbf{Reptiles} \\
\textbf{Session} & \textbf{6} & \textbf{7} & \textbf{8} & \textbf{9} & \textbf{10} \\
\midrule  
 \multirow{10}{*}{Classes} & Aptera fusca & Antilocapra americana & Ensis leei & Bryum argenteum & Alligator mississippiensis \\
 & Panchlora nivea & Balaenoptera acutorostrata & Clinocardium nuttallii & Rhodobryum ontariense & Caiman crocodilus \\
 & Pycnoscelus surinamensis & Megaptera novaeangliae & Dinocardium robustum & Leucolepis acanthoneura & Crocodylus acutus \\
 & Blatta orientalis & Aepyceros melampus & Tridacna maxima & Plagiomnium cuspidatum & Crocodylus moreletii \\
 & Periplaneta americana & Alcelaphus buselaphus & Donax gouldii & Plagiomnium insigne & Crocodylus niloticus \\
 & Periplaneta australasiae & Antidorcas marsupialis & Donax variabilis & Rhizomnium glabrescens & Crocodylus porosus \\
 & Periplaneta fuliginosa & Bison bison & Dreissena polymorpha & Dicranum scoparium & Sphenodon punctatus \\
 & Pseudomops septentrionalis & Bos taurus & Mya arenaria & Ceratodon purpureus & Acanthocercus atricollis \\
 & Arrhenodes minutus & Boselaphus tragocamelus & Cyrtopleura costata & Leucobryum glaucum & Agama atra \\
 & Agrilus planipennis & Bubalus bubalis & Geukensia demissa & Funaria hygrometrica & Agama picticauda \\
\bottomrule
\end{tabular}
%\end{sc}
\end{tiny}
\end{center}
\caption{Classes used from iNaturalist to create the few-shot CIL setting. The first part can be found in \Cref{table:inaturalist_classes_1}.}
\label{table:inaturalist_classes_2}
%\endgroup
\end{table*}

%----------------------------------------------

\section{Extra Training Details}
Due to the large number of experiments and datasets we tried to keep the hyperparameter tuning to a minimum by choosing a set of hyperparameters that works fairly well across all datasets and settings. We have not used any data augmentation in our experiments and all images have been scaled to $224 \times 224$ pixels. The only exception is the experiments on CIFAR100, and CUB200 under the few-shot+ setting. There, for comparability reasons, we followed the exact experimental settings as in \cite{zhou2022forward} where standard data augmentation techniques (e.g.~random flips and crops) were utilized. Moreover, when we used ResNet-20 for CIFAR100 we maintained the original image size ($32 \times 32$).

\paragraph{Computing Infrastructure Details \& Code.} All the experiments of Section 4 have been carried out on a Linux machine with a single NVIDIA-A100 (80GB memory) GPU. Our PyTorch-based code will be made available via a public repository after the review period. 

\paragraph{Optimization Details.} 
In all experiments, we train the models using a batch size of 256. Apart from GDumb, for the rest of the methods, EfficientNet-B0 backbones are optimized with the Adam optimizer \cite{kingma2014adam} while for ResNet architectures we opt for SGD with momentum set to 0.9. For GDumb, we follow \cite{prabhu2020gdumb} and we use SGD with momentum. For FACT~\cite{zhou2022forward} and FSA with pre-trained EfficientNet-B0 backbone, we set the initial learning rate to 0.0001 with scheduled decays by a factor of 0.5 every 50 epochs while for FSA-FiLM, we set it to 0.005. We train all full-body adaptation methods for 200 epochs and the FSA-FiLM for 150 epochs (except for the high-shot setting where we use 200 epochs for fair time comparisons). For the few-shot+ CIL scenario, we follow the training setup of~\cite{zhou2022forward}. The weights of the pre-trained EfficientNet-B0 have been obtained from \url{https://github.com/lukemelas/EfficientNet-PyTorch} while for the pre-trained weights of ResNet-18 and ConvNext, we use the following repository  \url{https://github.com/rwightman/pytorch-image-models}.

\paragraph{Competitors.} We found empirically that the recommended hyperparameter values (learning rates, cutmix parameters, SGDR schedule) for GDumb in~\cite{prabhu2020gdumb} work well in practice and we use these throughout the experiments. Similarly, for FACT, we use the default values $\alpha=0.5, \gamma=0.01, V = \text{number of new classes in total}$~\cite{zhou2022forward}. For ALICE, following~\cite{peng2022few}, the projection head is a two-layer MLP with a hidden feature size of
2048 and ReLU as the activation function. All the other hyperparameters (scale factor $s$, margin $m$, etc.) are set as in~\cite{peng2022few}. For E-EWC+SDC, a triplet loss~\cite{hoffer2015deep} is used as in~\cite{yu2020semantic} and the final embeddings of 640 dimensions are normalized.

%----------------------------------------------

\section{Additional Results}
In this section, we provide tables with the exact accuracies for each one of the datasets used in the experiments under different settings. We have run extra experiments on VTAB+ using meta-learned FiLM adapters in the offline setting and we report accuracies. Additionally, we perform a comparison between different backbones in the offline setting:  EfficientNet-B0 and ResNet-18. For the high-shot setting, apart from the four datasets utilized in the main paper, we also deploy the methods on SVHN and present accuracies by session. Finally, accuracies at each session for all three CIL settings are provided.

\paragraph{Head Comparison.} Here we provide the exact accuracies for each dataset based on Section 4.2 and Figure 1 of the main paper. Tables \ref{table:head_comparison_na_5shots}, \ref{table:head_comparison_na_10shots}, \ref{table:head_comparison_na_50shots}, and \ref{table:head_comparison_na_alldata} give the offline accuracies for the no adaptation (NA) method for 5, 10, 50 shots, and all training data, respectively. Similar information for the FiLM adaptation method (A-FiLM) is given in Tables \ref{table:head_comparison_film_5shots}, \ref{table:head_comparison_film_10shots}, \ref{table:head_comparison_film_50shots}, and \ref{table:head_comparison_film_alldata}. Finally, Tables \ref{table:head_comparison_full_5shots}, \ref{table:head_comparison_full_10shots}, \ref{table:head_comparison_full_50shots}, and \ref{table:head_comparison_full_alldata} provide the corresponding accuracies for the full-body adaptation method (A-FB).

\paragraph{Meta-learned FiLM Adapters.} We consider experiments in the offline setting with meta-learned FiLM adapters. We use the meta-trained FiLM adapters as presented in \cite{bronskill2021memory}. The results for meta-learned FiLM adapters, as well as for no-adaptation (NA), FiLM (fine-tuned) adaptation (FiLM), and full body adaptation methods, are summarized in Tables \ref{table:meta_effenet_5shots}, \ref{table:meta_effenet_10shots}, and \ref{table:meta_effenet_50shots}, for 5, 10, and 50 shots, respectively. We observe that the meta-trained FiLM adapters work better than NA in all cases, but they fail to compete with the fine-tuned FiLM adapters. Notice that as the number of shots increases, the accuracy difference between meta-learned and fine-tuned FiLM adapters also increases.

\paragraph{FiLM Adaptation: EfficientNet-B0 vs ResNet-18.} To assess how different backbone architectures affect the performance of the no-adaptation and FiLM adaptation method, we compare ResNet-18 and EfficientNet-B0 (EN) backbones in Tables \ref{table:resnet18_vs_effnet_film_5shots}, \ref{table:resnet18_vs_effnet_film_10shots}, and \ref{table:resnet18_vs_effnet_film_50shots}, for 5, 10, and 50 shots, respectively. All tables demonstrate the superiority of EfficientNet-B0, regardless of the adaptation method. The tables also show that, regardless of backbone architecture and number of shots, FiLM adaptation provides significant performance benefits. 

\paragraph{High-shot CIL: Accuracies per Session.} We provide detailed accuracies for each incremental session for all baselines in the high-shot CIL setting. The accuracies for CIFAR100, CORE50, SVHN, dSPrites-loc, FGVC-Aircraft, Cars, and Letters can be found in Tables \ref{table:high_shot_cifar100}, \ref{table:high_shot_core50}, \ref{table:high_shot_svhn}, \ref{table:high_shot_dspritesloc}, \ref{table:high_shot_fgvcaircraft}, \ref{table:high_shot_cars}, and
\ref{table:high_shot_letters}, respectively. For GDumb, we provide results with a memory buffer of size 1k and 5k.

\begin{figure*}[htb]
\centering
\includegraphics[width = \textwidth]{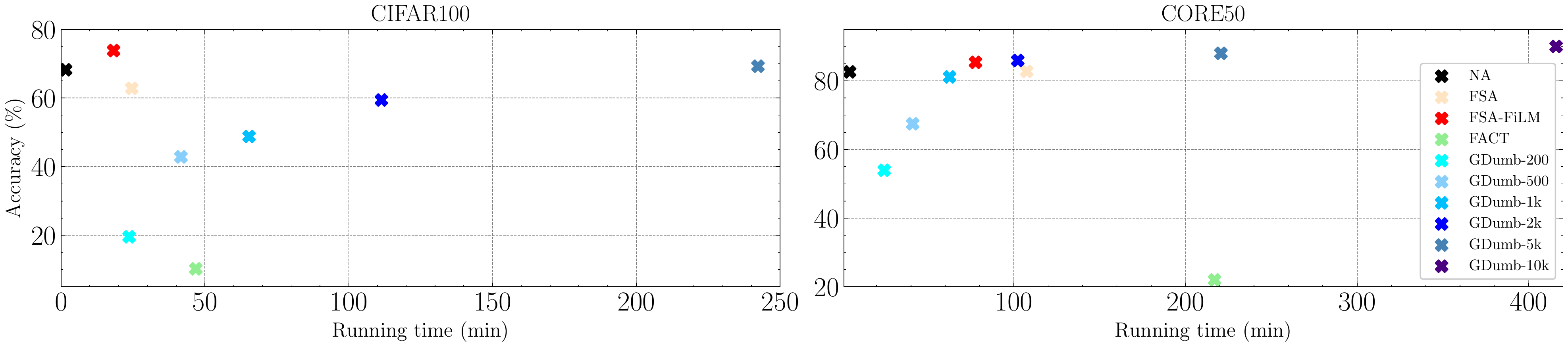} 
\caption{Last session's test accuracy ($\uparrow$) and run time ($\downarrow$)  for the ``high-shot CIL'' setting of \cref{sec:cil_comparisons}. GDumb-$m$ refers to memory buffer sizes $m \in \{ 200, 500, 1\text{k}, 2\text{k}, 5\text{k}, 10\text{k}^*  \}$. We use a memory buffer of 10k images only for CORE50.}
\label{fig:gdumb_cifar100_core50_time_acc}
\end{figure*}

\paragraph{Few-shot+ CIL: Accuracies per Session.}
We provide detailed accuracy for each incremental session for all baselines in the few-shot+ CIL setting. The accuracies for CIFAR100 and CUB200 can be found in Tables \ref{table:few_shot_plus_cifar100} and
\ref{table:few_shot_plus_cub200}, respectively. 

\paragraph{Few-shot CIL: Accuracies per Session.}
We provide detailed accuracy (+ error bars) for each incremental session for all baselines in the few-shot CIL setting. The accuracies for CIFAR100, SVHN, dSprites-location, FGVC-Aircraft, Letters, DomainNet, and iNaturalist can be found in Tables \ref{table:few_shot_cifar100}, \ref{table:few_shot_svhn}, \ref{table:few_shot_dsprites}, \ref{table:few_shot_fgvcaircraft}, \ref{table:few_shot_letters}, \ref{table:few_shot_domainnet}, and
\ref{table:few_shot_inaturalist}, respectively.

\paragraph{FSA-FiLM vs GDumb} The trade-off between accuracy and training time for different continual learning methods on CIFAR100 and CORE50 is illustrated in Fig.~\ref{fig:gdumb_cifar100_core50_time_acc}. Several different memory sizes are used for GDumb. FSA-FiLM attains the highest accuracy (and the lowest PPDR) $\approx$13.5x faster than GDumb with a 5k memory buffer on CIFAR100 while on CORE50, GDumb requires at least a 5K memory buffer to outperform FSA-FiLM and $\approx$3x more training time than FSA-FiLM. Notice that FACT is unable to perform well under this setting due to the small number of available classes in the first
session.
%%%%%%%%%%%%%%%%%%%%%%%%%%%%%%%%%
% Tables for head comparisons %%%
%%%%%%%%%%%%%%%%%%%%%%%%%%%%%%%%%

%%%%%%%%
% NA %%%
%%%%%%%%

\begin{table}[ht]
%\vspace*{1.5 cm}
%\begingroup

\setlength{\tabcolsep}{6.5pt} % Default value: 6pt
\begin{center}
    
\begin{small}
%\begin{sc}
\begin{tabular}{l c c c}
\toprule
 \textbf{Dataset} & \textbf{NCM} & \textbf{LDA} &  \textbf{Linear} \\
 \midrule
Caltech101 & 85.7 {\tiny $\pm 0.9$} & \textbf{88.2} {\tiny $\pm \textbf{0.8}$} & 87.2 {\tiny $\pm 0.7$} \\
CIFAR100 & 40.2 {\tiny $\pm 1.5$} & \textbf{42.7} {\tiny $\pm \textbf{1.6}$} & 42.3 {\tiny $\pm 1.3$} \\
Flowers102 & 71.5 {\tiny $\pm 0.6$} & \textbf{76.1} {\tiny $\pm \textbf{0.4}$} & 75.7 {\tiny $\pm 0.8$} \\
Pets & \textbf{83.0} {\tiny $\pm \textbf{1.5}$} & 82.4 {\tiny $\pm 1.6$} & 82.5 {\tiny $\pm 1.5$} \\
Sun397 & 41.0 {\tiny $\pm 0.9$} & 41.9 {\tiny $\pm 1.0$} & \textbf{42.9} {\tiny $\pm \textbf{0.7}$} \\
SVHN & 13.5 {\tiny $\pm 1.6$} & \textbf{16.5} {\tiny $\pm \textbf{1.1}$} & 15.0 {\tiny $\pm 1.4$} \\
DTD & 48.2 {\tiny $\pm 1.0$} & 48.9 {\tiny $\pm 1.7$} & \textbf{49.2} {\tiny $\pm \textbf{1.8}$} \\
\midrule
EuroSAT & 72.1 {\tiny $\pm 1.9$} & \textbf{76.3} {\tiny $\pm \textbf{1.8}$} & 74.6 {\tiny $\pm 2.0$} \\
Resics45 & 55.2 {\tiny $\pm 1.2$} & 58.8 {\tiny $\pm 1.4$} & \textbf{58.9} {\tiny $\pm \textbf{1.2}$} \\
\midrule
Patch Camelyon & 59.6 {\tiny $\pm 8.7$} & \textbf{59.8} {\tiny $\pm \textbf{7.2}$} & 59.0 {\tiny $\pm 6.4$} \\
Retinopathy & \textbf{26.3} {\tiny $\pm \textbf{2.5}$} & 25.6 {\tiny $\pm 1.6$} & 24.3 {\tiny $\pm 1.7$} \\
\midrule
CLEVR-count & 22.2 {\tiny $\pm 0.8$} & \textbf{23.1} {\tiny $\pm \textbf{1.1}$} & 22.6 {\tiny $\pm 0.7$} \\
CLEVR-dist & 23.0 {\tiny $\pm 2.7$} & \textbf{24.5} {\tiny $\pm \textbf{2.3}$} & 24.4 {\tiny $\pm 2.2$} \\
dSprites-loc & 7.5 {\tiny $\pm 0.7$} & \textbf{8.5} {\tiny $\pm \textbf{0.6}$} & 7.3 {\tiny $\pm 0.6$} \\
dSprites-ori & 13.1 {\tiny $\pm 1.2$} & \textbf{16.2} {\tiny $\pm \textbf{0.8}$} & 14.8 {\tiny $\pm 1.1$} \\
SmallNORB-azi & 6.8 {\tiny $\pm 0.6$} & \textbf{9.3} {\tiny $\pm \textbf{0.8}$} & 8.8 {\tiny $\pm 1.0$} \\
SmallNORB-elev & 13.0 {\tiny $\pm 1.5$} & \textbf{15.1} {\tiny $\pm \textbf{0.6}$} & 14.5 {\tiny $\pm 0.9$} \\
DMLab & 22.2 {\tiny $\pm 0.8$} & 22.1 {\tiny $\pm 1.3$} & \textbf{23.8} {\tiny $\pm \textbf{0.8}$} \\
KITTI-dist & 50.0 {\tiny $\pm 1.3$} & 51.4 {\tiny $\pm 2.7$} & \textbf{51.9} {\tiny $\pm \textbf{3.1}$} \\
\midrule
FGVC-Aircraft & 19.4 {\tiny $\pm 0.6$} & 22.1 {\tiny $\pm 1.0$} & \textbf{22.7} {\tiny $\pm \textbf{0.6}$} \\
Cars & 20.0 {\tiny $\pm 0.6$} & \textbf{22.6} {\tiny $\pm \textbf{0.6}$} & 22.0 {\tiny $\pm 0.7$} \\
Letters & 28.3 {\tiny $\pm 1.8$} & \textbf{36.1} {\tiny $\pm \textbf{2.1}$} & 34.0 {\tiny $\pm 1.3$} \\
\midrule
Average acc & 37.4 & \textbf{39.5} & 39.0 \\
\bottomrule
\end{tabular}
%\end{sc}
\end{small}
\end{center}
\caption{Accuracy comparison between NCM, LDA, and Linear head without using any adaptation (\textbf{NA} method). The reported results are based on \textbf{5} shots and averaged over 5 runs (mean±std). A pre-trained EfficientNet-B0 is used as a backbone in all cases.}
\label{table:head_comparison_na_5shots}
%\endgroup
\end{table}

\begin{table}[ht]
%\vspace*{1.5 cm}
%\begingroup

\setlength{\tabcolsep}{6.5pt} % Default value: 6pt %\renewcommand{\arraystretch}{.9} \begin{center}
\begin{center}
    
\begin{small}
%\begin{sc}
\begin{tabular}{l c c c}
\toprule
 \textbf{Dataset} & \textbf{NCM} & \textbf{LDA} &  \textbf{Linear} \\
 \midrule
Caltech101 & 88.5 {\tiny $\pm 0.9$} & \textbf{90.0} {\tiny $\pm \textbf{0.8}$} & 89.6 {\tiny $\pm 0.7$} \\
CIFAR100 & 45.8 {\tiny $\pm 1.0$} & \textbf{50.1} {\tiny $\pm \textbf{1.5}$} & 49.2 {\tiny $\pm 1.1$} \\
Flowers102 & 77.2 {\tiny $\pm 0.2$} & \textbf{83.9} {\tiny $\pm \textbf{0.3}$} & 81.9 {\tiny $\pm 0.2$} \\
Pets & 85.8 {\tiny $\pm 1.0$} & \textbf{86.4} {\tiny $\pm \textbf{0.7}$} & 86.1 {\tiny $\pm 0.5$} \\
Sun397 & 47.1 {\tiny $\pm 1.3$} & 49.0 {\tiny $\pm 1.3$} & \textbf{49.7} {\tiny $\pm \textbf{0.8}$} \\
SVHN & 15.5 {\tiny $\pm 2.3$} & \textbf{19.4} {\tiny $\pm \textbf{2.9}$} & 18.1 {\tiny $\pm 2.6$} \\
DTD & 53.8 {\tiny $\pm 0.3$} & 55.6 {\tiny $\pm 0.6$} & \textbf{56.3} {\tiny $\pm \textbf{0.7}$} \\
\midrule
EuroSAT & 76.6 {\tiny $\pm 1.2$} & \textbf{82.1} {\tiny $\pm \textbf{0.9}$} & 80.2 {\tiny $\pm 0.9$} \\
Resics45 & 60.7 {\tiny $\pm 1.3$} & 65.5 {\tiny $\pm 1.1$} & \textbf{65.9} {\tiny $\pm \textbf{1.2}$} \\
\midrule
Patch Camelyon & 63.0 {\tiny $\pm 5.3$} & \textbf{66.5} {\tiny $\pm \textbf{3.7}$} & 65.3 {\tiny $\pm 4.7$} \\
Retinopathy & \textbf{27.5} {\tiny $\pm \textbf{3.2}$} & 27.1 {\tiny $\pm 2.2$} & 26.1 {\tiny $\pm 1.8$} \\
\midrule
CLEVR-count & 24.0 {\tiny $\pm 0.4$} & \textbf{25.7} {\tiny $\pm \textbf{0.6}$} & 24.9 {\tiny $\pm 0.6$} \\
CLEVR-dist & 24.2 {\tiny $\pm 1.2$} & \textbf{26.3} {\tiny $\pm \textbf{1.1}$} & 26.3 {\tiny $\pm 1.4$} \\
dSprites-loc & 7.5 {\tiny $\pm 0.5$} & \textbf{8.7} {\tiny $\pm \textbf{0.3}$} & 7.9 {\tiny $\pm 0.3$} \\
dSprites-ori & 14.2 {\tiny $\pm 0.9$} & \textbf{18.2} {\tiny $\pm \textbf{0.7}$} & 16.4 {\tiny $\pm 1.0$} \\
SmallNORB-azi & 8.4 {\tiny $\pm 0.4$} & 9.5 {\tiny $\pm 1.1$} & \textbf{9.5} {\tiny $\pm \textbf{0.7}$} \\
SmallNORB-elev & 13.6 {\tiny $\pm 0.8$} & \textbf{16.5} {\tiny $\pm \textbf{1.1}$} & 16.2 {\tiny $\pm 0.7$} \\
DMLab & 25.1 {\tiny $\pm 1.1$} & 25.7 {\tiny $\pm 1.2$} & \textbf{27.2} {\tiny $\pm \textbf{1.3}$} \\
KITTI-dist & 50.1 {\tiny $\pm 0.8$} & \textbf{52.9} {\tiny $\pm \textbf{1.5}$} & 52.0 {\tiny $\pm 1.8$} \\
\midrule
FGVC-Aircraft & 23.1 {\tiny $\pm 0.4$} & \textbf{28.5} {\tiny $\pm \textbf{0.4}$} & 27.0 {\tiny $\pm 0.4$} \\
Cars & 25.4 {\tiny $\pm 0.5$} & 30.4 {\tiny $\pm 0.5$} & \textbf{30.7} {\tiny $\pm \textbf{0.4}$} \\
Letters & 34.2 {\tiny $\pm 0.8$} & \textbf{45.6} {\tiny $\pm \textbf{0.8}$} & 45.2 {\tiny $\pm 1.4$} \\
\midrule
Average acc & 40.5 & \textbf{43.8} & 43.3 \\
\bottomrule
\end{tabular}
%\end{sc}
\end{small}
\end{center}
\caption{Accuracy comparison between NCM, LDA, and Linear head without using any adaptation (\textbf{NA} method). The reported results are based on \textbf{10} shots and averaged over 5 runs (mean±std). A pre-trained EfficientNet-B0 is used as a backbone in all cases.}
\label{table:head_comparison_na_10shots}
%\endgroup
\end{table}

\begin{table}[ht]
%\vspace*{1.5 cm}
%\begingroup

\setlength{\tabcolsep}{6.5pt} % Default value: 6pt %\renewcommand{\arraystretch}{.9} \begin{center}
\begin{center}
    
\begin{small}
%\begin{sc}
\begin{tabular}{l c c c}
\toprule
 \textbf{Dataset} & \textbf{NCM} & \textbf{LDA} &  \textbf{Linear} \\
 \midrule
Caltech101 & 90.4 {\tiny $\pm 0.6$} & 91.9 {\tiny $\pm 0.5$} & \textbf{93.0} {\tiny $\pm \textbf{0.4}$} \\
CIFAR100 & 52.0 {\tiny $\pm 0.9$} & 57.4 {\tiny $\pm 1.0$} & \textbf{60.9} {\tiny $\pm \textbf{1.0}$} \\
Flowers102 & 77.2 {\tiny $\pm 0.2$} & \textbf{83.9} {\tiny $\pm \textbf{0.3}$} & 81.9 {\tiny $\pm 0.2$} \\
Pets & 88.2 {\tiny $\pm 0.3$} & 89.5 {\tiny $\pm 0.4$} & \textbf{89.9} {\tiny $\pm \textbf{0.6}$} \\
Sun397 & 52.6 {\tiny $\pm 1.1$} & 55.9 {\tiny $\pm 1.0$} & \textbf{58.4} {\tiny $\pm \textbf{0.8}$} \\
SVHN & 19.3 {\tiny $\pm 1.7$} & 28.3 {\tiny $\pm 1.1$} & \textbf{28.9} {\tiny $\pm \textbf{1.1}$} \\
DTD & 58.5 {\tiny $\pm 0.0$} & 61.1 {\tiny $\pm 0.0$} & \textbf{65.4} {\tiny $\pm \textbf{0.2}$} \\
\midrule
EuroSAT & 81.7 {\tiny $\pm 0.6$} & 87.7 {\tiny $\pm 0.9$} & \textbf{88.2} {\tiny $\pm \textbf{0.7}$} \\
Resics45 & 66.6 {\tiny $\pm 0.7$} & 73.5 {\tiny $\pm 0.9$} & \textbf{78.2} {\tiny $\pm \textbf{0.7}$} \\
\midrule
Patch Camelyon & 70.9 {\tiny $\pm 5.4$} & \textbf{76.2} {\tiny $\pm \textbf{1.1}$} & 76.2 {\tiny $\pm 1.5$} \\
Retinopathy & 29.2 {\tiny $\pm 2.2$} & \textbf{32.9} {\tiny $\pm \textbf{2.2}$} & 32.9 {\tiny $\pm 1.9$} \\
\midrule
CLEVR-count & 26.3 {\tiny $\pm 1.5$} & 30.1 {\tiny $\pm 1.0$} & \textbf{31.2} {\tiny $\pm \textbf{1.0}$} \\
CLEVR-dist & 27.8 {\tiny $\pm 0.8$} & \textbf{32.2} {\tiny $\pm \textbf{1.1}$} & 31.7 {\tiny $\pm 1.0$} \\
dSprites-loc & 9.3 {\tiny $\pm 0.8$} & \textbf{11.9} {\tiny $\pm \textbf{0.4}$} & 9.9 {\tiny $\pm 0.6$} \\
dSprites-ori & 14.9 {\tiny $\pm 0.5$} & \textbf{20.1} {\tiny $\pm \textbf{1.1}$} & 18.7 {\tiny $\pm 2.2$} \\
SmallNORB-azi & 9.5 {\tiny $\pm 0.6$} & \textbf{12.3} {\tiny $\pm \textbf{0.8}$} & 12.1 {\tiny $\pm 1.1$} \\
SmallNORB-elev & 15.2 {\tiny $\pm 1.3$} & 19.1 {\tiny $\pm 0.7$} & \textbf{20.0} {\tiny $\pm \textbf{0.5}$} \\
DMLab & 29.1 {\tiny $\pm 0.3$} & 30.6 {\tiny $\pm 0.3$} & \textbf{32.3} {\tiny $\pm \textbf{0.7}$} \\
KITTI-dist & 53.2 {\tiny $\pm 0.9$} & \textbf{61.4} {\tiny $\pm \textbf{2.3}$} & 60.7 {\tiny $\pm 3.2$} \\
\midrule
FGVC-Aircraft & 30.9 {\tiny $\pm 0.5$} & 41.0 {\tiny $\pm 0.7$} & \textbf{46.1} {\tiny $\pm \textbf{0.6}$} \\
Cars & 33.7 {\tiny $\pm 0.0$} & 43.3 {\tiny $\pm 0.0$} & \textbf{46.5} {\tiny $\pm \textbf{0.2}$} \\
Letters & 43.1 {\tiny $\pm 1.0$} & 57.6 {\tiny $\pm 0.8$} & \textbf{65.9} {\tiny $\pm \textbf{1.1}$} \\
\midrule
Average acc & 44.5 & 49.9 & \textbf{51.3} \\
\bottomrule
\end{tabular}
%\end{sc}
\end{small}
\end{center}
\caption{Accuracy comparison between NCM, LDA, and Linear head without using any adaptation (\textbf{NA} method). The reported results are based on \textbf{50} shots and averaged over 5 runs (mean±std). A pre-trained EfficientNet-B0 is used as a backbone in all cases.}
\label{table:head_comparison_na_50shots}
%\endgroup
\end{table}

\begin{table}[ht]
%\vspace*{1.5 cm}
%\begingroup

\setlength{\tabcolsep}{6.5pt} % Default value: 6pt %\renewcommand{\arraystretch}{.9} 
\begin{center}
\begin{small}
%\begin{sc}
\begin{tabular}{l c c c}
\toprule
 \textbf{Dataset} & \textbf{NCM} & \textbf{LDA} &  \textbf{Linear} \\
 \midrule
Caltech101 & 90.4 {\tiny $\pm 0.6$} & 91.9 {\tiny $\pm 0.5$} & \textbf{93.0} {\tiny $\pm \textbf{0.4}$} \\
CIFAR100 & 53.5 {\tiny $\pm 1.6$} & 68.2 {\tiny $\pm 1.7$} & \textbf{68.3} {\tiny $\pm \textbf{0.8}$} \\
Flowers102 & 77.2 {\tiny $\pm 0.2$} & \textbf{83.9} {\tiny $\pm \textbf{0.3}$} & 81.9 {\tiny $\pm 0.3$} \\
Pets & 88.7 {\tiny $\pm 0.2$} & 89.9 {\tiny $\pm 0.2$} & \textbf{90.7} {\tiny $\pm \textbf{0.3}$} \\
Sun397 & 53.9 {\tiny $\pm 1.1$} & 56.9 {\tiny $\pm 1.2$} & \textbf{58.4} {\tiny $\pm \textbf{0.6}$} \\
SVHN & 24.5 {\tiny $\pm 0.4$} & 36.8 {\tiny $\pm 0.6$} & \textbf{40.8} {\tiny $\pm \textbf{0.8}$} \\
DTD & 58.5 {\tiny $\pm 0.0$} & 61.1 {\tiny $\pm 0.0$} & \textbf{65.3} {\tiny $\pm \textbf{0.3}$} \\
\midrule
EuroSAT & 82.4 {\tiny $\pm 0.3$} & 88.1 {\tiny $\pm 0.1$} & \textbf{93.2} {\tiny $\pm \textbf{0.2}$} \\
Resics45 & 67.1 {\tiny $\pm 0.2$} & 74.6 {\tiny $\pm 0.2$} & \textbf{81.8} {\tiny $\pm \textbf{0.4}$} \\
\midrule
Patch Camelyon & 72.9 {\tiny $\pm 0.0$} & 79.1 {\tiny $\pm 0.0$} & \textbf{79.7} {\tiny $\pm \textbf{0.5}$} \\
Retinopathy & 33.0 {\tiny $\pm 0.3$} & 40.4 {\tiny $\pm 0.3$} & \textbf{46.9} {\tiny $\pm \textbf{0.4}$} \\
\midrule
CLEVR-count & 28.9 {\tiny $\pm 0.3$} & 40.1 {\tiny $\pm 0.4$} & \textbf{50.3} {\tiny $\pm \textbf{0.3}$} \\
CLEVR-dist & 29.2 {\tiny $\pm 0.5$} & 38.7 {\tiny $\pm 0.4$} & \textbf{45.5} {\tiny $\pm \textbf{1.3}$} \\
dSprites-loc & 14.6 {\tiny $\pm 0.6$} & 20.9 {\tiny $\pm 0.7$} & \textbf{30.8} {\tiny $\pm \textbf{1.4}$} \\
dSprites-ori & 15.5 {\tiny $\pm 0.4$} & 22.5 {\tiny $\pm 0.6$} & \textbf{33.3} {\tiny $\pm \textbf{0.9}$} \\
SmallNORB-azi & 11.5 {\tiny $\pm 0.5$} & 14.1 {\tiny $\pm 0.7$} & \textbf{14.2} {\tiny $\pm \textbf{0.7}$} \\
SmallNORB-elev & 19.3 {\tiny $\pm 0.4$} & 24.1 {\tiny $\pm 0.5$} & \textbf{26.3} {\tiny $\pm \textbf{0.8}$} \\
DMLab & 35.4 {\tiny $\pm 0.4$} & 39.7 {\tiny $\pm 0.3$} & \textbf{44.4} {\tiny $\pm \textbf{0.4}$} \\
KITTI-dist & 53.4 {\tiny $\pm 0.0$} & 66.7 {\tiny $\pm 0.0$} & \textbf{69.8} {\tiny $\pm \textbf{0.4}$} \\
\midrule
FGVC-Aircraft & 31.8 {\tiny $\pm 0.0$} & 41.3 {\tiny $\pm 0.0$} & \textbf{45.8} {\tiny $\pm \textbf{0.4}$} \\
Cars & 33.7 {\tiny $\pm 0.0$} & 43.3 {\tiny $\pm 0.0$} & \textbf{46.6} {\tiny $\pm \textbf{0.3}$} \\
Letters & 44.9 {\tiny $\pm 1.3$} & 59.7 {\tiny $\pm 0.5$} & \textbf{69.5} {\tiny $\pm \textbf{0.5}$} \\
\midrule
Average acc & 46.4 & 53.4 & \textbf{58.0} \\
\bottomrule
\end{tabular}
%\end{sc}
\end{small}
\end{center}
\caption{Accuracy comparison between NCM, LDA, and Linear head without using any adaptation (\textbf{NA} method). The reported results are based on the full training dataset and averaged over 5 runs (mean±std). A pre-trained EfficientNet-B0 is used as a backbone in all cases.}
\label{table:head_comparison_na_alldata}
%\endgroup
\end{table}

%%%%%%%%%%%%
% A-FiLM %%%
%%%%%%%%%%%%

\begin{table}[ht]
%\vspace*{1.5 cm}
%\begingroup

\setlength{\tabcolsep}{6.5pt} % Default value: 6pt %\renewcommand{\arraystretch}{.9} 
\begin{center}
\begin{small}
%\begin{sc}
\begin{tabular}{l c c c}
\toprule
 \textbf{Dataset} & \textbf{NCM} & \textbf{LDA} &  \textbf{Linear} \\
 \midrule
Caltech101 & 86.6 {\tiny $\pm 0.5$} & \textbf{89.0} {\tiny $\pm \textbf{0.6}$} & 88.5 {\tiny $\pm 0.4$} \\
CIFAR100 & 47.4 {\tiny $\pm 1.2$} & \textbf{51.8} {\tiny $\pm \textbf{1.3}$} & 50.3 {\tiny $\pm 1.4$} \\
Flowers102 & 80.2 {\tiny $\pm 0.4$} & \textbf{85.0} {\tiny $\pm \textbf{0.8}$} & 83.5 {\tiny $\pm 0.5$} \\
Pets & 81.8 {\tiny $\pm 1.5$} & 81.8 {\tiny $\pm 1.7$} & \textbf{82.6} {\tiny $\pm \textbf{1.6}$} \\
Sun397 & 40.9 {\tiny $\pm 0.7$} & \textbf{40.9} {\tiny $\pm \textbf{0.7}$} & 38.1 {\tiny $\pm 1.0$} \\
SVHN & 28.6 {\tiny $\pm 3.8$} & \textbf{31.7} {\tiny $\pm \textbf{3.7}$} & 30.1 {\tiny $\pm 4.5$} \\
DTD & 49.6 {\tiny $\pm 1.5$} & 50.2 {\tiny $\pm 0.9$} & \textbf{50.8} {\tiny $\pm \textbf{1.3}$} \\
\midrule
EuroSAT & 75.8 {\tiny $\pm 1.5$} & 78.1 {\tiny $\pm 1.2$} & \textbf{78.3} {\tiny $\pm \textbf{1.2}$} \\
Resics45 & 62.7 {\tiny $\pm 1.1$} & 64.7 {\tiny $\pm 1.2$} & \textbf{65.8} {\tiny $\pm \textbf{0.7}$} \\
\midrule
Patch Camelyon & 64.7 {\tiny $\pm 5.8$} & \textbf{64.9} {\tiny $\pm \textbf{6.6}$} & 62.9 {\tiny $\pm 5.4$} \\
Retinopathy & \textbf{27.4} {\tiny $\pm \textbf{3.1}$} & 26.0 {\tiny $\pm 2.0$} & 25.2 {\tiny $\pm 2.4$} \\
\midrule
CLEVR-count & \textbf{24.0} {\tiny $\pm \textbf{1.3}$} & 23.4 {\tiny $\pm 1.4$} & 23.2 {\tiny $\pm 0.7$} \\
CLEVR-dist & 23.1 {\tiny $\pm 1.4$} & 23.1 {\tiny $\pm 1.1$} & \textbf{24.0} {\tiny $\pm \textbf{1.3}$} \\
dSprites-loc & 19.5 {\tiny $\pm 1.8$} & \textbf{19.8} {\tiny $\pm \textbf{2.0}$} & 16.7 {\tiny $\pm 5.9$} \\
dSprites-ori & 20.6 {\tiny $\pm 1.6$} & \textbf{26.5} {\tiny $\pm \textbf{0.8}$} & 25.5 {\tiny $\pm 1.4$} \\
SmallNORB-azi & 9.0 {\tiny $\pm 1.0$} & 10.1 {\tiny $\pm 0.6$} & \textbf{10.3} {\tiny $\pm \textbf{0.6}$} \\
SmallNORB-elev & 14.6 {\tiny $\pm 0.9$} & 15.4 {\tiny $\pm 0.7$} & \textbf{15.5} {\tiny $\pm \textbf{0.8}$} \\
DMLab & 23.4 {\tiny $\pm 2.3$} & 23.3 {\tiny $\pm 1.9$} & \textbf{24.8} {\tiny $\pm \textbf{0.9}$} \\
KITTI-dist & \textbf{55.7} {\tiny $\pm \textbf{3.6}$} & 52.7 {\tiny $\pm 3.5$} & 53.2 {\tiny $\pm 1.8$} \\
\midrule
FGVC-Aircraft & 28.7 {\tiny $\pm 0.7$} & 32.6 {\tiny $\pm 0.9$} & \textbf{33.1} {\tiny $\pm \textbf{1.3}$} \\
Cars & 22.7 {\tiny $\pm 0.5$} & \textbf{28.1} {\tiny $\pm \textbf{0.4}$} & 27.2 {\tiny $\pm 0.7$} \\
Letters & 52.2 {\tiny $\pm 2.9$} & 55.9 {\tiny $\pm 2.5$} & \textbf{56.4} {\tiny $\pm \textbf{3.2}$} \\
\midrule
Average acc & 42.7 & \textbf{44.3} & 43.9 \\
\bottomrule
\end{tabular}
%\end{sc}
\end{small}
\end{center}

\caption{Accuracy comparison between NCM, LDA, and Linear head using FiLM adaptation (\textbf{A-FiLM} method). The reported results are based on \textbf{5} shots and averaged over 5 runs (mean±std). A pre-trained EfficientNet-B0 is used as a backbone in all cases.}
\label{table:head_comparison_film_5shots}
%\endgroup
\end{table}

\begin{table}[ht]
%\vspace*{1.5 cm}
%\begingroup
\setlength{\tabcolsep}{8.6pt} % Default value: 6pt
\begin{center}
\begin{small}
%\begin{sc}
\begin{tabular}{l c c c}
\toprule
 \textbf{Dataset} & \textbf{NCM} & \textbf{LDA} &  \textbf{Linear} \\
 \midrule
Caltech101 & 89.9 {\tiny $\pm 0.2$} & \textbf{91.5} {\tiny $\pm \textbf{0.4}$} & 91.1 {\tiny $\pm 0.6$} \\
CIFAR100 & 59.2 {\tiny $\pm 1.7$} & \textbf{62.8} {\tiny $\pm \textbf{1.1}$} & 60.6 {\tiny $\pm 0.5$} \\
Flowers102 & 86.2 {\tiny $\pm 0.5$} & 91.1 {\tiny $\pm 0.3$} & \textbf{91.2} {\tiny $\pm \textbf{0.3}$} \\
Pets & 85.8 {\tiny $\pm 1.0$} & 86.3 {\tiny $\pm 0.8$} & \textbf{86.6} {\tiny $\pm \textbf{0.8}$} \\
Sun397 & 48.5 {\tiny $\pm 0.8$} & \textbf{49.3} {\tiny $\pm \textbf{0.6}$} & 44.8 {\tiny $\pm 1.2$} \\
SVHN & 40.3 {\tiny $\pm 2.5$} & \textbf{45.3} {\tiny $\pm \textbf{3.0}$} & 43.9 {\tiny $\pm 3.3$} \\
DTD & 58.3 {\tiny $\pm 0.8$} & 59.1 {\tiny $\pm 0.8$} & \textbf{59.2} {\tiny $\pm \textbf{1.1}$} \\
\midrule
EuroSAT & 81.9 {\tiny $\pm 1.4$} & \textbf{84.4} {\tiny $\pm \textbf{1.6}$} & 83.5 {\tiny $\pm 0.7$} \\
Resics45 & 70.2 {\tiny $\pm 1.1$} & 73.0 {\tiny $\pm 1.3$} & \textbf{73.5} {\tiny $\pm \textbf{0.7}$} \\
\midrule
Patch Camelyon & \textbf{69.2} {\tiny $\pm \textbf{5.2}$} & 68.5 {\tiny $\pm 6.1$} & 67.1 {\tiny $\pm 4.8$} \\
Retinopathy & 26.7 {\tiny $\pm 1.3$} & \textbf{26.9} {\tiny $\pm \textbf{1.1}$} & 25.6 {\tiny $\pm 1.0$} \\
\midrule
CLEVR-count & \textbf{30.1} {\tiny $\pm \textbf{1.9}$} & 27.6 {\tiny $\pm 1.6$} & 29.2 {\tiny $\pm 1.2$} \\
CLEVR-dist & 25.3 {\tiny $\pm 1.7$} & 26.2 {\tiny $\pm 1.3$} & \textbf{26.5} {\tiny $\pm \textbf{0.7}$} \\
dSprites-loc & \textbf{26.2} {\tiny $\pm \textbf{11.4}$} & 26.1 {\tiny $\pm 10.6$} & 24.2 {\tiny $\pm 14.7$} \\
dSprites-ori & 26.7 {\tiny $\pm 2.4$} & \textbf{34.0} {\tiny $\pm \textbf{2.6}$} & 33.8 {\tiny $\pm 2.4$} \\
SmallNORB-azi & 11.0 {\tiny $\pm 0.8$} & \textbf{11.7} {\tiny $\pm \textbf{1.2}$} & 11.3 {\tiny $\pm 1.4$} \\
SmallNORB-elev & 15.6 {\tiny $\pm 0.7$} & 16.3 {\tiny $\pm 0.4$} & \textbf{16.6} {\tiny $\pm \textbf{0.3}$} \\
DMLab & 27.2 {\tiny $\pm 1.9$} & 26.6 {\tiny $\pm 1.5$} & \textbf{29.3} {\tiny $\pm \textbf{1.7}$} \\
KITTI-dist & \textbf{56.7} {\tiny $\pm \textbf{3.7}$} & 55.4 {\tiny $\pm 3.7$} & 56.2 {\tiny $\pm 4.5$} \\
\midrule
FGVC-Aircraft & 37.5 {\tiny $\pm 0.7$} & 43.3 {\tiny $\pm 1.1$} & \textbf{43.4} {\tiny $\pm \textbf{1.5}$} \\
Cars & 36.3 {\tiny $\pm 0.8$} & \textbf{43.1} {\tiny $\pm \textbf{1.0}$} & 42.1 {\tiny $\pm 1.0$} \\
Letters & 64.0 {\tiny $\pm 1.5$} & 67.5 {\tiny $\pm 1.3$} & \textbf{68.1} {\tiny $\pm \textbf{1.4}$} \\
\midrule
Average acc & 48.8 & \textbf{50.7} & 50.4 \\
\bottomrule
\end{tabular}
%\end{sc}
\end{small}
\end{center}
\caption{Accuracy comparison between NCM, LDA, and Linear head using FiLM adaptation (\textbf{A-FiLM} method). The reported results are based on \textbf{10} shots and averaged over 5 runs (mean±std). A pre-trained EfficientNet-B0 is used as a backbone in all cases.}
\label{table:head_comparison_film_10shots}
%\endgroup
\end{table}

\begin{table}[ht]
%\vspace*{1.5 cm}
%\begingroup

\setlength{\tabcolsep}{6.5pt} % Default value: 6pt %\renewcommand{\arraystretch}{.9} 
\begin{center}
\begin{small}
%\begin{sc}
\begin{tabular}{l c c c}
\toprule
 \textbf{Dataset} & \textbf{NCM} & \textbf{LDA} &  \textbf{Linear} \\
 \midrule
Caltech101 & 93.4 {\tiny $\pm 0.7$} & \textbf{93.8} {\tiny $\pm \textbf{0.5}$} & 93.5 {\tiny $\pm 1.0$} \\
CIFAR100 & 72.6 {\tiny $\pm 0.7$} & \textbf{73.8} {\tiny $\pm \textbf{0.9}$} & 73.7 {\tiny $\pm 0.5$} \\
Flowers102 & 86.2 {\tiny $\pm 0.5$} & 91.1 {\tiny $\pm 0.3$} & \textbf{91.2} {\tiny $\pm \textbf{0.3}$} \\
Pets & 89.3 {\tiny $\pm 0.6$} & \textbf{89.9} {\tiny $\pm \textbf{0.7}$} & 89.9 {\tiny $\pm 0.7$} \\
Sun397 & 58.5 {\tiny $\pm 0.7$} & 59.7 {\tiny $\pm 0.6$} & \textbf{60.8} {\tiny $\pm \textbf{0.7}$} \\
SVHN & 73.9 {\tiny $\pm 1.1$} & \textbf{77.2} {\tiny $\pm \textbf{0.8}$} & 76.8 {\tiny $\pm 1.0$} \\
DTD & 66.8 {\tiny $\pm 0.3$} & 68.4 {\tiny $\pm 0.2$} & \textbf{68.7} {\tiny $\pm \textbf{0.7}$} \\
\midrule
EuroSAT & 91.0 {\tiny $\pm 0.6$} & 93.0 {\tiny $\pm 0.6$} & \textbf{93.2} {\tiny $\pm \textbf{0.6}$} \\
Resics45 & 81.7 {\tiny $\pm 0.2$} & 83.4 {\tiny $\pm 0.6$} & \textbf{85.3} {\tiny $\pm \textbf{0.6}$} \\
\midrule
Patch Camelyon & \textbf{78.5} {\tiny $\pm \textbf{2.0}$} & 77.9 {\tiny $\pm 2.4$} & 78.1 {\tiny $\pm 2.4$} \\
Retinopathy & 34.2 {\tiny $\pm 1.6$} & \textbf{35.2} {\tiny $\pm \textbf{1.2}$} & 33.5 {\tiny $\pm 1.9$} \\
\midrule
CLEVR-count & \textbf{56.8} {\tiny $\pm \textbf{0.9}$} & 46.6 {\tiny $\pm 1.1$} & 53.1 {\tiny $\pm 1.1$} \\
CLEVR-dist & 40.2 {\tiny $\pm 1.8$} & 38.8 {\tiny $\pm 1.0$} & \textbf{41.2} {\tiny $\pm \textbf{1.5}$} \\
dSprites-loc & 83.6 {\tiny $\pm 5.4$} & \textbf{83.7} {\tiny $\pm \textbf{5.6}$} & 81.6 {\tiny $\pm 5.9$} \\
dSprites-ori & 41.2 {\tiny $\pm 0.8$} & 52.1 {\tiny $\pm 1.3$} & \textbf{53.8} {\tiny $\pm \textbf{1.4}$} \\
SmallNORB-azi & 17.0 {\tiny $\pm 0.8$} & 16.8 {\tiny $\pm 0.8$} & \textbf{17.5} {\tiny $\pm \textbf{1.0}$} \\
SmallNORB-elev & 23.1 {\tiny $\pm 1.2$} & 22.9 {\tiny $\pm 0.5$} & \textbf{24.0} {\tiny $\pm \textbf{0.6}$} \\
DMLab & 35.0 {\tiny $\pm 0.6$} & 34.6 {\tiny $\pm 0.6$} & \textbf{35.8} {\tiny $\pm \textbf{0.4}$} \\
KITTI-dist & \textbf{67.3} {\tiny $\pm \textbf{2.1}$} & 66.8 {\tiny $\pm 3.0$} & 66.5 {\tiny $\pm 2.7$} \\
\midrule
FGVC-Aircraft & 60.6 {\tiny $\pm 0.9$} & 65.1 {\tiny $\pm 0.7$} & \textbf{68.0} {\tiny $\pm \textbf{0.6}$} \\
Cars & 60.9 {\tiny $\pm 0.2$} & 67.9 {\tiny $\pm 0.2$} & \textbf{74.5} {\tiny $\pm \textbf{0.4}$} \\
Letters & 76.7 {\tiny $\pm 0.5$} & 79.7 {\tiny $\pm 0.4$} & \textbf{81.9} {\tiny $\pm \textbf{0.8}$} \\
\midrule
Average acc & 63.1 & 64.5 & \textbf{65.6} \\
\bottomrule
\end{tabular}
%\end{sc}
\end{small}
\end{center}
\caption{Accuracy comparison between NCM, LDA, and Linear head using FiLM adaptation (\textbf{A-FiLM} method). The reported results are based on \textbf{50} shots and averaged over 5 runs (mean±std). A pre-trained EfficientNet-B0 is used as a backbone in all cases.}
\label{table:head_comparison_film_50shots}
%\endgroup
\end{table}

\begin{table}[ht]
%\vspace*{1.5 cm}
%\begingroup

\setlength{\tabcolsep}{6.5pt} % Default value: 6pt %\renewcommand{\arraystretch}{.9} 
\begin{center}
\begin{small}
%\begin{sc}
\begin{tabular}{l c c c}
\toprule
 \textbf{Dataset} & \textbf{NCM} & \textbf{LDA} &  \textbf{Linear} \\
 \midrule
Caltech101 & 93.6 {\tiny $\pm 0.4$} & \textbf{94.4} {\tiny $\pm \textbf{0.3}$} & 94.1 {\tiny $\pm 0.6$} \\
CIFAR100 & 77.4 {\tiny $\pm 1.1$} & 78.2 {\tiny $\pm 1.1$} & \textbf{82.1} {\tiny $\pm \textbf{1.1}$} \\
Flowers102 & 88.6 {\tiny $\pm 0.6$} & \textbf{91.2} {\tiny $\pm \textbf{0.4}$} & 90.6 {\tiny $\pm 0.5$} \\
Pets & 90.2 {\tiny $\pm 0.2$} & 90.8 {\tiny $\pm 0.3$} & \textbf{91.0} {\tiny $\pm \textbf{0.4}$} \\
Sun397 & 61.4 {\tiny $\pm 0.5$} & 62.1 {\tiny $\pm 0.3$} & \textbf{63.7} {\tiny $\pm \textbf{0.9}$} \\
SVHN & 92.9 {\tiny $\pm 0.4$} & 93.1 {\tiny $\pm 0.4$} & \textbf{95.1} {\tiny $\pm \textbf{0.4}$} \\
DTD & 64.8 {\tiny $\pm 0.2$} & 66.8 {\tiny $\pm 0.5$} & \textbf{67.6} {\tiny $\pm \textbf{0.6}$} \\
\midrule
EuroSAT & 96.5 {\tiny $\pm 0.3$} & 97.2 {\tiny $\pm 0.1$} & \textbf{98.1} {\tiny $\pm \textbf{0.3}$} \\
Resics45 & 88.0 {\tiny $\pm 0.3$} & 89.4 {\tiny $\pm 0.4$} & \textbf{94.2} {\tiny $\pm \textbf{0.4}$} \\
\midrule
Patch Camelyon & 84.3 {\tiny $\pm 1.5$} & \textbf{85.9} {\tiny $\pm \textbf{1.2}$} & 85.8 {\tiny $\pm 1.6$} \\
Retinopathy & 52.3 {\tiny $\pm 0.7$} & 52.6 {\tiny $\pm 0.8$} & \textbf{59.5} {\tiny $\pm \textbf{0.8}$} \\
\midrule
CLEVR-count & 94.5 {\tiny $\pm 0.2$} & 93.5 {\tiny $\pm 0.7$} & \textbf{95.3} {\tiny $\pm \textbf{0.6}$} \\
CLEVR-dist & 79.3 {\tiny $\pm 1.4$} & 80.1 {\tiny $\pm 2.3$} & \textbf{84.5} {\tiny $\pm \textbf{2.0}$} \\
dSprites-loc & 98.0 {\tiny $\pm 0.5$} & 98.5 {\tiny $\pm 0.6$} & \textbf{99.3} {\tiny $\pm \textbf{0.4}$} \\
dSprites-ori & 69.8 {\tiny $\pm 2.4$} & 80.0 {\tiny $\pm 0.8$} & \textbf{90.7} {\tiny $\pm \textbf{0.8}$} \\
SmallNORB-azi & \textbf{26.1} {\tiny $\pm \textbf{1.2}$} & 24.4 {\tiny $\pm 0.6$} & 23.5 {\tiny $\pm 0.5$} \\
SmallNORB-elev & 47.0 {\tiny $\pm 0.9$} & \textbf{47.9} {\tiny $\pm \textbf{1.2}$} & 47.9 {\tiny $\pm 1.4$} \\
DMLab & 60.2 {\tiny $\pm 0.6$} & 61.3 {\tiny $\pm 0.5$} & \textbf{65.8} {\tiny $\pm \textbf{1.0}$} \\
KITTI-dist & 78.5 {\tiny $\pm 0.9$} & \textbf{80.1} {\tiny $\pm \textbf{1.1}$} & 79.7 {\tiny $\pm 0.4$} \\
\midrule
FGVC-Aircraft & 63.3 {\tiny $\pm 0.2$} & 67.5 {\tiny $\pm 0.4$} & \textbf{71.5} {\tiny $\pm \textbf{0.5}$} \\
Cars & 60.1 {\tiny $\pm 0.1$} & 67.3 {\tiny $\pm 0.3$} & \textbf{73.6} {\tiny $\pm \textbf{0.3}$} \\
Letters & 77.1 {\tiny $\pm 0.3$} & 81.7 {\tiny $\pm 0.4$} & \textbf{85.2} {\tiny $\pm \textbf{0.3}$} \\
\midrule
Average acc & 74.7 & 76.6 & \textbf{79.0} \\
\bottomrule
\end{tabular}
%\end{sc}
\end{small}
\end{center}
\caption{Accuracy comparison between NCM, LDA, and Linear head using FiLM adaptation (\textbf{A-FiLM} method). The reported results are based on the full training dataset and averaged over 5 runs (mean±std). A pre-trained EfficientNet-B0 is used as a backbone in all cases.}
\label{table:head_comparison_film_alldata}
%\endgroup
\end{table}

%%%%%%%%%%%%
% A-Full Body %%%
%%%%%%%%%%%%

\begin{table}[ht]
%\vspace*{1.5 cm}
%\begingroup

\setlength{\tabcolsep}{6.5pt} % Default value: 6pt %\renewcommand{\arraystretch}{.9} 
\begin{center}
\begin{small}
%\begin{sc}
\begin{tabular}{l c c c}
\toprule
 \textbf{Dataset} & \textbf{NCM} & \textbf{LDA} &  \textbf{Linear} \\
 \midrule
Caltech101 & 87.1 {\tiny $\pm 0.6$} & \textbf{89.4} {\tiny $\pm \textbf{0.7}$} & 86.1 {\tiny $\pm 0.9$} \\
CIFAR100 & 48.1 {\tiny $\pm 0.7$} & \textbf{49.3} {\tiny $\pm \textbf{1.1}$} & 49.2 {\tiny $\pm 1.0$} \\
Flowers102 & 81.5 {\tiny $\pm 0.8$} & 81.7 {\tiny $\pm 0.6$} & \textbf{83.6} {\tiny $\pm \textbf{0.8}$} \\
Pets & \textbf{80.9} {\tiny $\pm \textbf{1.7}$} & 80.8 {\tiny $\pm 1.7$} & 71.1 {\tiny $\pm 1.2$} \\
Sun397 & 35.5 {\tiny $\pm 0.3$} & 35.2 {\tiny $\pm 0.5$} & \textbf{37.2} {\tiny $\pm \textbf{0.3}$} \\
SVHN & 19.1 {\tiny $\pm 1.5$} & \textbf{19.6} {\tiny $\pm \textbf{1.2}$} & 19.2 {\tiny $\pm 1.8$} \\
DTD & 48.4 {\tiny $\pm 1.3$} & \textbf{49.0} {\tiny $\pm \textbf{1.3}$} & 41.6 {\tiny $\pm 0.8$} \\
\midrule
EuroSAT & 75.4 {\tiny $\pm 4.1$} & 76.7 {\tiny $\pm 2.9$} & \textbf{78.5} {\tiny $\pm \textbf{1.2}$} \\
Resics45 & 61.7 {\tiny $\pm 2.3$} & 62.3 {\tiny $\pm 2.5$} & \textbf{63.5} {\tiny $\pm \textbf{2.5}$} \\
\midrule
Patch Camelyon & 59.2 {\tiny $\pm 4.9$} & 59.4 {\tiny $\pm 4.9$} & \textbf{60.5} {\tiny $\pm \textbf{7.0}$} \\
Retinopathy & 24.6 {\tiny $\pm 2.7$} & 24.5 {\tiny $\pm 2.5$} & \textbf{26.1} {\tiny $\pm \textbf{2.4}$} \\
\midrule
CLEVR-count & 23.9 {\tiny $\pm 2.9$} & 23.5 {\tiny $\pm 3.0$} & \textbf{24.2} {\tiny $\pm \textbf{3.1}$} \\
CLEVR-dist & 25.1 {\tiny $\pm 3.3$} & 25.6 {\tiny $\pm 3.5$} & \textbf{25.6} {\tiny $\pm \textbf{2.4}$} \\
dSprites-loc & 26.1 {\tiny $\pm 2.7$} & \textbf{27.1} {\tiny $\pm \textbf{1.6}$} & 25.5 {\tiny $\pm 3.6$} \\
dSprites-ori & 18.3 {\tiny $\pm 1.6$} & \textbf{19.9} {\tiny $\pm \textbf{1.5}$} & 15.6 {\tiny $\pm 1.9$} \\
SmallNORB-azi & 10.0 {\tiny $\pm 0.7$} & \textbf{10.4} {\tiny $\pm \textbf{0.8}$} & 10.3 {\tiny $\pm 0.9$} \\
SmallNORB-elev & 15.8 {\tiny $\pm 1.1$} & \textbf{16.2} {\tiny $\pm \textbf{1.1}$} & 15.6 {\tiny $\pm 0.8$} \\
DMLab & 21.3 {\tiny $\pm 1.6$} & 22.1 {\tiny $\pm 1.3$} & \textbf{22.6} {\tiny $\pm \textbf{0.9}$} \\
KITTI-dist & 51.1 {\tiny $\pm 2.5$} & \textbf{52.8} {\tiny $\pm \textbf{2.5}$} & 52.1 {\tiny $\pm 2.0$} \\
\midrule
FGVC-Aircraft & 23.8 {\tiny $\pm 0.7$} & 23.6 {\tiny $\pm 0.8$} & \textbf{25.1} {\tiny $\pm \textbf{1.0}$} \\
Cars & 23.1 {\tiny $\pm 0.3$} & 23.5 {\tiny $\pm 0.3$} & \textbf{25.3} {\tiny $\pm \textbf{0.5}$} \\
Letters & 35.5 {\tiny $\pm 3.3$} & 35.7 {\tiny $\pm 3.3$} & \textbf{37.0} {\tiny $\pm \textbf{3.2}$} \\
\midrule
Average acc & 40.7 & \textbf{41.3} & 40.7 \\
\bottomrule
\end{tabular}
%\end{sc}
\end{small}
\end{center}
\caption{Accuracy comparison between NCM, LDA, and Linear head using full-body adaptation (\textbf{A-FB} method). The reported results are based on \textbf{5} shots and averaged over 5 runs (mean±std). A pre-trained EfficientNet-B0 is used as a backbone in all cases.}
\label{table:head_comparison_full_5shots}
%\endgroup
\end{table}

\begin{table}[ht]
%\vspace*{1.5 cm}
%\begingroup

\setlength{\tabcolsep}{6.5pt} % Default value: 6pt
\begin{center}
\begin{small}
%\begin{sc}
\begin{tabular}{l c c c}
\toprule
 \textbf{Dataset} & \textbf{NCM} & \textbf{LDA} &  \textbf{Linear} \\
 \midrule
Caltech101 & 90.4 {\tiny $\pm 0.8$} & \textbf{91.7} {\tiny $\pm \textbf{0.7}$} & 89.3 {\tiny $\pm 0.8$} \\
CIFAR100 & 58.8 {\tiny $\pm 0.4$} & \textbf{59.5} {\tiny $\pm \textbf{0.3}$} & 59.5 {\tiny $\pm 0.8$} \\
Flowers102 & 89.9 {\tiny $\pm 0.6$} & 90.0 {\tiny $\pm 0.5$} & \textbf{91.3} {\tiny $\pm \textbf{0.6}$} \\
Pets & 85.5 {\tiny $\pm 0.9$} & \textbf{85.7} {\tiny $\pm \textbf{0.5}$} & 78.4 {\tiny $\pm 1.7$} \\
Sun397 & 45.2 {\tiny $\pm 0.7$} & 45.0 {\tiny $\pm 0.4$} & \textbf{46.3} {\tiny $\pm \textbf{0.3}$} \\
SVHN & 26.4 {\tiny $\pm 3.4$} & \textbf{27.1} {\tiny $\pm \textbf{3.8}$} & 26.4 {\tiny $\pm 3.6$} \\
DTD & 55.2 {\tiny $\pm 0.8$} & \textbf{56.7} {\tiny $\pm \textbf{1.0}$} & 48.0 {\tiny $\pm 0.8$} \\
\midrule
EuroSAT & 83.9 {\tiny $\pm 1.5$} & 84.9 {\tiny $\pm 1.5$} & \textbf{86.3} {\tiny $\pm \textbf{1.5}$} \\
Resics45 & 72.6 {\tiny $\pm 1.2$} & 72.8 {\tiny $\pm 1.3$} & \textbf{74.3} {\tiny $\pm \textbf{1.1}$} \\
\midrule
Patch Camelyon & 61.3 {\tiny $\pm 4.5$} & 61.2 {\tiny $\pm 4.5$} & \textbf{63.0} {\tiny $\pm \textbf{5.1}$} \\
Retinopathy & 25.1 {\tiny $\pm 4.2$} & 24.6 {\tiny $\pm 3.3$} & \textbf{27.5} {\tiny $\pm \textbf{2.0}$} \\
\midrule
CLEVR-count & 28.0 {\tiny $\pm 1.9$} & 27.7 {\tiny $\pm 2.0$} & \textbf{28.7} {\tiny $\pm \textbf{2.1}$} \\
CLEVR-dist & \textbf{30.1} {\tiny $\pm \textbf{4.1}$} & 30.0 {\tiny $\pm 4.2$} & 29.9 {\tiny $\pm 3.6$} \\
dSprites-loc & 42.9 {\tiny $\pm 2.7$} & \textbf{44.8} {\tiny $\pm \textbf{1.3}$} & 42.2 {\tiny $\pm 3.1$} \\
dSprites-ori & 26.4 {\tiny $\pm 6.5$} & \textbf{29.6} {\tiny $\pm \textbf{6.8}$} & 22.9 {\tiny $\pm 10.4$} \\
SmallNORB-azi & \textbf{12.2} {\tiny $\pm \textbf{0.6}$} & 11.5 {\tiny $\pm 0.8$} & 11.7 {\tiny $\pm 0.6$} \\
SmallNORB-elev & 18.2 {\tiny $\pm 1.3$} & \textbf{18.3} {\tiny $\pm \textbf{1.5}$} & 17.3 {\tiny $\pm 1.1$} \\
DMLab & 25.3 {\tiny $\pm 1.2$} & 25.6 {\tiny $\pm 1.3$} & \textbf{26.1} {\tiny $\pm \textbf{1.1}$} \\
KITTI-dist & 53.5 {\tiny $\pm 2.3$} & \textbf{54.9} {\tiny $\pm \textbf{1.3}$} & 52.2 {\tiny $\pm 2.1$} \\
\midrule
FGVC-Aircraft & 37.4 {\tiny $\pm 0.5$} & 36.7 {\tiny $\pm 0.6$} & \textbf{38.6} {\tiny $\pm \textbf{0.4}$} \\
Cars & 43.8 {\tiny $\pm 0.8$} & 43.5 {\tiny $\pm 0.6$} & \textbf{46.9} {\tiny $\pm \textbf{0.8}$} \\
Letters & 55.8 {\tiny $\pm 1.5$} & 55.2 {\tiny $\pm 1.3$} & \textbf{56.6} {\tiny $\pm \textbf{1.4}$} \\
\midrule
Average acc & 48.5 & \textbf{49.0} & 48.3 \\
\bottomrule
\end{tabular}
%\end{sc}
\end{small}
\end{center}
\caption{Accuracy comparison between NCM, LDA, and Linear head using full-body adaptation (\textbf{A-FB} method). The reported results are based on \textbf{10} shots and averaged over 5 runs (mean±std). A pre-trained EfficientNet-B0 is used as a backbone in all cases.}
\label{table:head_comparison_full_10shots}
%\endgroup
\end{table}

\begin{table}[ht]
%\vspace*{1.5 cm}
%\begingroup

\setlength{\tabcolsep}{6.5pt} % Default value: 6pt %\renewcommand{\arraystretch}{.9} 
\begin{center}
\begin{small}
%\begin{sc}
\begin{tabular}{l c c c}
\toprule
 \textbf{Dataset} & \textbf{NCM} & \textbf{LDA} &  \textbf{Linear} \\
 \midrule
Caltech101 & 92.8 {\tiny $\pm 0.3$} & \textbf{93.9} {\tiny $\pm \textbf{0.3}$} & 92.6 {\tiny $\pm 0.5$} \\
CIFAR100 & 72.9 {\tiny $\pm 0.9$} & \textbf{73.1} {\tiny $\pm \textbf{0.7}$} & 72.7 {\tiny $\pm 0.8$} \\
Flowers102 & 89.9 {\tiny $\pm 0.6$} & 90.0 {\tiny $\pm 0.5$} & \textbf{91.3} {\tiny $\pm \textbf{0.6}$} \\
Pets & 88.7 {\tiny $\pm 0.5$} & \textbf{89.6} {\tiny $\pm \textbf{0.6}$} & 84.9 {\tiny $\pm 1.1$} \\
Sun397 & 59.9 {\tiny $\pm 0.6$} & 60.4 {\tiny $\pm 0.6$} & \textbf{62.4} {\tiny $\pm \textbf{0.4}$} \\
SVHN & 63.9 {\tiny $\pm 1.5$} & 64.5 {\tiny $\pm 1.3$} & \textbf{64.6} {\tiny $\pm \textbf{1.6}$} \\
DTD & 60.9 {\tiny $\pm 0.3$} & \textbf{64.4} {\tiny $\pm \textbf{0.2}$} & 57.4 {\tiny $\pm 0.7$} \\
\midrule
EuroSAT & 93.8 {\tiny $\pm 0.5$} & \textbf{94.1} {\tiny $\pm \textbf{0.6}$} & 93.9 {\tiny $\pm 1.0$} \\
Resics45 & 87.4 {\tiny $\pm 0.8$} & 87.6 {\tiny $\pm 0.7$} & \textbf{88.0} {\tiny $\pm \textbf{0.8}$} \\
\midrule
Patch Camelyon & 71.6 {\tiny $\pm 1.7$} & 72.2 {\tiny $\pm 1.8$} & \textbf{74.4} {\tiny $\pm \textbf{1.8}$} \\
Retinopathy & 31.1 {\tiny $\pm 3.3$} & 31.3 {\tiny $\pm 2.9$} & \textbf{31.9} {\tiny $\pm \textbf{2.2}$} \\
\midrule
CLEVR-count & \textbf{46.5} {\tiny $\pm \textbf{1.0}$} & 45.2 {\tiny $\pm 1.1$} & 45.4 {\tiny $\pm 1.9$} \\
CLEVR-dist & 44.0 {\tiny $\pm 2.1$} & 44.7 {\tiny $\pm 2.2$} & \textbf{45.7} {\tiny $\pm \textbf{2.2}$} \\
dSprites-loc & 85.3 {\tiny $\pm 4.0$} & \textbf{87.1} {\tiny $\pm \textbf{2.4}$} & 85.2 {\tiny $\pm 3.2$} \\
dSprites-ori & 42.6 {\tiny $\pm 3.2$} & \textbf{44.8} {\tiny $\pm \textbf{2.3}$} & 39.3 {\tiny $\pm 4.7$} \\
SmallNORB-azi & \textbf{19.6} {\tiny $\pm \textbf{0.6}$} & 18.3 {\tiny $\pm 0.7$} & 19.1 {\tiny $\pm 1.1$} \\
SmallNORB-elev & \textbf{31.4} {\tiny $\pm \textbf{1.4}$} & 31.3 {\tiny $\pm 1.7$} & 31.3 {\tiny $\pm 2.1$} \\
DMLab & 32.6 {\tiny $\pm 1.4$} & 32.7 {\tiny $\pm 1.6$} & \textbf{34.0} {\tiny $\pm \textbf{0.9}$} \\
KITTI-dist & 65.8 {\tiny $\pm 2.1$} & \textbf{66.9} {\tiny $\pm \textbf{2.2}$} & 65.8 {\tiny $\pm 1.8$} \\
\midrule
FGVC-Aircraft & 74.2 {\tiny $\pm 0.8$} & 73.5 {\tiny $\pm 0.4$} & \textbf{74.6} {\tiny $\pm \textbf{0.3}$} \\
Cars & 79.3 {\tiny $\pm 0.1$} & 79.4 {\tiny $\pm 0.1$} & \textbf{81.5} {\tiny $\pm \textbf{0.2}$} \\
Letters & 82.1 {\tiny $\pm 0.7$} & 82.3 {\tiny $\pm 0.9$} & \textbf{83.1} {\tiny $\pm \textbf{0.6}$} \\
\midrule
Average acc & 64.4 & \textbf{64.9} & 64.5 \\
\bottomrule
\end{tabular}
%\end{sc}
\end{small}
\end{center}
\caption{Accuracy comparison between NCM, LDA, and Linear head using full-body adaptation (\textbf{A-FB} method). The reported results are based on \textbf{50} shots and averaged over 5 runs (mean±std). A pre-trained EfficientNet-B0 is used as a backbone in all cases.}
\label{table:head_comparison_full_50shots}
%\endgroup
\end{table}

\begin{table}[ht]
%\vspace*{1.5 cm}
%\begingroup
\setlength{\tabcolsep}{6.5pt} % Default value: 6pt %\renewcommand{\arraystretch}{.9}
\begin{center}
\begin{small}
%\begin{sc}
\begin{tabular}{l c c c}
\toprule
 \textbf{Dataset} & \textbf{NCM} & \textbf{LDA} &  \textbf{Linear} \\
 \midrule
Caltech101 & 94.2 {\tiny $\pm 0.5$} & 94.6 {\tiny $\pm 0.3$} & \textbf{94.8} {\tiny $\pm \textbf{0.3}$} \\
CIFAR100 & 84.2 {\tiny $\pm 1.2$} & 84.3 {\tiny $\pm 0.9$} & \textbf{85.0} {\tiny $\pm \textbf{1.1}$} \\
Flowers102 & 90.3 {\tiny $\pm 0.3$} & 90.3 {\tiny $\pm 0.4$} & \textbf{91.4} {\tiny $\pm \textbf{0.5}$} \\
Pets & 89.7 {\tiny $\pm 0.4$} & 89.5 {\tiny $\pm 0.3$} & \textbf{90.0} {\tiny $\pm \textbf{0.4}$} \\
Sun397 & 65.9 {\tiny $\pm 0.3$} & 66.1 {\tiny $\pm 1.0$} & \textbf{66.7} {\tiny $\pm \textbf{0.4}$} \\
SVHN & \textbf{95.6} {\tiny $\pm \textbf{0.2}$} & 95.3 {\tiny $\pm 0.5$} & 95.5 {\tiny $\pm 0.3$} \\
DTD & 67.6 {\tiny $\pm 0.8$} & 68.1 {\tiny $\pm 0.4$} & \textbf{68.5} {\tiny $\pm \textbf{0.5}$} \\
\midrule
EuroSAT & 98.1 {\tiny $\pm 0.2$} & 98.5 {\tiny $\pm 0.3$} & \textbf{98.6} {\tiny $\pm \textbf{0.2}$} \\
Resics45 & 95.3 {\tiny $\pm 0.1$} & 95.5 {\tiny $\pm 0.2$} & \textbf{95.9} {\tiny $\pm \textbf{0.1}$} \\
\midrule
Patch Camelyon & 81.1 {\tiny $\pm 2.2$} & 85.0 {\tiny $\pm 0.7$} & \textbf{86.5} {\tiny $\pm \textbf{0.9}$} \\
Retinopathy & 55.8 {\tiny $\pm 0.9$} & 56.1 {\tiny $\pm 1.4$} & \textbf{57.7} {\tiny $\pm \textbf{0.7}$} \\
\midrule
CLEVR-count & 98.5 {\tiny $\pm 0.3$} & \textbf{98.7} {\tiny $\pm \textbf{0.3}$} & 98.3 {\tiny $\pm 0.3$} \\
CLEVR-dist & 89.0 {\tiny $\pm 0.6$} & 89.0 {\tiny $\pm 0.6$} & \textbf{89.4} {\tiny $\pm \textbf{1.5}$} \\
dSprites-loc & 99.7 {\tiny $\pm 0.3$} & \textbf{99.8} {\tiny $\pm \textbf{0.1}$} & 99.6 {\tiny $\pm 0.4$} \\
dSprites-ori & 89.2 {\tiny $\pm 1.0$} & \textbf{94.0} {\tiny $\pm \textbf{0.7}$} & 93.0 {\tiny $\pm 1.1$} \\
SmallNORB-azi & \textbf{29.8} {\tiny $\pm \textbf{1.0}$} & 28.7 {\tiny $\pm 0.6$} & 28.9 {\tiny $\pm 0.8$} \\
SmallNORB-elev & 74.3 {\tiny $\pm 4.3$} & \textbf{81.8} {\tiny $\pm \textbf{3.1}$} & 77.2 {\tiny $\pm 4.6$} \\
DMLab & 64.8 {\tiny $\pm 0.6$} & \textbf{65.7} {\tiny $\pm \textbf{0.4}$} & 65.6 {\tiny $\pm 0.7$} \\
KITTI-dist & 78.2 {\tiny $\pm 0.7$} & 82.1 {\tiny $\pm 0.6$} & \textbf{82.3} {\tiny $\pm \textbf{1.1}$} \\
\midrule
FGVC-Aircraft & 76.0 {\tiny $\pm 0.5$} & 75.8 {\tiny $\pm 0.7$} & \textbf{76.7} {\tiny $\pm \textbf{0.8}$} \\
Cars & 79.1 {\tiny $\pm 0.1$} & 78.9 {\tiny $\pm 0.2$} & \textbf{81.3} {\tiny $\pm \textbf{0.2}$} \\
Letters & 86.0 {\tiny $\pm 0.5$} & 85.7 {\tiny $\pm 0.5$} & \textbf{87.2} {\tiny $\pm \textbf{0.3}$} \\
\midrule
Average acc & 81.0 & 82.0 & \textbf{82.3} \\
\bottomrule
\end{tabular}
%\end{sc}
\end{small}
\end{center}
\caption{Accuracy comparison between NCM, LDA, and Linear head using full-body adaptation (\textbf{A-FB} method). The reported results are based on the full training dataset and averaged over 5 runs (mean±std). A pre-trained EfficientNet-B0 is used as a backbone in all cases.}
\label{table:head_comparison_full_alldata}
%\endgroup
\end{table}

%%%%%%%%%%%%%%%%%%%%%%%%%
% Tables for Resnet18 %%%
%%%%%%%%%%%%%%%%%%%%%%%%%

\begin{table}[ht]
%\vspace*{1.5 cm}
%\begingroup

\setlength{\tabcolsep}{.2pt} % Default value: 6pt
\begin{center}
\begin{footnotesize}
%\begin{sc}
\begin{tabular}{l c c c c}
\toprule
 \textbf{Dataset} & \textbf{NA(RN)} & \textbf{FiLM(RN)} &  \textbf{NA(EN)} &  \textbf{FiLM(EN)} \\
 \midrule
Caltech101 & 80.9 {\tiny $\pm 0.7$} & 81.1 {\tiny $\pm 0.7$} & 88.2 {\tiny $\pm 0.8$} & 89.0 {\tiny $\pm 0.6$} \\
CIFAR100 & 40.4 {\tiny $\pm 0.5$} & 42.2 {\tiny $\pm 0.9$} & 42.7 {\tiny $\pm 1.6$} & 51.8 {\tiny $\pm 1.3$} \\
Flowers102 & 72.6 {\tiny $\pm 0.8$} & 79.4 {\tiny $\pm 1.4$} & 76.1 {\tiny $\pm 0.4$} & 85.0 {\tiny $\pm 0.8$} \\
Pets & 76.9 {\tiny $\pm 1.2$} & 74.5 {\tiny $\pm 1.4$} & 82.4 {\tiny $\pm 1.6$} & 81.8 {\tiny $\pm 1.7$} \\
Sun397 & 35.1 {\tiny $\pm 0.9$} & 29.1 {\tiny $\pm 0.7$} & 41.9 {\tiny $\pm 1.0$} & 40.9 {\tiny $\pm 0.7$} \\
SVHN & 20.9 {\tiny $\pm 1.3$} & 28.8 {\tiny $\pm 2.7$} & 16.5 {\tiny $\pm 1.1$} & 31.7 {\tiny $\pm 3.7$} \\
DTD & 43.2 {\tiny $\pm 1.7$} & 42.2 {\tiny $\pm 0.7$} & 48.9 {\tiny $\pm 1.7$} & 50.2 {\tiny $\pm 0.9$} \\
\midrule
EuroSAT & 75.1 {\tiny $\pm 1.9$} & 79.3 {\tiny $\pm 1.2$} & 76.3 {\tiny $\pm 1.8$} & 78.1 {\tiny $\pm 1.2$} \\
Resics45 & 56.8 {\tiny $\pm 1.3$} & 57.0 {\tiny $\pm 1.2$} & 58.8 {\tiny $\pm 1.4$} & 64.7 {\tiny $\pm 1.2$} \\
\midrule
Patch Camelyon & 62.4 {\tiny $\pm 5.5$} & 64.6 {\tiny $\pm 7.2$} & 59.8 {\tiny $\pm 7.2$} & 64.9 {\tiny $\pm 6.6$} \\
Retinopathy & 23.0 {\tiny $\pm 2.5$} & 23.6 {\tiny $\pm 1.7$} & 25.6 {\tiny $\pm 1.6$} & 26.0 {\tiny $\pm 2.0$} \\
\midrule
CLEVR-count & 21.6 {\tiny $\pm 1.7$} & 23.0 {\tiny $\pm 1.3$} & 23.1 {\tiny $\pm 1.1$} & 23.4 {\tiny $\pm 1.4$} \\
CLEVR-dist & 22.9 {\tiny $\pm 1.4$} & 24.4 {\tiny $\pm 1.3$} & 24.5 {\tiny $\pm 2.3$} & 23.1 {\tiny $\pm 1.1$} \\
dSprites-loc & 13.0 {\tiny $\pm 1.0$} & 15.9 {\tiny $\pm 1.0$} & 8.5 {\tiny $\pm 0.6$} & 19.8 {\tiny $\pm 2.0$} \\
dSprites-ori & 14.4 {\tiny $\pm 0.6$} & 22.9 {\tiny $\pm 0.8$} & 16.2 {\tiny $\pm 0.8$} & 26.5 {\tiny $\pm 0.8$} \\
SmallNORB-azi & 9.4 {\tiny $\pm 0.8$} & 9.8 {\tiny $\pm 1.1$} & 9.3 {\tiny $\pm 0.8$} & 10.1 {\tiny $\pm 0.6$} \\
SmallNORB-elev & 15.8 {\tiny $\pm 0.7$} & 15.9 {\tiny $\pm 0.7$} & 15.1 {\tiny $\pm 0.6$} & 15.4 {\tiny $\pm 0.7$} \\
DMLab & 21.6 {\tiny $\pm 1.5$} & 22.1 {\tiny $\pm 1.8$} & 22.1 {\tiny $\pm 1.3$} & 23.3 {\tiny $\pm 1.9$} \\
KITTI-dist & 54.3 {\tiny $\pm 2.8$} & 54.0 {\tiny $\pm 2.5$} & 51.4 {\tiny $\pm 2.7$} & 52.7 {\tiny $\pm 3.5$} \\
\midrule
FGVC-Aircraft & 19.1 {\tiny $\pm 0.9$} & 20.3 {\tiny $\pm 0.7$} & 22.1 {\tiny $\pm 1.0$} & 32.6 {\tiny $\pm 0.9$} \\
Cars & 14.8 {\tiny $\pm 0.5$} & 13.9 {\tiny $\pm 0.3$} & 22.6 {\tiny $\pm 0.6$} & 28.1 {\tiny $\pm 0.4$} \\
Letters & 32.4 {\tiny $\pm 1.9$} & 45.5 {\tiny $\pm 2.4$} & 36.1 {\tiny $\pm 2.1$} & 55.9 {\tiny $\pm 2.5$} \\
\midrule
Average acc & 37.6 & 39.5 & 39.5 & 44.3 \\
\bottomrule
\end{tabular}
%\end{sc}
\end{footnotesize}
\end{center}
\caption{Accuracy comparison between NA and FiLM methods in offline mode using either a pre-trained ResNet-18 (RN) or a pre-trained EfficientNet-B0 (EN) backbone. We use an LDA head. The reported results are based on \textbf{5} shots. Results are averaged over 5 runs (mean±std).}
\label{table:resnet18_vs_effnet_film_5shots}
%\endgroup
\end{table}

\begin{table}[ht]
%\vspace*{1.5 cm}
%\begingroup

\setlength{\tabcolsep}{.2pt} % Default value: 6pt
\begin{center}
\begin{footnotesize}
%\begin{sc}
\begin{tabular}{l c c c c}
\toprule
 \textbf{Dataset} & \textbf{NA(RN)} & \textbf{FiLM(RN)} &  \textbf{NA(EN)} &  \textbf{FiLM(EN)} \\
 \midrule
Caltech101 & 85.0 {\tiny $\pm 0.6$} & 86.6 {\tiny $\pm 0.6$} & 90.0 {\tiny $\pm 0.8$} & 91.5 {\tiny $\pm 0.4$} \\
CIFAR100 & 48.5 {\tiny $\pm 0.4$} & 52.2 {\tiny $\pm 0.8$} & 50.1 {\tiny $\pm 1.5$} & 62.8 {\tiny $\pm 1.1$} \\
Flowers102 & 81.2 {\tiny $\pm 0.7$} & 87.1 {\tiny $\pm 0.2$} & 83.9 {\tiny $\pm 0.3$} & 91.1 {\tiny $\pm 0.3$} \\
Pets & 82.3 {\tiny $\pm 0.7$} & 80.5 {\tiny $\pm 1.1$} & 86.4 {\tiny $\pm 0.7$} & 86.3 {\tiny $\pm 0.8$} \\
Sun397 & 42.8 {\tiny $\pm 0.9$} & 36.3 {\tiny $\pm 1.0$} & 49.0 {\tiny $\pm 1.3$} & 49.3 {\tiny $\pm 0.6$} \\
SVHN & 24.6 {\tiny $\pm 1.7$} & 35.8 {\tiny $\pm 4.1$} & 19.4 {\tiny $\pm 2.9$} & 45.3 {\tiny $\pm 3.0$} \\
DTD & 51.9 {\tiny $\pm 0.9$} & 50.5 {\tiny $\pm 0.6$} & 55.6 {\tiny $\pm 0.6$} & 59.1 {\tiny $\pm 0.8$} \\
\midrule
EuroSAT & 82.0 {\tiny $\pm 0.7$} & 85.4 {\tiny $\pm 0.9$} & 82.1 {\tiny $\pm 0.9$} & 84.4 {\tiny $\pm 1.6$} \\
Resics45 & 64.8 {\tiny $\pm 1.5$} & 67.2 {\tiny $\pm 2.0$} & 65.5 {\tiny $\pm 1.1$} & 73.0 {\tiny $\pm 1.3$} \\
\midrule
Patch Camelyon & 66.9 {\tiny $\pm 3.7$} & 68.8 {\tiny $\pm 3.7$} & 66.5 {\tiny $\pm 3.7$} & 68.5 {\tiny $\pm 6.1$} \\
Retinopathy & 25.5 {\tiny $\pm 1.3$} & 25.8 {\tiny $\pm 3.7$} & 27.1 {\tiny $\pm 2.2$} & 26.9 {\tiny $\pm 1.1$} \\
\midrule
CLEVR-count & 23.8 {\tiny $\pm 0.7$} & 25.6 {\tiny $\pm 2.0$} & 25.7 {\tiny $\pm 0.6$} & 27.6 {\tiny $\pm 1.6$} \\
CLEVR-dist & 24.9 {\tiny $\pm 0.7$} & 27.2 {\tiny $\pm 1.2$} & 26.3 {\tiny $\pm 1.1$} & 26.2 {\tiny $\pm 1.3$} \\
dSprites-loc & 14.7 {\tiny $\pm 0.4$} & 25.4 {\tiny $\pm 1.5$} & 8.7 {\tiny $\pm 0.3$} & 26.1 {\tiny $\pm 10.6$} \\
dSprites-ori & 16.6 {\tiny $\pm 0.9$} & 29.7 {\tiny $\pm 1.3$} & 18.2 {\tiny $\pm 0.7$} & 34.0 {\tiny $\pm 2.6$} \\
SmallNORB-azi & 10.4 {\tiny $\pm 0.9$} & 12.2 {\tiny $\pm 1.1$} & 9.5 {\tiny $\pm 1.1$} & 11.7 {\tiny $\pm 1.2$} \\
SmallNORB-elev & 16.9 {\tiny $\pm 0.8$} & 17.2 {\tiny $\pm 1.1$} & 16.5 {\tiny $\pm 1.1$} & 16.3 {\tiny $\pm 0.4$} \\
DMLab & 24.6 {\tiny $\pm 1.8$} & 25.6 {\tiny $\pm 1.4$} & 25.7 {\tiny $\pm 1.2$} & 26.6 {\tiny $\pm 1.5$} \\
KITTI-dist & 53.0 {\tiny $\pm 2.0$} & 55.9 {\tiny $\pm 3.5$} & 52.9 {\tiny $\pm 1.5$} & 55.4 {\tiny $\pm 3.7$} \\
\midrule
FGVC-Aircraft & 25.9 {\tiny $\pm 0.8$} & 29.5 {\tiny $\pm 0.8$} & 28.5 {\tiny $\pm 0.4$} & 43.3 {\tiny $\pm 1.1$} \\
Cars & 21.5 {\tiny $\pm 0.5$} & 24.0 {\tiny $\pm 0.2$} & 30.4 {\tiny $\pm 0.5$} & 43.1 {\tiny $\pm 1.0$} \\
Letters & 41.5 {\tiny $\pm 1.2$} & 62.7 {\tiny $\pm 1.9$} & 45.6 {\tiny $\pm 0.8$} & 67.5 {\tiny $\pm 1.3$} \\
\midrule
Average acc & 42.2 & 46.0 & 43.8 & 50.7 \\
\bottomrule
\end{tabular}
%\end{sc}
\end{footnotesize}
\end{center}
\caption{Accuracy comparison between NA and FiLM methods in offline mode using either a pre-trained ResNet-18 (RN) or a pre-trained EfficientNet-B0 (EN) backbone. We use an LDA head. The reported results are based on \textbf{10} shots. Results are averaged over 5 runs (mean±std).}
\label{table:resnet18_vs_effnet_film_10shots}
%\endgroup
\end{table}

\begin{table}[ht]
%\vspace*{1.5 cm}
%\begingroup

\setlength{\tabcolsep}{.2pt} % Default value: 6pt
\begin{center}

\begin{footnotesize}
%\begin{sc}
\begin{tabular}{l c c c c}
\toprule
 \textbf{Dataset} & \textbf{NA(RN)} & \textbf{FiLM(RN)} &  \textbf{NA(EN)} &  \textbf{FiLM(EN)} \\
 \midrule
Caltech101 & 88.0 {\tiny $\pm 0.3$} & 87.7 {\tiny $\pm 0.7$} & 91.9 {\tiny $\pm 0.5$} & 93.8 {\tiny $\pm 0.5$} \\
CIFAR100 & 58.2 {\tiny $\pm 0.9$} & 61.8 {\tiny $\pm 0.7$} & 57.4 {\tiny $\pm 1.0$} & 73.8 {\tiny $\pm 0.9$} \\
Flowers102 & 81.2 {\tiny $\pm 0.7$} & 87.1 {\tiny $\pm 0.2$} & 83.9 {\tiny $\pm 0.3$} & 91.1 {\tiny $\pm 0.3$} \\
Pets & 86.9 {\tiny $\pm 0.4$} & 86.8 {\tiny $\pm 0.6$} & 89.5 {\tiny $\pm 0.4$} & 89.9 {\tiny $\pm 0.7$} \\
Sun397 & 51.4 {\tiny $\pm 1.3$} & 47.8 {\tiny $\pm 0.9$} & 55.9 {\tiny $\pm 1.0$} & 59.7 {\tiny $\pm 0.6$} \\
SVHN & 37.3 {\tiny $\pm 0.8$} & 74.5 {\tiny $\pm 0.9$} & 28.3 {\tiny $\pm 1.1$} & 77.2 {\tiny $\pm 0.8$} \\
DTD & 59.9 {\tiny $\pm 0.0$} & 59.3 {\tiny $\pm 0.6$} & 61.1 {\tiny $\pm 0.0$} & 68.4 {\tiny $\pm 0.2$} \\
\midrule
EuroSAT & 88.3 {\tiny $\pm 0.5$} & 92.6 {\tiny $\pm 0.5$} & 87.7 {\tiny $\pm 0.9$} & 93.0 {\tiny $\pm 0.6$} \\
Resics45 & 73.7 {\tiny $\pm 1.1$} & 77.5 {\tiny $\pm 1.1$} & 73.5 {\tiny $\pm 0.9$} & 83.4 {\tiny $\pm 0.6$} \\
\midrule
Patch Camelyon & 76.3 {\tiny $\pm 0.9$} & 77.4 {\tiny $\pm 1.1$} & 76.2 {\tiny $\pm 1.1$} & 77.9 {\tiny $\pm 2.4$} \\
Retinopathy & 29.2 {\tiny $\pm 2.1$} & 30.4 {\tiny $\pm 1.0$} & 32.9 {\tiny $\pm 2.2$} & 35.2 {\tiny $\pm 1.2$} \\
\midrule
CLEVR-count & 29.5 {\tiny $\pm 1.3$} & 39.9 {\tiny $\pm 1.4$} & 30.1 {\tiny $\pm 1.0$} & 46.6 {\tiny $\pm 1.1$} \\
CLEVR-dist & 31.7 {\tiny $\pm 0.8$} & 45.3 {\tiny $\pm 2.6$} & 32.2 {\tiny $\pm 1.1$} & 38.8 {\tiny $\pm 1.0$} \\
dSprites-loc & 21.8 {\tiny $\pm 0.6$} & 70.2 {\tiny $\pm 2.0$} & 11.9 {\tiny $\pm 0.4$} & 83.7 {\tiny $\pm 5.6$} \\
dSprites-ori & 20.8 {\tiny $\pm 0.4$} & 52.1 {\tiny $\pm 1.5$} & 20.1 {\tiny $\pm 1.1$} & 52.1 {\tiny $\pm 1.3$} \\
SmallNORB-azi & 13.9 {\tiny $\pm 0.5$} & 16.4 {\tiny $\pm 0.7$} & 12.3 {\tiny $\pm 0.8$} & 16.8 {\tiny $\pm 0.8$} \\
SmallNORB-elev & 21.4 {\tiny $\pm 1.0$} & 25.1 {\tiny $\pm 0.6$} & 19.1 {\tiny $\pm 0.7$} & 22.9 {\tiny $\pm 0.5$} \\
DMLab & 29.7 {\tiny $\pm 0.5$} & 31.6 {\tiny $\pm 0.9$} & 30.6 {\tiny $\pm 0.3$} & 34.6 {\tiny $\pm 0.6$} \\
KITTI-dist & 62.0 {\tiny $\pm 3.5$} & 66.0 {\tiny $\pm 2.7$} & 61.4 {\tiny $\pm 2.3$} & 66.8 {\tiny $\pm 3.0$} \\
\midrule
FGVC-Aircraft & 38.9 {\tiny $\pm 0.5$} & 53.1 {\tiny $\pm 0.3$} & 41.0 {\tiny $\pm 0.7$} & 65.1 {\tiny $\pm 0.7$} \\
Cars & 34.4 {\tiny $\pm 0.0$} & 49.5 {\tiny $\pm 0.2$} & 43.3 {\tiny $\pm 0.0$} & 67.9 {\tiny $\pm 0.2$} \\
Letters & 56.1 {\tiny $\pm 1.2$} & 77.7 {\tiny $\pm 1.3$} & 57.6 {\tiny $\pm 0.8$} & 79.7 {\tiny $\pm 0.4$} \\
\midrule
Average acc & 49.6 & 59.5 & 49.9 & 64.5 \\
\bottomrule
\end{tabular}
%\end{sc}
\end{footnotesize}
\end{center}
\caption{Accuracy comparison between NA and FiLM methods in offline mode using either a pre-trained ResNet-18 (RN) or a pre-trained EfficientNet-B0 (EN) backbone. We use an LDA head. The reported results are based on \textbf{50} shots. Results are averaged over 5 runs (mean±std).}
\label{table:resnet18_vs_effnet_film_50shots}
%\endgroup
\end{table}

%%%%%%%%%%%%%%%%%%%%%%%%%%%%%%
% Tables for meta-learning %%%
%%%%%%%%%%%%%%%%%%%%%%%%%%%%%%

\begin{table}[ht]
%\vspace*{1.5 cm}
%\begingroup

\setlength{\tabcolsep}{1.pt} % Default value: 6pt
\begin{center}
\begin{footnotesize}
%\begin{sc}
\begin{tabular}{l c c c c}
\toprule
 \textbf{Dataset} & \textbf{NA} & \textbf{Meta-Learn} &  \textbf{FiLM} &  \textbf{Full-body} \\
 \midrule
Caltech101 & 88.2 {\tiny $\pm 0.8$} & 86.7 {\tiny $\pm 0.9$} & 89.0 {\tiny $\pm 0.6$} & \textbf{89.4} {\tiny $\pm \textbf{0.7}$} \\
CIFAR100 & 42.7 {\tiny $\pm 1.6$} & 42.1 {\tiny $\pm 1.3$} & \textbf{51.8} {\tiny $\pm \textbf{1.3}$} & 49.3 {\tiny $\pm 1.1$} \\
Flowers102 & 76.1 {\tiny $\pm 0.4$} & 78.2 {\tiny $\pm 0.4$} & \textbf{85.0} {\tiny $\pm \textbf{0.8}$} & 81.7 {\tiny $\pm 0.6$} \\
Pets & 82.4 {\tiny $\pm 1.6$} & \textbf{83.8} {\tiny $\pm \textbf{1.0}$} & 81.8 {\tiny $\pm 1.7$} & 80.8 {\tiny $\pm 1.7$} \\
Sun397 & \textbf{41.9} {\tiny $\pm \textbf{1.0}$} & 40.2 {\tiny $\pm 0.6$} & 40.9 {\tiny $\pm 0.7$} & 35.2 {\tiny $\pm 0.5$} \\
SVHN & 16.5 {\tiny $\pm 1.1$} & 27.4 {\tiny $\pm 3.6$} & \textbf{31.7} {\tiny $\pm \textbf{3.7}$} & 19.6 {\tiny $\pm 1.2$} \\
DTD & 48.9 {\tiny $\pm 1.7$} & \textbf{50.5} {\tiny $\pm \textbf{1.7}$} & 50.2 {\tiny $\pm 0.9$} & 49.0 {\tiny $\pm 1.3$} \\
\midrule
EuroSAT & 76.3 {\tiny $\pm 1.8$} & 75.2 {\tiny $\pm 1.5$} & \textbf{78.1} {\tiny $\pm \textbf{1.2}$} & 76.7 {\tiny $\pm 2.9$} \\
Resics45 & 58.8 {\tiny $\pm 1.4$} & 62.7 {\tiny $\pm 1.0$} & \textbf{64.7} {\tiny $\pm \textbf{1.2}$} & 62.3 {\tiny $\pm 2.5$} \\
\midrule
Patch Camelyon & 59.8 {\tiny $\pm 7.2$} & 64.2 {\tiny $\pm 7.3$} & \textbf{64.9} {\tiny $\pm \textbf{6.6}$} & 59.4 {\tiny $\pm 4.9$} \\
Retinopathy & 25.6 {\tiny $\pm 1.6$} & \textbf{26.8} {\tiny $\pm \textbf{3.5}$} & 26.0 {\tiny $\pm 2.0$} & 24.5 {\tiny $\pm 2.5$} \\
\midrule
CLEVR-count & 23.1 {\tiny $\pm 1.1$} & 22.6 {\tiny $\pm 0.8$} & 23.4 {\tiny $\pm 1.4$} & \textbf{23.5} {\tiny $\pm \textbf{3.0}$} \\
CLEVR-dist & 24.5 {\tiny $\pm 2.3$} & 23.8 {\tiny $\pm 1.1$} & 23.1 {\tiny $\pm 1.1$} & \textbf{25.6} {\tiny $\pm \textbf{3.5}$} \\
dSprites-loc & 8.5 {\tiny $\pm 0.6$} & 8.9 {\tiny $\pm 0.5$} & 19.8 {\tiny $\pm 2.0$} & \textbf{27.1} {\tiny $\pm \textbf{1.6}$} \\
dSprites-ori & 16.2 {\tiny $\pm 0.8$} & 19.2 {\tiny $\pm 0.7$} & \textbf{26.5} {\tiny $\pm \textbf{0.8}$} & 19.9 {\tiny $\pm 1.5$} \\
SmallNORB-azi & 9.3 {\tiny $\pm 0.8$} & 8.7 {\tiny $\pm 1.0$} & 10.1 {\tiny $\pm 0.6$} & \textbf{10.4} {\tiny $\pm \textbf{0.8}$} \\
SmallNORB-elev & 15.1 {\tiny $\pm 0.6$} & 15.4 {\tiny $\pm 0.5$} & 15.4 {\tiny $\pm 0.7$} & \textbf{16.2} {\tiny $\pm \textbf{1.1}$} \\
DMLab & 22.1 {\tiny $\pm 1.3$} & \textbf{24.9} {\tiny $\pm \textbf{1.5}$} & 23.3 {\tiny $\pm 1.9$} & 22.1 {\tiny $\pm 1.3$} \\
KITTI-dist & 51.4 {\tiny $\pm 2.7$} & \textbf{55.0} {\tiny $\pm \textbf{1.5}$} & 52.7 {\tiny $\pm 3.5$} & 52.8 {\tiny $\pm 2.5$} \\
\midrule
FGVC-Aircraft & 22.1 {\tiny $\pm 1.0$} & 31.9 {\tiny $\pm 0.6$} & \textbf{32.6} {\tiny $\pm \textbf{0.9}$} & 23.6 {\tiny $\pm 0.8$} \\
Cars & 22.6 {\tiny $\pm 0.6$} & 22.8 {\tiny $\pm 0.4$} & \textbf{28.1} {\tiny $\pm \textbf{0.4}$} & 23.5 {\tiny $\pm 0.3$} \\
Letters & 36.1 {\tiny $\pm 2.1$} & 46.5 {\tiny $\pm 3.2$} & \textbf{55.9} {\tiny $\pm \textbf{2.5}$} & 35.7 {\tiny $\pm 3.3$} \\
\midrule
Average acc & 39.5 & 41.7 & \textbf{44.3} & 41.3 \\
\bottomrule
\end{tabular}
%\end{sc}
\end{footnotesize}
\end{center}
\caption{Accuracy comparison between different adaptation methods in offline mode using a pre-trained EfficientNet-B0 backbone. We use an LDA head. The reported results are based on \textbf{5} shots and averaged over 5 runs (mean±std).}
\label{table:meta_effenet_5shots}
%\endgroup
\end{table}

\begin{table}[ht]
%\vspace*{1.5 cm}
%\begingroup

\setlength{\tabcolsep}{1.1pt} % Default value: 6pt
\begin{center}
\begin{footnotesize}
%\begin{sc}
\begin{tabular}{l c c c c}
\toprule
 \textbf{Dataset} & \textbf{NA} & \textbf{Meta-Learn} &  \textbf{FiLM} &  \textbf{Full-body} \\
 \midrule
Caltech101 & 90.0 {\tiny $\pm 0.8$} & 89.1 {\tiny $\pm 0.3$} & 91.5 {\tiny $\pm 0.4$} & \textbf{91.7} {\tiny $\pm \textbf{0.7}$} \\
CIFAR100 & 50.1 {\tiny $\pm 1.5$} & 50.1 {\tiny $\pm 1.2$} & \textbf{62.8} {\tiny $\pm \textbf{1.1}$} & 59.5 {\tiny $\pm 0.3$} \\
Flowers102 & 83.9 {\tiny $\pm 0.3$} & 84.4 {\tiny $\pm 0.3$} & \textbf{91.1} {\tiny $\pm \textbf{0.3}$} & 90.0 {\tiny $\pm 0.5$} \\
Pets & 86.4 {\tiny $\pm 0.7$} & \textbf{86.8} {\tiny $\pm \textbf{0.2}$} & 86.3 {\tiny $\pm 0.8$} & 85.7 {\tiny $\pm 0.5$} \\
Sun397 & 49.0 {\tiny $\pm 1.3$} & 46.3 {\tiny $\pm 0.9$} & \textbf{49.3} {\tiny $\pm \textbf{0.6}$} & 45.0 {\tiny $\pm 0.4$} \\
SVHN & 19.4 {\tiny $\pm 2.9$} & 33.0 {\tiny $\pm 2.2$} & \textbf{45.3} {\tiny $\pm \textbf{3.0}$} & 27.1 {\tiny $\pm 3.8$} \\
DTD & 55.6 {\tiny $\pm 0.6$} & 57.7 {\tiny $\pm 1.4$} & \textbf{59.1} {\tiny $\pm \textbf{0.8}$} & 56.7 {\tiny $\pm 1.0$} \\
\midrule
EuroSAT & 82.1 {\tiny $\pm 0.9$} & 81.2 {\tiny $\pm 0.7$} & 84.4 {\tiny $\pm 1.6$} & \textbf{84.9} {\tiny $\pm \textbf{1.5}$} \\
Resics45 & 65.5 {\tiny $\pm 1.1$} & 68.4 {\tiny $\pm 1.2$} & \textbf{73.0} {\tiny $\pm \textbf{1.3}$} & 72.8 {\tiny $\pm 1.3$} \\
\midrule
Patch Camelyon & 66.5 {\tiny $\pm 3.7$} & 67.5 {\tiny $\pm 5.5$} & \textbf{68.5} {\tiny $\pm \textbf{6.1}$} & 61.2 {\tiny $\pm 4.5$} \\
Retinopathy & \textbf{27.1} {\tiny $\pm \textbf{2.2}$} & 26.9 {\tiny $\pm 0.4$} & 26.9 {\tiny $\pm 1.1$} & 24.6 {\tiny $\pm 3.3$} \\
\midrule
CLEVR-count & 25.7 {\tiny $\pm 0.6$} & 24.3 {\tiny $\pm 1.1$} & 27.6 {\tiny $\pm 1.6$} & \textbf{27.7} {\tiny $\pm \textbf{2.0}$} \\
CLEVR-dist & 26.3 {\tiny $\pm 1.1$} & 25.5 {\tiny $\pm 0.8$} & 26.2 {\tiny $\pm 1.3$} & \textbf{30.0} {\tiny $\pm \textbf{4.2}$} \\
dSprites-loc & 8.7 {\tiny $\pm 0.3$} & 8.9 {\tiny $\pm 0.4$} & 26.1 {\tiny $\pm 10.6$} & \textbf{44.8} {\tiny $\pm \textbf{1.3}$} \\
dSprites-ori & 18.2 {\tiny $\pm 0.7$} & 20.4 {\tiny $\pm 1.2$} & \textbf{34.0} {\tiny $\pm \textbf{2.6}$} & 29.6 {\tiny $\pm 6.8$} \\
SmallNORB-azi & 9.5 {\tiny $\pm 1.1$} & 10.5 {\tiny $\pm 0.2$} & \textbf{11.7} {\tiny $\pm \textbf{1.2}$} & 11.5 {\tiny $\pm 0.8$} \\
SmallNORB-elev & 16.5 {\tiny $\pm 1.1$} & 15.8 {\tiny $\pm 0.6$} & 16.3 {\tiny $\pm 0.4$} & \textbf{18.3} {\tiny $\pm \textbf{1.5}$} \\
DMLab & 25.7 {\tiny $\pm 1.2$} & \textbf{27.8} {\tiny $\pm \textbf{1.7}$} & 26.6 {\tiny $\pm 1.5$} & 25.6 {\tiny $\pm 1.3$} \\
KITTI-dist & 52.9 {\tiny $\pm 1.5$} & \textbf{56.4} {\tiny $\pm \textbf{1.8}$} & 55.4 {\tiny $\pm 3.7$} & 54.9 {\tiny $\pm 1.3$} \\
\midrule
FGVC-Aircraft & 28.5 {\tiny $\pm 0.4$} & 39.0 {\tiny $\pm 0.8$} & \textbf{43.3} {\tiny $\pm \textbf{1.1}$} & 36.7 {\tiny $\pm 0.6$} \\
Cars & 30.4 {\tiny $\pm 0.5$} & 29.8 {\tiny $\pm 0.1$} & 43.1 {\tiny $\pm 1.0$} & \textbf{43.5} {\tiny $\pm \textbf{0.6}$} \\
Letters & 45.6 {\tiny $\pm 0.8$} & 54.5 {\tiny $\pm 1.5$} & \textbf{67.5} {\tiny $\pm \textbf{1.3}$} & 55.2 {\tiny $\pm 1.3$} \\
\midrule
Average acc & 43.8 & 45.7 & \textbf{50.7} & 49.0 \\
\bottomrule
\end{tabular}
%\end{sc}
\end{footnotesize}
\end{center}
\caption{Accuracy comparison between different adaptation methods in offline mode using a pre-trained EfficientNet-B0 backbone. We use an LDA head. The reported results are based on \textbf{10} shots and averaged over 5 runs (mean±std).}
\label{table:meta_effenet_10shots}
%\endgroup
\end{table}

\begin{table}[ht]
%\vspace*{1.5 cm}
%\begingroup

\setlength{\tabcolsep}{1.1pt} % Default value: 6pt
\begin{center}
\begin{footnotesize}
%\begin{sc}
\begin{tabular}{l c c c c}
\toprule
 \textbf{Dataset} & \textbf{NA} & \textbf{Meta-Learn} &  \textbf{FiLM} &  \textbf{Full-body} \\
 \midrule
Caltech101 & 91.9 {\tiny $\pm 0.5$} & 91.0 {\tiny $\pm 0.4$} & 93.8 {\tiny $\pm 0.5$} & \textbf{93.9} {\tiny $\pm \textbf{0.3}$} \\
CIFAR100 & 57.4 {\tiny $\pm 1.0$} & 58.0 {\tiny $\pm 0.9$} & \textbf{73.8} {\tiny $\pm \textbf{0.9}$} & 73.1 {\tiny $\pm 0.7$} \\
Flowers102 & 83.9 {\tiny $\pm 0.3$} & 84.4 {\tiny $\pm 0.3$} & \textbf{91.1} {\tiny $\pm \textbf{0.3}$} & 90.0 {\tiny $\pm 0.5$} \\
Pets & 89.5 {\tiny $\pm 0.4$} & 89.6 {\tiny $\pm 0.3$} & \textbf{89.9} {\tiny $\pm \textbf{0.7}$} & 89.6 {\tiny $\pm 0.6$} \\
Sun397 & 55.9 {\tiny $\pm 1.0$} & 53.7 {\tiny $\pm 0.8$} & 59.7 {\tiny $\pm 0.6$} & \textbf{60.4} {\tiny $\pm \textbf{0.6}$} \\
SVHN & 28.3 {\tiny $\pm 1.1$} & 47.3 {\tiny $\pm 1.4$} & \textbf{77.2} {\tiny $\pm \textbf{0.8}$} & 64.5 {\tiny $\pm 1.3$} \\
DTD & 61.1 {\tiny $\pm 0.0$} & 63.8 {\tiny $\pm 0.0$} & \textbf{68.4} {\tiny $\pm \textbf{0.2}$} & 64.4 {\tiny $\pm 0.2$} \\
\midrule
EuroSAT & 87.7 {\tiny $\pm 0.9$} & 85.7 {\tiny $\pm 0.6$} & 93.0 {\tiny $\pm 0.6$} & \textbf{94.1} {\tiny $\pm \textbf{0.6}$} \\
Resics45 & 73.5 {\tiny $\pm 0.9$} & 75.5 {\tiny $\pm 1.0$} & 83.4 {\tiny $\pm 0.6$} & \textbf{87.6} {\tiny $\pm \textbf{0.7}$} \\
\midrule
Patch Camelyon & 76.2 {\tiny $\pm 1.1$} & \textbf{78.0} {\tiny $\pm \textbf{1.4}$} & 77.9 {\tiny $\pm 2.4$} & 72.2 {\tiny $\pm 1.8$} \\
Retinopathy & 32.9 {\tiny $\pm 2.2$} & 31.6 {\tiny $\pm 1.2$} & \textbf{35.2} {\tiny $\pm \textbf{1.2}$} & 31.3 {\tiny $\pm 2.9$} \\
\midrule
CLEVR-count & 30.1 {\tiny $\pm 1.0$} & 28.7 {\tiny $\pm 1.1$} & \textbf{46.6} {\tiny $\pm \textbf{1.1}$} & 45.2 {\tiny $\pm 1.1$} \\
CLEVR-dist & 32.2 {\tiny $\pm 1.1$} & 30.5 {\tiny $\pm 1.7$} & 38.8 {\tiny $\pm 1.0$} & \textbf{44.7} {\tiny $\pm \textbf{2.2}$} \\
dSprites-loc & 11.9 {\tiny $\pm 0.4$} & 12.2 {\tiny $\pm 0.5$} & 83.7 {\tiny $\pm 5.6$} & \textbf{87.1} {\tiny $\pm \textbf{2.4}$} \\
dSprites-ori & 20.1 {\tiny $\pm 1.1$} & 24.7 {\tiny $\pm 1.9$} & \textbf{52.1} {\tiny $\pm \textbf{1.3}$} & 44.8 {\tiny $\pm 2.3$} \\
SmallNORB-azi & 12.3 {\tiny $\pm 0.8$} & 12.4 {\tiny $\pm 0.5$} & 16.8 {\tiny $\pm 0.8$} & \textbf{18.3} {\tiny $\pm \textbf{0.7}$} \\
SmallNORB-elev & 19.1 {\tiny $\pm 0.7$} & 18.9 {\tiny $\pm 1.0$} & 22.9 {\tiny $\pm 0.5$} & \textbf{31.3} {\tiny $\pm \textbf{1.7}$} \\
DMLab & 30.6 {\tiny $\pm 0.3$} & 32.6 {\tiny $\pm 0.8$} & \textbf{34.6} {\tiny $\pm \textbf{0.6}$} & 32.7 {\tiny $\pm 1.6$} \\
KITTI-dist & 61.4 {\tiny $\pm 2.3$} & 62.6 {\tiny $\pm 2.0$} & 66.8 {\tiny $\pm 3.0$} & \textbf{66.9} {\tiny $\pm \textbf{2.2}$} \\
\midrule
FGVC-Aircraft & 41.0 {\tiny $\pm 0.7$} & 50.9 {\tiny $\pm 0.7$} & 65.1 {\tiny $\pm 0.7$} & \textbf{73.5} {\tiny $\pm \textbf{0.4}$} \\
Cars & 43.3 {\tiny $\pm 0.0$} & 40.1 {\tiny $\pm 0.0$} & 67.9 {\tiny $\pm 0.2$} & \textbf{79.4} {\tiny $\pm \textbf{0.1}$} \\
Letters & 57.6 {\tiny $\pm 0.8$} & 64.2 {\tiny $\pm 0.6$} & 79.7 {\tiny $\pm 0.4$} & \textbf{82.3} {\tiny $\pm \textbf{0.9}$} \\
\midrule
Average acc & 49.9 & 51.7 & 64.5 & \textbf{64.9} \\
\bottomrule
\end{tabular}
%\end{sc}
\end{footnotesize}
\end{center}
\caption{Accuracy comparison between different adaptation methods in offline mode using a pre-trained EfficientNet-B0 backbone. We use an LDA head. The reported results are based on \textbf{50} shots and averaged over 5 runs (mean±std).}
\label{table:meta_effenet_50shots}
%\endgroup
\end{table}

%%%%%%%%%%%%%%%%%%%%%%%%%%%%%%%%%%%%%%%%%%
% Tables with detailed acc per session %%%
%%%%%%%%%%%%%%%%%%%%%%%%%%%%%%%%%%%%%%%%%%

%%%%%%%%%%%%%%%
% High-shot %%%
%%%%%%%%%%%%%%%

% CIFAR100
\begin{table*}[ht]
%\vspace*{1.5 cm}
%\begingroup

\setlength{\tabcolsep}{9.8pt} % Default value: 6pt
\begin{center}
\begin{normalsize}
%\begin{sc}
\begin{tabular}{l c c c c c c c c c c}
\toprule
\multirow{2}{*}{Method} &  \multicolumn{10}{c}{Accuracy (\%) in each session ($\uparrow$)}  \\
\cmidrule(lr){2-11} 
        &  \textbf{1} & \textbf{2} & \textbf{3} &  \textbf{4} & \textbf{5} & \textbf{6} &  \textbf{7} & \textbf{8} & \textbf{9} & \textbf{10}\\
 \midrule
NA & 93.2 & 87.1 & 81.9 & 80.2 & 76.8 & 74.3 & 72.8 & 70.6 & 70.1 & 68.2\\
\hdashline
E-EWC+SDC & \textbf{97.2} & 70.5 & 63.6 & 46.4 & 40.2 & 40.7 & 38.8 & 35.5 & 33.9 & 32.4 \\
FACT & 96.6 & 48.2 & 32.8 & 24.2 & 19.3 & 16.4 & 13.9 & 12.5 & 11.3 & 10.2 \\
ALICE & 96.6  &  80.2  & 73.7 & 69.4  & 64.8  & 61.7 &  58.1 & 55.7 &  54.3 &  52.4 \\
FSA & 96.0 & 86.3 & 80.5 & 77.7 & 74.2 & 70.9 & 68.2 & 65.8 & 64.2 & 62.8 \\
FSA-LL & 96.4 &  84.9 &  79.1 &  75.4 &  71.6 &  68.5 &  66.4 &  64.1 &  62.7 &  60.5 \\
FSA-FiLM & 96.4 & \textbf{90.4} & \textbf{86.8} & \textbf{84.7} & \textbf{82.0} & \textbf{79.8} & \textbf{78.2} & \textbf{76.1} & \textbf{75.7} & \textbf{73.8} \\
\midrule
GDumb-1k & 94.17 & 86.2 & 81.0 & 76.1 & 70.8 & 64.3 & 62.0 & 59.7 & 57.1 & 54.5 \\
GDumb-5k & 97.0 & 91.6 & 88.1 & 85.1 & 81.8 & 77.9 & 75.6 & 73.2 & 71.8 & 69.3 \\
\bottomrule
\end{tabular}
%\end{sc}
\end{normalsize}
\end{center}
\caption{Detailed accuracy for each incremental session on \textbf{CIFAR100} under the \textbf{high-shot CIL} setting. The best results across all methods per session are in bold while the best results across the no-memory methods are underlined. A pre-trained EfficientNet-B0 on Imagenet-1k is used as a backbone for all methods.}
\label{table:high_shot_cifar100}
%\endgroup
\end{table*}

% CORE50
\begin{table*}[ht]
%\vspace*{1.5 cm}
%\begingroup

\setlength{\tabcolsep}{11.8pt} % Default value: 6pt
\begin{center}
\begin{normalsize}
%\begin{sc}
\begin{tabular}{l c c c c c c c c c}
\toprule
\multirow{2}{*}{Method} &  \multicolumn{9}{c}{Accuracy (\%) in each session ($\uparrow$)}  \\
\cmidrule(lr){2-10} 
        &  \textbf{1} & \textbf{2} & \textbf{3} &  \textbf{4} & \textbf{5} & \textbf{6} &  \textbf{7} & \textbf{8} & \textbf{9}\\
 \midrule
NA & 96.3 & 94.5 & 91.6 & 89.9 & 87.7 & 84.8 & 82.0 & 82.8 & 82.6\\
\hdashline
E-EWC+SDC & 98.7 & 89.7 & 72.6 & 65.4 & 43.0 & 40.8 & 35.1 & 26.6 & 21.7 \\
FACT & 98.5 & 66.9 & 50.5 & 40.4 & 33.8 & 28.9 & 25.3 & 23.9 & 22.0 \\
ALICE & 98.1 & 92.2 & 87.0 & 84.4 & 80.4 & 76.4 & 72.6 & 73.4 & 72.8 \\
FSA & 98.0 & 93.6 & 89.8 & 88.8 & 86.8 & 84.4 & 81.7 & 82.3 & 82.8 \\
FSA-LL & 97.8 & 92.1 & 87.3 & 85.6 & 83.6 & 81.4 & 78.5 & 78.6 & 79.0 \\
FSA-FiLM & \textbf{98.5} & \textbf{96.0} & \textbf{92.6} & \textbf{90.6} & \textbf{89.5} & \textbf{87.3} & \textbf{85.0} & \textbf{85.6} & \textbf{85.4} \\
\midrule
GDumb-1k & 96.9 &  94.3 &  91.9 &  90.6 &  88.2 & 85.4 & 80.7 & 82.3 & 82.4 \\
GDumb-5k & 97.5 & 96.7 & 94.6 & 92.7 & 91.8 & 90.3 & 89.0 & 90.0 & 90.0 \\
\bottomrule
\end{tabular}
%\end{sc}
\end{normalsize}
\end{center}
\caption{Detailed accuracy for each incremental session on \textbf{CORE50} under the \textbf{high-shot CIL} setting. The best results across all methods per session are in bold while the best results across the no-memory methods are underlined. A pre-trained EfficientNet-B0 on Imagenet-1k is used as a backbone for all methods.}
\label{table:high_shot_core50}
%\endgroup
\end{table*}

% SVHN
\begin{table*}[ht]
%\vspace*{1.5 cm}
%\begingroup

\setlength{\tabcolsep}{12.8pt} % Default value: 6pt
\begin{center}
\begin{normalsize}
%\begin{sc}
\begin{tabular}{l c c c c c }
\toprule
\multirow{2}{*}{Method} &  \multicolumn{5}{c}{Accuracy (\%) in each session ($\uparrow$)}  \\
\cmidrule(lr){2-6} 
        &  \textbf{1} & \textbf{2} & \textbf{3} &  \textbf{4} & \textbf{5} \\
 \midrule
NA & 81.5 & 61.5 & 48.5 & 42.5 & 39.9\\
\hdashline
E-EWC+SDC & 99.0 & 55.5 & 48.1 & 44.7 & 39.5 \\
FACT & \textbf{99.3} & 63.5 & 47.4 & 38.7 & 33.8 \\
ALICE & 99.3 & 73.5 &  56.4 &  49.9 &  46.1 \\
FSA & 97.2 & 86.4 & 78.2 & 73.0 & 71.3 \\
FSA-LL & 96.7 & 82.7 & 72.6 & 67.4 & 64.6 \\
FSA-FiLM & 99.1 & \textbf{89.0} & \textbf{81.7} & \textbf{77.6} & \textbf{75.9} \\
\midrule
GDumb-1k  & 97.4 &  91.7 &  87.3 &  83.8 & 78.3 \\
GDumb-5k & 98.6 & 97.3 & 95.8 & 93.7 & 93.2 \\
\bottomrule
\end{tabular}
%\end{sc}
\end{normalsize}
\end{center}
%\endgroup
\caption{Detailed accuracy for each incremental session on \textbf{SVHN} under the \textbf{high-shot CIL} setting. The best results across all methods per session are in bold while the best results across the no-memory methods are underlined. A pre-trained EfficientNet-B0 on Imagenet-1k is used as a backbone for all methods.}
\label{table:high_shot_svhn}
\end{table*}

% dSprites-loc
\begin{table*}[ht]
%\vspace*{1.5 cm}
%\begingroup

\setlength{\tabcolsep}{12.8pt} % Default value: 6pt
\begin{center}
\begin{normalsize}
%\begin{sc}
\begin{tabular}{l c c c c c c c }
\toprule
\multirow{2}{*}{Method} &  \multicolumn{5}{c}{Accuracy (\%) in each session ($\uparrow$)}  \\
\cmidrule(lr){2-8} 
        &  \textbf{1} & \textbf{2} & \textbf{3} &  \textbf{4} & \textbf{5} & \textbf{6} & \textbf{7} \\
 \midrule
NA & 44.9 & 37.4 & 31.6 & 27.1 & 23.8 & 21.5 & 20.6 \\
\hdashline
E-EWC+SDC & 99.5 & 58.1 & 29.9 & 33.0 & 19.4 & 21.7 & 18.6 \\
FACT & \textbf{100.0} &  16.6 &  12.7 &  10.1 &  8.4 &  7.2 &  6.4 \\
ALICE & \textbf{100.0} & 92.6 & 77.6 &  69.6 & 65.6 &  69.8 &  68.3 \\
FSA & \textbf{100.0} & \textbf{95.4} & \textbf{92.3} & \textbf{87.7} & \textbf{89.9} &  \textbf{90.7} & \textbf{91.5} \\
FSA-LL & 99.8 &  94.0 &  93.9 &  90.5 &  91.5 &  91.4 &  91.3 \\
FSA-FiLM & 99.6 &  89.6 &  84.6 &  78.7 &  77.5 &   77.0 &  76.9 \\
\midrule
GDumb-1k  & 91.2 &  85.5 &  85.6  & 83.8 &  76.3 &  78.1 &   79.5 \\
GDumb-5k & 99.4 &  99.5 &  99.6 &  98.5 &  99.4 &  98.4 &  99.4 \\
\bottomrule
\end{tabular}
%\end{sc}
\end{normalsize}
\end{center}
%\endgroup
\caption{Detailed accuracy for each incremental session on \textbf{dSprites-loc} under the \textbf{high-shot CIL} setting. The best results across all methods per session are in bold while the best results across the no-memory methods are underlined. A pre-trained EfficientNet-B0 on Imagenet-1k is used as a backbone for all methods.}
\label{table:high_shot_dspritesloc}
\end{table*}

% FGVC-Airicraft
\begin{table*}[ht]
%\vspace*{1.5 cm}
%\begingroup

\setlength{\tabcolsep}{9.8pt} % Default value: 6pt
\begin{center}
\begin{normalsize}
%\begin{sc}
\begin{tabular}{l c c c c c c c c c c}
\toprule
\multirow{2}{*}{Method} &  \multicolumn{10}{c}{Accuracy (\%) in each session ($\uparrow$)}  \\
\cmidrule(lr){2-11} 
        &  \textbf{1} & \textbf{2} & \textbf{3} &  \textbf{4} & \textbf{5} & \textbf{6} &  \textbf{7} & \textbf{8} & \textbf{9} & \textbf{10}\\
 \midrule
NA & 42.0 & 38.0 & 29.9 & 37.0 & 43.9 & 41.3 & 40.7 & 42.4 & 41.2 & 41.2\\
\hdashline
E-EWC-SDC & 58.0 & 35.3 & 27.3 & 27.7 & 27.4 & 28.2 & 23.5 & 25.4 & 25.7 & 25.6\\
FACT & 58.7 & 30.6 & 24.9 & 21.4 & 19.1 & 19.0 & 17.6 & 16.9 & 15.5 & 14.7 \\
ALICE & \textbf{61.3} & 43.5 & 36.7 &  39.5 & 41.5 & 41.8 & 41.0 & 41.6 & 40.0 & 39.8 \\
FSA & 54.2 & 44.4 & 39.7 & 45.9 & 52.2 & 49.6 & 48.9 & 51.4 & 51.1 & 50.8 \\
FSA-LL & 58.0 & 40.5 & 37.0 & 41.6 & 44.3 & 44.9 & 44.9 & 46.1 & 45.1 & 45.4 \\
FSA-FiLM & 52.9 & \textbf{46.2} & \textbf{44.7} & \textbf{50.3} & \textbf{53.3} & \textbf{55.0} & \textbf{54.5} & \textbf{56.3} & \textbf{55.5} & \textbf{55.9} \\
\midrule
GDumb-1k & 59.8 & 47.3 & 42.6 & 46.5 & 51.8 & 43.7 & 43.4 & 43.0 & 39.2 & 38.4 \\
GDumb-5k & 58.6 & 43.0 & 36.9 & 40.1 & 41.0 & 36.1 & 30.3 & 30.3 & 29.5 & 25.3 \\
\bottomrule
\end{tabular}
%\end{sc}
\end{normalsize}   
\end{center}
\caption{Detailed accuracy for each incremental session on \textbf{FGVC-Aircraft} under the \textbf{high-shot CIL} setting. The best results across all methods per session are in bold while the best results across the no-memory methods are underlined. A pre-trained EfficientNet-B0 on Imagenet-1k is used as a backbone for all methods.}
\label{table:high_shot_fgvcaircraft}
%\endgroup
\end{table*}

% Stanford Cars
\begin{table*}[ht]
%\vspace*{1.5 cm}
%\begingroup

\setlength{\tabcolsep}{9.8pt} % Default value: 6pt
\begin{center}
\begin{normalsize}
%\begin{sc}
\begin{tabular}{l c c c c c c c c c c}
\toprule
\multirow{2}{*}{Method} &  \multicolumn{10}{c}{Accuracy (\%) in each session ($\uparrow$)}  \\
\cmidrule(lr){2-11} 
        &  \textbf{1} & \textbf{2} & \textbf{3} &  \textbf{4} & \textbf{5} & \textbf{6} &  \textbf{7} & \textbf{8} & \textbf{9} & \textbf{10}\\
 \midrule
NA & 72.7 & 49.9 & 49.9 & 46.4 & 47.1 & 47.2 & 44.5 & 44.7 & 44.6 & 43.3 \\
\hdashline
E-EWC+SDC & 80.2 & 47.3 & 38.7 & 37.2 & 34.8 & 33.5 & 31.0 & 31.6 & 31.2 & 30.0 \\
FACT & 79.8 & 2.9 & 1.9 & 1.4 & 1.1 & 0.9 & 0.8 & 0.7 & 0.6 & 0.6 \\
ALICE & 82.7 & 52.9 & 49.6 & 45.1 & 42.8 & 41.1 & 39.0 & 38.8 & 38.1 & 36.4 \\
FSA & 79.3 & 55.0 & 55.9 & 54.0 & 53.3 & 52.7 & 51.5 & 51.5 & 51.4 & 50.3 \\
FSA-LL & \textbf{81.2} & 49.3 & 50.5 & 49.9 & 49.0 & 46.9 & 46.8 & 46.9 & 46.6 & 45.7 \\
FSA-FiLM & 80.1 & \textbf{59.5} & \textbf{60.0} & \textbf{59.4} & \textbf{58.9} & \textbf{58.2} & \textbf{56.8} & \textbf{57.3} & \textbf{56.7} & \textbf{55.9} \\
\midrule
GDumb-1k & 81.5 & 59.9 & 54.0 & 47.3 & 40.5 & 33.9 & 32.7 & 27.6 & 22.8 & 18.1 \\
GDumb-5k & 82.1 & 65.8 & 59.0 & 51.8 & 46.7 & 38.0 & 37.6 & 32.3 & 28.7 & 24.2 \\
\bottomrule
\end{tabular}
%\end{sc}
\end{normalsize}    
\end{center}
\caption{Detailed accuracy for each incremental session on \textbf{Cars} under the \textbf{high-shot CIL} setting. The best results across all methods per session are in bold while the best results across the no-memory methods are underlined. A pre-trained EfficientNet-B0 on Imagenet-1k is used as a backbone for all methods.}
\label{table:high_shot_cars}
%\endgroup
\end{table*}

% Letters
\begin{table*}[ht]
%\vspace*{1.5 cm}
%\begingroup

\setlength{\tabcolsep}{8.8pt} % Default value: 6pt
\begin{center}
\begin{normalsize}
%\begin{sc}
\begin{tabular}{l c c c c c c c c c c c}
\toprule
\multirow{2}{*}{Method} &  \multicolumn{10}{c}{Accuracy (\%) in each session ($\uparrow$)}  \\
\cmidrule(lr){2-11} 
        &  \textbf{1} & \textbf{2} & \textbf{3} &  \textbf{4} & \textbf{5} & \textbf{6} &  \textbf{7} & \textbf{8} & \textbf{9} & \textbf{10} & \textbf{11}\\
 \midrule
NA & 90.12 & 84.3 & 82.4 & 80.1 & 78.7 & 78.0 & 76.0 & 75.3 & 72.8 & 71.6 & 68.4 \\
\hdashline
E-EWC+SDC & 99.9 & 83.4 & 64.5 & 59.9 & 47.8 & 54.1 & 42.6 & 40.4 & 31.1 & 30.0 & 33.6 \\
FACT & 99.9 & 69.9 & 53.9 & 43.4 & 36.3 & 32.7 & 29.3 & 27.0 & 24.4 & 22.4 & 20.9 \\
ALICE & \textbf{99.9} & 96.1 & 93.4 & 89.5 & 88.7 & 87.9 & 85.8 & 83.7 & 81.1 & 79.3 & 75.7 \\
FSA & 99.8 &  96.4 & 94.6 &  91.3 & 90.3  & 89.6 & 87.9 & 86.3 & 83.4 & 82.0 & 78.4 \\
FSA-LL & 99.8 & 95.9 & 94.0 & 90.4 & 89.0 & 88.3 & 86.3 & 85.3 & 82.4 & 81.0 & 77.2  \\
FSA-FiLM & 99.6 & \textbf{96.0} & \textbf{94.4} &  \textbf{92.0}  & \textbf{91.1} & \textbf{90.6} & \textbf{88.5} & \textbf{87.7} &  \textbf{85.0} &  \textbf{83.4} & \textbf{79.7} \\
\midrule
GDumb-1k & 96.0 & 92.2 & 89.4 & 86.7 & 85.7 & 83.9 &  81.1 &  80.3 & 76.2 & 75.2 & 70.1 \\
GDumb-5k & 99.4 & 98.3 & 97.2 & 95.2 & 94.6 & 94.3 & 92.2 & 91.3 & 88.5 & 86.9 & 82.6\\
\bottomrule
\end{tabular}
%\end{sc}
\end{normalsize}    
\end{center}
\caption{Detailed accuracy for each incremental session on \textbf{Letters} under the \textbf{high-shot CIL} setting. The best results across all methods per session are in bold while the best results across the no-memory methods are underlined. A pre-trained EfficientNet-B0 on Imagenet-1k is used as a backbone for all methods.}
\label{table:high_shot_letters}
%\endgroup
\end{table*}

%%%%%%%%%%%%%%%
% Few-shot+ %%%
%%%%%%%%%%%%%%%

% CIFAR100
\begin{table*}[ht]
%\vspace*{1.5 cm}
%\begingroup

\setlength{\tabcolsep}{8.5pt} % Default value: 6pt
\begin{center}
\begin{normalsize}

%\begin{sc}
\begin{tabular}{l c c c c c c c c c c}
\toprule
\multirow{2}{*}{Method} & \multirow{2}{*}{Backbone} & \multicolumn{9}{c}{Accuracy (\%) in each session ($\uparrow$)}  \\
\cmidrule(lr){3-11} 
     &   &  \textbf{1} & \textbf{2} & \textbf{3} &  \textbf{4} & \textbf{5} & \textbf{6} &  \textbf{7} & \textbf{8} & \textbf{9}\\
 \midrule
Decoupled-Cos* &  \multirow{4}{*}{RN-20} & 74.6 & 67.4 & 63.6 & 59.6 & 56.1 & 53.8 & 51.7 & 49.7 & 47.7 \\
CEC* &  & 73.1 & 68.9 & 65.3 & 61.2 & 58.1 & 55.6 & 53.2 & 51.3 & 49.1 \\
FACT* & & 74.6 & 72.1 & 67.6 & 63.5 & 61.4 & 58.4 & 56.3 & 54.2 & 52.1 \\
FSA & & 75.1 & 71.2 & 67.5 & 63.3 & 60.0 & 57.6 & 55.5 & 54.2 & 52.0 \\
\hdashline
NA &  \multirow{5}{*}{RN-18} & 68.9 & 65.4 & 62.4 & 58.7 & 57.2 & 54.7 & 53.3 & 51.9 & 50.4 \\
FACT &  & 75.8 & 71.0 & 66.3 & 62.5 & 59.1 & 56.3 & 54.1 & 51.8 & 49.5 \\
ALICE$^{\dagger}$ & & 79.0 & 70.5 & 67.1 & 63.4 & 61.2 & 59.2 & 58.1 & 56.3 & 54.1 \\
FSA-FiLM  &  & 73.0 & 69.7 & 66.3 & 63.2 & 61.9 & 59.3 & 58.3 & 57.2 & 55.2 \\
FSA &  & 82.0 & 78.2 & 74.8 & 70.22 & 68.7 & 66.2 & 65.3 & 63.8 & 61.4 \\
\hdashline
NA & \multirow{5}{*}{EN-B0} & 74.4 & 70.4 & 67.4 & 63.4 & 62.4 & 59.8 & 58.4 & 56.9 & 55.2 \\
FACT & & 86.4 & 80.6 & 75.6 & 71.1 & 67.6 & 64.4 & 61.8 & 59.2 & 56.5 \\
ALICE & & 87.7 & 83.3 & 78.7 & 74.4 & 72.1 & 69.6 & 67.4 & 65.4 & 62.7 \\
FSA-FiLM & & 79.6 & 75.6 & 72.9 & 68.8 & 68.2 & 65.4 & 64.9 & 63.9 & 61.8 \\
FSA & & \textbf{87.6} & \textbf{83.5} & \textbf{79.7} & \textbf{75.4} & \textbf{73.8} & \textbf{70.9} & \textbf{70.2} & \textbf{68.8} & \textbf{66.1}  \\
\bottomrule
\end{tabular}
%\end{sc}
\end{normalsize}   
\end{center}
\caption{Detailed accuracy for each incremental session on \textbf{CIFAR100} under the \textbf{few-shot+ CIL} setting. Asterisk (*) indicates that the reported results of a method are from \cite{zhou2022forward} and $\dagger$ that the reported results of a method are from \cite{peng2022few}. We use three different backbones, EfficientNet-B0 (EN-B0) and ResNet-18/20 (RN-18/20); EN-B0 and RN-18 are pre-trained on Imagenet-1k.}
\label{table:few_shot_plus_cifar100}
%\endgroup
\end{table*}

% CUB200
\begin{table*}[ht]
%\vspace*{1.5 cm}
%\begingroup

\setlength{\tabcolsep}{5.8pt} % Default value: 6pt
\begin{center}
\begin{normalsize}
%\begin{sc}
\begin{tabular}{l c c c c c c c c c c c c}
\toprule
\multirow{2}{*}{Method} & \multirow{2}{*}{Backbone} & \multicolumn{11}{c}{Accuracy (\%) in each session ($\uparrow$)}  \\
\cmidrule(lr){3-13} 
     &   &  \textbf{1} & \textbf{2} & \textbf{3} &  \textbf{4} & \textbf{5} & \textbf{6} &  \textbf{7} & \textbf{8} & \textbf{9} & \textbf{10} & \textbf{11} \\
 \midrule
NA &  \multirow{7}{*}{RN-18} & 70.7 &  66.7 & 63.4 & 59.0 & 58.2 & 56.4 & 54.0 & 52.3 & 50.5 & 50.5 & 50.0 \\
Decoupled-Cos* &  & 75.5 & 71.0 & 66.5 & 61.2 & 60.9 & 56.9 & 55.4 & 53.5 & 51.9 & 50.9 & 49.3 \\
CEC* &  & 75.9 & 71.9 & 68.5 & 63.5 & 62.4 & 58.3 & 57.7 & 55.8 & 54.8 & 53.5 & 52.3 \\
FACT* &  & 75.9 & 73.2 & 70.8 & 66.1 & 65.6 & 62.2 & 61.7 & 59.8 & 58.4 & 57.9 & 56.9 \\
ALICE$^{\dagger}$ &  & 77.4&  72.7&  70.6&  67.2&  65.9&  63.4&  62.9&  61.9&  60.5&  60.6&  60.1 \\
FSA-FiLM  &  & 72.7 & 68.2 & 64.9 & 60.8 & 60.2 & 58.1 & 55.4 & 54.8 & 53.5 & 53.4 & 52.7 \\
FSA &  & 76.1 & 72.6 & 69.6 & 65.0 & 64.6 & 62.3 & 61.6 & 59.6 & 58.2 & 58.2 & 57.6  \\
\hdashline
NA & \multirow{5}{*}{EN-B0} & 78.6 & 75.8 & 73.4 & 69.5 & 69.2 & 67.3 & 66.5 & 64.3 & 62.7 & 63.1 & 63.2 \\
FACT & & \textbf{82.0} & \textbf{77.5} & 74.4 & 70.0 & 69.3 & 66.6 & 66.2 & 64.7 & \textbf{64.0} & 63.3 & 62.9 \\
ALICE & & 81.6 & 77.1 & \textbf{75.1} & \textbf{71.9} & \textbf{70.5} & \textbf{67.8} & \textbf{66.8} & \textbf{65.7} & \textbf{64.1} & \textbf{64.0} & \textbf{63.5} \\
FSA-FiLM & & 79.0 & 75.3 & 72.7 & 69.5 & 68.3 & 66.5 & 65.3 & 64.1 & 62.8 & 62.9 & 62.9 \\
FSA & & 80.2 & 77.1 & 74.2 & 69.3 & 69.3 & 66.9 & 66.4 & 64.8 & 63.6 & 63.8 & 63.4 \\
\bottomrule
\end{tabular}
%\end{sc}
\end{normalsize}
    
\end{center}
\caption{Detailed accuracy for each incremental session on \textbf{CUB200} under the \textbf{few-shot+ CIL} setting. Asterisk (*) indicates that the reported results of a method are from \cite{zhou2022forward} and $\dagger$ that the reported results of a method are from \cite{peng2022few}. We use two different backbones, EfficientNet-B0 (EN-B0) and ResNet-18 (RN-18); EN-B0 and RN-18 are pre-trained on Imagenet-1k.}
\label{table:few_shot_plus_cub200}
%\endgroup
\end{table*}

%%%%%%%%%%%%%%
% Few-shot %%%
%%%%%%%%%%%%%%

% CIFAR100
\begin{table*}[ht]
%\vspace*{1.5 cm}
%\begingroup

\setlength{\tabcolsep}{3.2pt} % Default value: 6pt
\begin{center}
\begin{footnotesize}
\begin{tabular}{l c c c c c c c c c}
\toprule
\multirow{2}{*}{Method} & \multicolumn{9}{c}{Accuracy (\%) in each session ($\uparrow$)}  \\
\cmidrule(lr){2-10} 
   &  \textbf{1} & \textbf{2} & \textbf{3} &  \textbf{4} & \textbf{5} & \textbf{6} &  \textbf{7} & \textbf{8} & \textbf{9}\\
 \midrule
NA & 80.2 {\footnotesize $ \pm $ 0.9} & 76.2 {\footnotesize $ \pm $ 0.7} & 72.4 {\footnotesize $ \pm $ 1.5} & 68.7 {\footnotesize $ \pm $ 1.5} & 65.2 {\footnotesize $ \pm $ 1.4} & 63.5 {\footnotesize $ \pm $ 1.3} & 60.8 {\footnotesize $ \pm $ 1.3} & 59.5 {\footnotesize $ \pm $ 1.9} & 57.4 {\footnotesize $ \pm $ 1.0}\\
\hdashline
GDumb & 83.9 {\footnotesize $ \pm $ 1.4} & 80.1 {\footnotesize $ \pm $ 0.5} & 77.6 {\footnotesize $ \pm $ 1.6} & 71.7 {\footnotesize $ \pm $ 2.2} & 67.5 {\footnotesize $ \pm $ 1.8} & 64.0 {\footnotesize $ \pm $ 1.5} & 59.4 {\footnotesize $ \pm $ 1.2} & 58.6 {\footnotesize $ \pm $ 1.9} & 55.7 {\footnotesize $ \pm $ 1.0}\\
\hdashline
FACT & 83.0 {\footnotesize $ \pm $ 1.2} & 54.5 {\footnotesize $ \pm $ 1.4} & 40.7 {\footnotesize $ \pm $ 1.0} & 32.6 {\footnotesize $ \pm $ 1.0} & 27.6 {\footnotesize $ \pm $ 0.9} & 23.8 {\footnotesize $ \pm $ 0.7} & 20.8 {\footnotesize $ \pm $ 0.5} & 18.7 {\footnotesize $ \pm $ 0.4} & 16.8 {\footnotesize $ \pm $ 0.3} \\
FSA & 82.7 {\footnotesize $ \pm $ 2.1} & 75.4 {\footnotesize $ \pm $ 1.7} & 72.5 {\footnotesize $ \pm $ 1.8} & 69.5 {\footnotesize $ \pm $ 1.6} & 66.8 {\footnotesize $ \pm $ 1.8} & 64.5 {\footnotesize $ \pm $ 1.4} & 63.1 {\footnotesize $ \pm $ 1.5} & 62.7 {\footnotesize $ \pm $ 1.5} & 60.3 {\footnotesize $ \pm $ 1.3} \\
FSA-FiLM & \underline{\textbf{89.2} {\footnotesize $ \pm $ \textbf{0.9}}} & \underline{\textbf{85.8} {\footnotesize $ \pm $ \textbf{1.3}}} & \underline{\textbf{84.3} {\footnotesize $ \pm $ \textbf{1.3}}} & \underline{\textbf{81.0} {\footnotesize $ \pm $ \textbf{1.3}}} & \underline{\textbf{77.9} {\footnotesize $ \pm $ \textbf{1.7}}} & \underline{\textbf{75.9} {\footnotesize $ \pm $ \textbf{1.0}}} & \underline{\textbf{74.3} {\footnotesize $ \pm $ \textbf{1.4}}} & \underline{\textbf{74.2} {\footnotesize $ \pm $ \textbf{1.1}}} & \underline{\textbf{70.9} {\footnotesize $ \pm $ \textbf{1.0}}} \\
\bottomrule
\end{tabular}
\end{footnotesize}
\end{center}
%\endgroup

\caption{Detailed accuracy for each incremental session on \textbf{CIFAR100} under the \textbf{few-shot CIL}  setting. GDumb is the only
memory-based method used for comparisons; we use a buffer size equal to the first session’s number of images $N_1$. The best results across
all methods are in bold while the best results across the no-memory methods are underlined. A pre-trained EfficientNet-B0 on Imagenet-1k
is used as a backbone for all methods.}
\label{table:few_shot_cifar100}
\end{table*}

% SVHN
\begin{table*}[ht]
%\vspace*{1.5 cm}
%\begingroup

\setlength{\tabcolsep}{9.5pt} % Default value: 6pt
\begin{center}
\begin{normalsize}
%\begin{sc}
\begin{tabular}{l c c c c c}
\toprule
\multirow{2}{*}{Method} & \multicolumn{5}{c}{Accuracy (\%) in each session ($\uparrow$)}  \\
\cmidrule(lr){2-6} 
   &  \textbf{1} & \textbf{2} & \textbf{3} &  \textbf{4} & \textbf{5} \\
 \midrule
NA & 73.5 {\tiny $ \pm $ 4.6} & 53.3 {\tiny $ \pm $ 3.7} & 37.3 {\tiny $ \pm $ 1.8} & 33.6 {\tiny $ \pm $ 1.5} & 28.3 {\tiny $ \pm $ 1.1} \\
\hdashline
GDumb & 78.2 {\tiny $ \pm $ 4.9} & 46.7 {\tiny $ \pm $ 5.1} & 35.2 {\tiny $ \pm $ 1.6} & 23.3 {\tiny $ \pm $ 3.8} & 21.0 {\tiny $ \pm $ 2.1}\\
\hdashline
FACT & 71.3 {\tiny $ \pm $ 1.0} & 46.6 {\tiny $ \pm $ 3.5} & 34.0 {\tiny $ \pm $ 1.9} & 27.7 {\tiny $ \pm $ 2.5} & 24.1 {\tiny $ \pm $ 2.0} \\
FSA & 70.7 {\tiny $ \pm $ 2.9} & 50.8 {\tiny $ \pm $ 3.9} & 38.4 {\tiny $ \pm $ 3.2} & 35.7 {\tiny $ \pm $ 1.6} & 32.9 {\tiny $ \pm $ 1.0} \\
FSA-FiLM & \underline{\textbf{90.7} {\tiny $ \pm $ \textbf{1.8}}} & \underline{\textbf{70.4} {\tiny $ \pm $ \textbf{1.4}}} & \underline{\textbf{60.5} {\tiny $ \pm $ \textbf{1.5}}} & \underline{\textbf{55.5} {\tiny $ \pm $ \textbf{2.3}}} & \underline{\textbf{51.3} {\tiny $ \pm $ \textbf{2.1}}} \\
\bottomrule
\end{tabular}
%\end{sc}
\end{normalsize}
\end{center}
\caption{Detailed accuracy for each incremental session on \textbf{SVHN} under the \textbf{few-shot CIL}  setting. GDumb is the only
memory-based method used for comparisons; we use a buffer size equal to the first session’s number of images $N_1$. The best results across
all methods are in bold while the best results across the no-memory methods are underlined. A pre-trained EfficientNet-B0 on Imagenet-1k
is used as a backbone for all methods.}
\label{table:few_shot_svhn}
%\endgroup
\end{table*}

% dSprites-loc
\begin{table*}[ht]
%\vspace*{1.5 cm}
%\begingroup

\setlength{\tabcolsep}{7.8pt} % Default value: 6pt
\begin{center}
%\begin{sc}
\begin{tabular}{l c c c c c c c c}
\toprule
\multirow{2}{*}{Method} & \multicolumn{7}{c}{Accuracy (\%) in each session ($\uparrow$)}  \\
\cmidrule(lr){2-8} 
   &  \textbf{1} & \textbf{2} & \textbf{3} &  \textbf{4} & \textbf{5} &  \textbf{6} & \textbf{7} \\
 \midrule
NA & 35.7 {\footnotesize $ \pm $ 1.4} & 26.4 {\footnotesize $ \pm $ 1.7} & 22.1 {\footnotesize $ \pm $ 1.7} & 18.1 {\footnotesize $ \pm $ 1.2} & 15.4 {\footnotesize $ \pm $ 0.8} & 13.5 {\footnotesize $ \pm $ 0.7} & 11.9 {\footnotesize $ \pm $ 0.4}\\
\hdashline
GDumb & 36.4 {\footnotesize $ \pm $ 7.9} & 29.9 {\footnotesize $ \pm $ 6.3} & 20.3 {\footnotesize $ \pm $ 3.6} & 22.1 {\footnotesize $ \pm $ 5.5} & 13.1 {\footnotesize $ \pm $ 2.1} & 11.7 {\footnotesize $ \pm $ 3.2} & 16.4 {\footnotesize $ \pm $ 2.6}\\
\hdashline
FACT & 32.6 {\footnotesize $ \pm $ 1.4} & 22.7 {\footnotesize $ \pm $ 2.2} & 18.4 {\footnotesize $ \pm $ 2.1} & 14.4 {\footnotesize $ \pm $ 1.9} & 12.4 {\footnotesize $ \pm $ 1.5} & 11.9 {\footnotesize $ \pm $ 1.6} & 11.7 {\footnotesize $ \pm $ 1.7} \\
FSA & 57.4 {\footnotesize $ \pm $ 2.3} & 44.8 {\footnotesize $ \pm $ 2.8} & 39.3 {\footnotesize $ \pm $ 1.6} & 34.8 {\footnotesize $ \pm $ 2.1} & 32.9 {\footnotesize $ \pm $ 1.9} & 33.2 {\footnotesize $ \pm $ 2.6} & 33.7 {\footnotesize $ \pm $ 1.7} \\
FSA-FiLM & \underline{\textbf{62.7} {\footnotesize $ \pm $ \textbf{2.1}}} & \underline{\textbf{50.2} {\footnotesize $ \pm $ \textbf{2.4}}} & \underline{\textbf{46.9} {\footnotesize $ \pm $ \textbf{2.3}}} & \underline{\textbf{40.1} {\footnotesize $ \pm $ \textbf{2.2}}} & \underline{\textbf{37.3} {\footnotesize $ \pm $ \textbf{2.5}}} & \underline{\textbf{36.1} {\footnotesize $ \pm $ \textbf{2.2}}} & \underline{\textbf{35.7} {\footnotesize $ \pm $ \textbf{2.1}}} \\
\bottomrule
\end{tabular}
%\end{sc}
\end{center}
\caption{Detailed accuracy for each incremental session on \textbf{dSprites-position} under the \textbf{few-shot CIL}  setting. GDumb is the only
memory-based method used for comparisons; we use a buffer size equal to the first session’s number of images $N_1$. The best results across
all methods are in bold while the best results across the no-memory methods are underlined. A pre-trained EfficientNet-B0 on Imagenet-1k
is used as a backbone for all methods.}
\label{table:few_shot_dsprites}
%\endgroup
\end{table*}

% FGVC-Aircraft
\begin{table*}[ht]
%\vspace*{1.5 cm}
%\begingroup

\setlength{\tabcolsep}{2.7pt} % Default value: 6pt
\begin{center}
\begin{small}
\begin{tabular}{l c c c c c c c c c}
\toprule
\multirow{2}{*}{Method} & \multicolumn{9}{c}{Accuracy (\%) in each session ($\uparrow$)}  \\
\cmidrule(lr){2-10} 
   &  \textbf{1} & \textbf{2} & \textbf{3} &  \textbf{4} & \textbf{5} & \textbf{6} &  \textbf{7} & \textbf{8} & \textbf{9}\\
 \midrule
NA & 35.5 {\footnotesize $ \pm $ 0.9} & 28.7 {\footnotesize $ \pm $ 0.7} & 36.4 {\footnotesize $ \pm $ 0.8} & 42.8 {\footnotesize $ \pm $ 0.4} & 40.0 {\footnotesize $ \pm $ 0.4} & 39.4 {\footnotesize $ \pm $ 0.6} & 41.4 {\footnotesize $ \pm $ 0.7} & 40.3 {\footnotesize $ \pm $ 0.9} & 41.0 {\footnotesize $ \pm $ 0.7}\\
\hdashline
GDumb & \textbf{51.1} {\footnotesize $ \pm $ \textbf{1.7}} & \textbf{45.4} {\footnotesize $ \pm $ \textbf{1.2}} & 46.9 {\footnotesize $ \pm $ 1.8} & 52.2 {\footnotesize $ \pm $ 1.0} & 45.4 {\footnotesize $ \pm $ 1.7} & 42.5 {\footnotesize $ \pm $ 1.6} & 41.8 {\footnotesize $ \pm $ 0.9} & 39.5 {\footnotesize $ \pm $ 1.9} & 38.6 {\footnotesize $ \pm $ 1.0}\\
\hdashline
FACT & 41.4 {\footnotesize $ \pm $ 0.6} & 25.9 {\footnotesize $ \pm $ 0.9} & 19.6 {\footnotesize $ \pm $ 0.4} & 16.3 {\footnotesize $ \pm $ 0.4} & 13.7 {\footnotesize $ \pm $ 0.4} & 11.3 {\footnotesize $ \pm $ 0.5} & 10.4 {\footnotesize $ \pm $ 0.5} & 9.4 {\footnotesize $ \pm $ 0.6} & 8.3 {\footnotesize $ \pm $ 0.6} \\
FSA & 42.9 {\footnotesize $ \pm $ 2.6} & 39.5 {\footnotesize $ \pm $ 2.0} & 45.5 {\footnotesize $ \pm $ 1.5} & 51.6 {\footnotesize $ \pm $ 1.7} & 48.2 {\footnotesize $ \pm $ 1.8} & 47.3 {\footnotesize $ \pm $ 1.4} & 49.8 {\footnotesize $ \pm $ 1.5} & 49.1 {\footnotesize $ \pm $ 1.7} & 50.1 {\footnotesize $ \pm $ 1.5} \\
FSA-FiLM & \underline{46.6 {\footnotesize $ \pm $ 1.9}} & \underline{44.8 {\footnotesize $ \pm $ 1.1}} & \underline{\textbf{49.4} {\footnotesize $ \pm $ \textbf{0.8}}} & \underline{\textbf{52.9} {\footnotesize $ \pm $ \textbf{1.6}}} & \underline{\textbf{54.0} {\footnotesize $ \pm $ \textbf{0.7}}} & \underline{\textbf{53.5} {\footnotesize $ \pm $ \textbf{1.0}}} & \underline{\textbf{55.8} {\footnotesize $ \pm $ \textbf{0.7}}} & \underline{\textbf{55.2} {\footnotesize $ \pm $ \textbf{0.5}}} & \underline{\textbf{55.8} {\footnotesize $ \pm $ \textbf{0.6}}} \\
\bottomrule
\end{tabular}
\end{small}
\end{center}
\caption{Detailed accuracy for each incremental session on \textbf{FGVC-Aircraft} under the \textbf{few-shot CIL}  setting. GDumb is the only
memory-based method used for comparisons; we use a buffer size equal to the first session’s number of images $N_1$. The best results across
all methods are in bold while the best results across the no-memory methods are underlined. A pre-trained EfficientNet-B0 on Imagenet-1k
is used as a backbone for all methods.}
\label{table:few_shot_fgvcaircraft}
%\endgroup
\end{table*}

% Letters
\begin{table*}[ht]
%\vspace*{1.5 cm}
%\begingroup

\setlength{\tabcolsep}{2.1pt} % Default value: 6pt
\begin{center}
\begin{scriptsize}
      
%\begin{sc}
\begin{tabular}{l c c c c c c c c c c c}
\toprule
\multirow{2}{*}{Method} & \multicolumn{11}{c}{Accuracy (\%) in each session ($\uparrow$)}  \\
\cmidrule(lr){2-12} 
   &  \textbf{1} & \textbf{2} & \textbf{3} &  \textbf{4} & \textbf{5} & \textbf{6} &  \textbf{7} & \textbf{8} & \textbf{9} & \textbf{10} & \textbf{11} \\
 \midrule
NA & 82.1 {\tiny $ \pm $ 0.8} & 76.6 {\tiny $ \pm $ 0.9} & 73.0 {\tiny $ \pm $ 1.5} & 69.6 {\tiny $ \pm $ 1.0} & 69.0 {\tiny $ \pm $ 1.3} & 67.6 {\tiny $ \pm $ 1.3} & 65.8 {\tiny $ \pm $ 1.2} & 64.1 {\tiny $ \pm $ 1.0} & 60.4 {\tiny $ \pm $ 0.6} & 59.0 {\tiny $ \pm $ 0.8} & 57.6 {\tiny $ \pm $ 0.8}\\
\hdashline
GDumb & 91.3 {\tiny $ \pm $ 1.6} & \textbf{91.8} {\tiny $ \pm $ \textbf{1.2}} & 80.0 {\tiny $ \pm $ 1.4} & 72.0 {\tiny $ \pm $ 1.9} & 69.2 {\tiny $ \pm $ 3.0} & 63.5 {\tiny $ \pm $ 1.5} & 59.9 {\tiny $ \pm $ 1.1} & 54.5 {\tiny $ \pm $ 0.3} & 48.0 {\tiny $ \pm $ 0.4} & 44.3 {\tiny $ \pm $ 1.9} & 41.2 {\tiny $ \pm $ 1.7}\\
\hdashline
FACT & 84.3 {\tiny $ \pm $ 1.4} & 72.0 {\tiny $ \pm $ 1.2} & 68.0 {\tiny $ \pm $ 2.2} & 63.2 {\tiny $ \pm $ 1.0} & 62.3 {\tiny $ \pm $ 1.0} & 59.5 {\tiny $ \pm $ 1.2} & 58.0 {\tiny $ \pm $ 1.0} & 55.8 {\tiny $ \pm $ 1.0} & 52.8 {\tiny $ \pm $ 0.7} & 51.7 {\tiny $ \pm $ 0.8} & 49.8 {\tiny $ \pm $ 0.8} \\
FSA & 87.0 {\tiny $ \pm $ 1.4} & 79.6 {\tiny $ \pm $ 1.1} & 76.4 {\tiny $ \pm $ 0.9} & 72.7 {\tiny $ \pm $ 0.7} & 73.0 {\tiny $ \pm $ 0.9} & 71.3 {\tiny $ \pm $ 0.4} & 69.7 {\tiny $ \pm $ 0.4} & 68.4 {\tiny $ \pm $ 0.4} & 64.7 {\tiny $ \pm $ 0.5} & 62.9 {\tiny $ \pm $ 0.4} & 62.2 {\tiny $ \pm $ 0.4} \\
FSA-FiLM & \underline{\textbf{94.3} {\tiny $ \pm $ \textbf{0.9}}} & \underline{90.6 {\tiny $ \pm $ 0.3}} & \underline{\textbf{88.6} {\tiny $ \pm $ \textbf{1.0}}} & \underline{\textbf{85.1} {\tiny $ \pm $ \textbf{0.6}}} & \underline{\textbf{84.9} {\tiny $ \pm $ \textbf{0.4}}} & \underline{\textbf{84.0} {\tiny $ \pm $ \textbf{0.4}}} & \underline{\textbf{82.5} {\tiny $ \pm $ \textbf{0.7}}} & \underline{\textbf{81.1} {\tiny $ \pm $ \textbf{0.4}}} & \underline{\textbf{76.8} {\tiny $ \pm $ \textbf{0.4}}} & \underline{\textbf{75.0} {\tiny $ \pm $ \textbf{0.4}}} & \underline{\textbf{73.4} {\tiny $ \pm $ \textbf{0.4}}} \\
\bottomrule
\end{tabular}
%\end{sc}
\end{scriptsize}   
\end{center}

\caption{Detailed accuracy for each incremental session on \textbf{Letters} under the \textbf{few-shot CIL}  setting. GDumb is the only
memory-based method used for comparisons; we use a buffer size equal to the first session’s number of images $N_1$. The best results across
all methods are in bold while the best results across the no-memory methods are underlined. A pre-trained EfficientNet-B0 on Imagenet-1k
is used as a backbone for all methods.}
\label{table:few_shot_letters}
%\endgroup
\end{table*}

% DomainNet
\begin{table*}[ht]
%\vspace*{1.5 cm}
%\begingroup

\setlength{\tabcolsep}{3.2pt} % Default value: 6pt
\begin{center}
\begin{small}
\begin{tabular}{l c c c c c c c c c}
\toprule
\multirow{2}{*}{Method} & \multicolumn{9}{c}{Accuracy (\%) in each session ($\uparrow$)}  \\
\cmidrule(lr){2-10} 
   &  \textbf{1} & \textbf{2} & \textbf{3} &  \textbf{4} & \textbf{5} & \textbf{6} &  \textbf{7} & \textbf{8} & \textbf{9}\\
 \midrule
NA & 83.3 {\scriptsize $ \pm $ 1.0} & 62.6 {\scriptsize $ \pm $ 0.5} & 60.9 {\scriptsize $ \pm $ 1.0} & 61.5 {\scriptsize $ \pm $ 0.3} & 68.3 {\scriptsize $ \pm $ 1.2} & 71.1 {\scriptsize $ \pm $ 0.6} & 72.0 {\scriptsize $ \pm $ 0.7} & 70.8 {\scriptsize $ \pm $ 1.0} & 69.0 {\scriptsize $ \pm $ 0.3}\\
\hdashline
GDumb & \textbf{88.3} {\scriptsize $ \pm $ \textbf{0.4}} & 63.6 {\scriptsize $ \pm $ 0.8} & 58.6 {\scriptsize $ \pm $ 0.6} & 56.4 {\scriptsize $ \pm $ 1.0} & 66.0 {\scriptsize $ \pm $ 1.1} & 68.7 {\scriptsize $ \pm $ 0.4} & 68.9 {\scriptsize $ \pm $ 0.6} & 65.8 {\scriptsize $ \pm $ 1.0} & 63.2 {\scriptsize $ \pm $ 1.1}\\
\hdashline
FACT & 84.3 {\scriptsize $ \pm $ 0.6} & 53.6 {\scriptsize $ \pm $ 0.6} & 43.0 {\scriptsize $ \pm $ 0.5} & 35.9 {\scriptsize $ \pm $ 0.6} & 28.3 {\scriptsize $ \pm $ 0.6} & 24.2 {\scriptsize $ \pm $ 0.3} & 22.6 {\scriptsize $ \pm $ 0.3} & 22.0 {\scriptsize $ \pm $ 0.5} & 20.6 {\scriptsize $ \pm $ 0.2} \\
FSA & 85.2 {\scriptsize $ \pm $ 0.3} & 63.3 {\scriptsize $ \pm $ 0.3} & 61.4 {\scriptsize $ \pm $ 0.7} & 61.6 {\scriptsize $ \pm $ 0.5} & 68.5 {\scriptsize $ \pm $ 0.9} & 71.2 {\scriptsize $ \pm $ 1.1} & 72.2 {\scriptsize $ \pm $ 0.5} & 71.2 {\scriptsize $ \pm $ 0.7} & 70.3 {\scriptsize $ \pm $ 0.4} \\
FSA-FiLM & \underline{87.7 {\scriptsize $ \pm $ 0.3}} & \underline{\textbf{68.5} {\scriptsize $ \pm $ \textbf{0.6}}} & \underline{\textbf{66.9} {\scriptsize $ \pm $ \textbf{0.4}}} & \underline{\textbf{66.7} {\scriptsize $ \pm $ \textbf{0.9}}} & \underline{\textbf{73.7} {\scriptsize $ \pm $ \textbf{0.6}}} & \underline{\textbf{76.0} {\scriptsize $ \pm $ \textbf{0.7}}} & \underline{\textbf{76.0} {\scriptsize $ \pm $ \textbf{0.5}}} & \underline{\textbf{75.0} {\scriptsize $ \pm $ \textbf{0.6}}} & \underline{\textbf{74.0} {\scriptsize $ \pm $ \textbf{0.3}}} \\
\bottomrule
\end{tabular}
\end{small}
\end{center}
%\endgroup
\caption{Detailed accuracy for each incremental session on \textbf{DomainNet} under the \textbf{few-shot CIL}  setting. GDumb is the only
memory-based method used for comparisons; we use a buffer size equal to the first session’s number of images $N_1$. The best results across
all methods are in bold while the best results across the no-memory methods are underlined. A pre-trained EfficientNet-B0 on Imagenet-1k
is used as a backbone for all methods.}
\label{table:few_shot_domainnet}
\end{table*}

% iNaturalist
\begin{table*}[ht]
%\vspace*{1.5 cm}
%\begingroup

\setlength{\tabcolsep}{9.7pt} % Default value: 6pt
\begin{center}
\begin{normalsize}
%\begin{sc}
\begin{tabular}{l c c c c c c c c c}
\toprule
\multirow{2}{*}{Method} & \multicolumn{9}{c}{Accuracy (\%) in each session ($\uparrow$)}  \\
\cmidrule(lr){2-10} 
   &  \textbf{1} & \textbf{2} & \textbf{3} &  \textbf{4} & \textbf{5} & \textbf{6} &  \textbf{7} & \textbf{8} & \textbf{9}\\
 \midrule
NA & 51.9 & 52.8 & 44.9 & 46.8 & 49.3 & 51.7 & 54.4 & 53.8 & 49.7\\
\hdashline
GDumb & 56.4 & 50.1 & 36.3 & 47.5 & 44.2 & 44.7 & 46.4 & 40.9 & 40.4\\
\hdashline
FACT & 54.9 & 29.9 & 24.6 & 23.8 & 20.4 & 17.8 & 15.7 & 16.4 & 14.3 \\
FSA & 52.1 & 53.6 & 40.0 & 47.8 & 49.2 & 51.3 & 55.1 & 55.1 & 51.5 \\
FSA-FiLM & \underline{\textbf{61.8}} & \underline{\textbf{61.6}} & \underline{\textbf{52.0}} & \underline{\textbf{56.5}} & \underline{\textbf{57.0}} & \underline{\textbf{59.4}} & \underline{\textbf{61.8}} & \underline{\textbf{61.2}} & \underline{\textbf{58.8}} \\
\bottomrule
\end{tabular}
%\end{sc}
\end{normalsize}    
\end{center}
\caption{Detailed accuracy for each incremental session on \textbf{iNaturalist} under the \textbf{few-shot CIL}  setting. GDumb is the only
memory-based method used for comparisons; we use a buffer size equal to the first session’s number of images $N_1$. The best results across
all methods are in bold while the best results across the no-memory methods are underlined. A pre-trained EfficientNet-B0 on Imagenet-1k
is used as a backbone for all methods.}
\label{table:few_shot_inaturalist}
%\endgroup
\end{table*}

\end{document}